\documentclass{article}

\PassOptionsToPackage{numbers, compress}{natbib}

\usepackage[preprint]{neurips_2026}     % prepring

\usepackage[utf8]{inputenc} % allow utf-8 input
\usepackage[T1]{fontenc}    % use 8-bit T1 fonts
\usepackage[hidelinks]{hyperref}       % hyperlinks
\usepackage{url}            % simple URL typesetting
\usepackage{booktabs}       % professional-quality tables
\usepackage{amsfonts}       % blackboard math symbols
\usepackage{nicefrac}       % compact symbols for 1/2, etc.
\usepackage{microtype}      % microtypography
\usepackage{xcolor}         % colors

\title{How Much of a Model Do We Need? Redundancy and Slimmability in Remote Sensing Foundation Models}

\author{%
  Leonard Hackel\thanks{Corresponding author} \\
  Big Data Analytics for Earth Observation\\BIFOLD~/~TU Berlin\\
  10587 Berlin, Germany \\
  \texttt{l.hackel@tu-berlin.de} \\
  \And
  Tom Burgert\\
  Big Data Analytics for Earth Observation\\BIFOLD~/~TU Berlin\\
  10587 Berlin, Germany \\
  \texttt{t.burgert@tu-berlin.de} \\
  \AND
  Begüm Demir\\
  Big Data Analytics for Earth Observation\\BIFOLD~/~TU Berlin\\
  10587 Berlin, Germany \\
  \texttt{demir@tu-berlin.de} \\
}
\usepackage{makecell}
\usepackage{multirow}
\usepackage{csquotes}
\usepackage{pifont}
\usepackage{siunitx}
\DeclareSIUnit{\nothing}{\relax}
\usepackage{subcaption}
\usepackage{booktabs} % for professional tables
\usepackage{placeins}
\usepackage[table]{xcolor}
\usepackage{graphicx}
\usepackage[shortlabels]{enumitem}
\usepackage{amsmath}
\usepackage{amssymb}
\usepackage{mathtools}
\usepackage{amsthm}

\usepackage[capitalize,noabbrev]{cleveref}

\usepackage{acronym}
\acrodef{AA}{average accuracy}
\acrodef{AP}{average precision}
\acrodef{AI}{artificial intelligence}
\acrodef{BN}{batch normalization}
\acrodef{C2C}[C$_2$C]{capacity-to-compute}
\acrodef{CBIR}{content-based image retrieval}
\acrodef{CD}{change detection}
\acrodef{CL}{contrastive learning}
\acrodef{CSMAE}{Cross-Sensor Masked Autoencoder}
\acrodef{CSMoE}{Cross-Sensor Mixture-of-Experts}
\acrodef{CV}{computer vision}
\acrodef{DL}{deep learning}
\acrodef{EO}{earth observation}
\acrodef{EVR}{explained variance ratio}
\acrodef{PC1}{first principal component}
\acrodef{FLOP}{floating-point operation}
\acrodef{FFN}{feed-forward network}
\acrodef{FM}{foundation model}
\acrodef{GSD}{ground sample distance}
\acrodef{IoU}{mean intersection-over-union}
\acrodef{KNN}[k-NN]{k-nearest neighbors}
\acrodef{LULC}{land-use/land-cover}
\acrodef{MAE}{masked autoencoder}
\acrodef{mAP}[mAP$_M$]{macro-mean average precision}
\acrodef{MHSA}{multi-head self-attention}
\acrodef{MiM}{mutual information maximization}
\acrodef{MIM}{masked image modeling}
\acrodef{MLP}{multi-layer perceptron}
\acrodef{MoCo}{Momentum Contrast}
\acrodef{MoE}{mixture-of-experts}
\acrodef{MSE}{mean squared error}
\acrodef{MTC}[MTom]{Major TOM Core}
\acrodef{mTC}[MTom$_\mu$]{Minor TOM Core}
\acrodef{rTC}[MTom$_r$]{Minor TOM Core (random)}
\acrodef{NTXent}[NT-Xent]{normalized temperature-scaled cross entropy}
\acrodef{PIMask}{patch-incomplete masking}
\acrodef{RS}{remote sensing}
\acrodef{S1}{Sentinel-1}
\acrodef{S2}{Sentinel-2}
\acrodef{SAR}{synthetic aperture radar}
\acrodef{SOTA}{state-of-the-art}
\acrodef{SSL}{self-supervised learning}
\acrodef{UAV}{unmanned aerial vehicle}
\acrodef{UMR}{intra-modal reconstruction loss}
\acrodef{CMR}{cross-modal reconstruction loss}
\acrodef{ViT}{vision transformer}
\acrodef{VQA}{visual question answering}

\acrodefplural{MoE}[MoEs]{mixture-of-experts}

\usepackage{calc}
\usepackage{tikz}
\usepackage{wrapfig}

\usepackage{xparse}
\usepackage{ifthen}
\NewDocumentCommand{\tlabeledimage}{m m m O{4cm} O{1mm} O{2.5mm} O{2mm} O{1mm}}{%
    \begin{tikzpicture}[trim left=(img), trim right=(img)]
        \node (img) {\includegraphics[width=#4]{#1}};
        \ifthenelse{\equal{#2}{}}{}{%
            \node[rotate=90, anchor=center] (ylabel)
                at ([xshift=#5, yshift=#6] img.west)
                {\scriptsize\textsf{#2}};
        }%
        \ifthenelse{\equal{#3}{}}{}{%
            \node[anchor=center] (xlabel)
                at ([xshift=#7, yshift=#8] img.south)
                {\scriptsize\textsf{#3}};
        }%
    \end{tikzpicture}%
}

\newcommand{\rowgrouplabel}[2][10mm]{%
    \begin{subfigure}[t]{0.03\textwidth}
        \raisebox{#1}{\rotatebox{90}{\textsf{\underline{\small{#2}}}}}%
    \end{subfigure}%
}

\newcommand{\newPageToggle}{}

\begin{document}

\maketitle

\begin{abstract}
    Large-scale \acp{FM} in \acf{RS} (denoted as \ac{RS} \acp{FM}) are developed following paradigms established in \ac{CV}, yet the validity of transferring \ac{CV} scaling laws to \ac{RS} has not been systematically examined. 
    We hypothesize that \ac{RS} \acp{FM} enter an overparameterized regime at substantially smaller scales than their \ac{CV} counterparts, with task-relevant information encoded redundantly across model dimensions.
    To test this hypothesis, we apply post-hoc slimmability, uniform width reduction of pretrained encoder transformer blocks, as a tool to measure representational redundancy across ten state-of-the-art \ac{RS} \acp{FM} on classification, segmentation, and change detection tasks. 
    \ac{RS} \acp{FM} retain on average over \SI{85}{\percent} relative accuracy on \ac{RS} classification datasets under aggressive width reduction, while \ac{MAE} and DINOv2 pretrained on natural images (denoted as \ac{CV} \ac{MAE} and \ac{CV} DINOv2) degrade sharply on ImageNet subsets of matched class count over the same range of computational requirements.
    A \ac{CV} \ac{MAE} evaluated directly on the same \ac{RS} dataset narrows but does not close the gap, indicating that both dataset characteristics and domain-specific pretraining contribute to the differences between the models.
    Analyses of the representation space of the models such as feature correlation, explained variance, and effective dimensionality indicate that task-relevant variance concentrates in few principal components and is redundantly encoded across model dimensions.
    We further show that learned slimmable training improves over post-hoc slimmability for contrastive objectives, while reconstruction-based objectives do not benefit consistently from current slimmable training protocols. 
    Our findings establish post-hoc slimming as a practical deployment strategy for resource-constrained \ac{RS} applications and as a diagnostic tool for representational redundancy in \ac{RS} \acp{FM}.
    Code is available at \url{https://github.com/hackelle/slimmable-remote-sensing-foundation-models}.
\end{abstract}

%\FloatBarrier
\acresetall
\newPageToggle

\section{Introduction} % ~1 page, little more
\Ac{RS} images acquired by satellite or airborne systems are a rich source of information to monitor the Earth surface, e.g., for climate change analysis, urban area studies, risk and damage assessment, and crop monitoring~\cite{maktav2005remote, omia2023remote, yamazaki2007remote, yang2013role}.
% eniolorunda2014climate,rahman2020systematic
With the rapid emergence of large-scale \acp{FM} trained on extensive \ac{RS} image archives (denoted as \ac{RS} \acp{FM}), practitioners have shifted from training single-purpose models to finetuning pretrained \ac{RS} \acp{FM}~\cite{jiao2023brain, lu2025fmsurvey, xiao2025foundation}.
Many of these \ac{RS} \acp{FM} follow the paradigms established in \ac{CV}, most notably \ac{SSL} paradigms such as contrastive learning (\ac{MoCo}~\cite{he2020momentum}), self-distillation (DINO~\cite{caron2021emerging}, DINOv2~\cite{oquab2024dinov}), and \ac{MIM} (\ac{MAE}~\cite{he2022masked}), as well as multi-modal pretraining strategies inspired by CLIP~\cite{radford2021learning}.
In line with prevailing practices in \ac{CV}, \ac{RS} \acp{FM} are typically scaled by increasing parameter count, with the implicit assumption that larger models yield more universal and transferable representations.
However, no prior work has tested whether the benefits of model scaling observed in \ac{CV} equally transfer to \ac{RS}.
While both domains rely on visual data, they differ in data characteristics such as spatial and spectral resolution~\cite{rolf2024position}. 
Recent analyses show that models trained on \ac{RS} data depend predominantly on low- and mid-level features, whereas those trained on natural images rely more on higher-level semantic abstractions~\cite{burgert2025imagenettrained}.
This suggests that \ac{RS} \acp{FM} may not require the same representational capacity as their \ac{CV} counterparts.

Building on these findings, we hypothesize that many \ac{RS} downstream tasks require less representational capacity than \ac{CV} downstream tasks, and that \ac{RS} \acp{FM} distribute task-relevant information redundantly across their parameters.
According to our hypothesis, uniform, task-agnostic parameter removal preserves substantial downstream performance. 
%The alternative outcome, that retaining performance requires structured or task-aware pruning, would indicate sparse encoding and contradict our hypothesis.
We therefore use uniform width reduction of pretrained encoder transformer blocks, which we call \emph{post-hoc slimmability}, as a tool to quantify representational redundancy across ten state-of-the-art \ac{RS} \acp{FM}.

Beyond its diagnostic value, post-hoc slimmability has direct practical relevance for \ac{RS}. 
Many downstream users operate under limited computational budgets, which makes deploying or finetuning increasingly large \acp{FM} difficult. 
These constraints are most severe for edge and on-board satellite inference, where memory, power, and latency are limited, and for time-critical applications such as disaster response or search and rescue. 
By adapting pretrained models to available hardware without requiring retraining or access to the original pretraining data, post-hoc slimmability provides a practical mechanism for making \acp{FM} accessible across a broad range of deployment scenarios.

\begin{figure}
    \centering
    \noindent\includegraphics[width=0.98\linewidth]{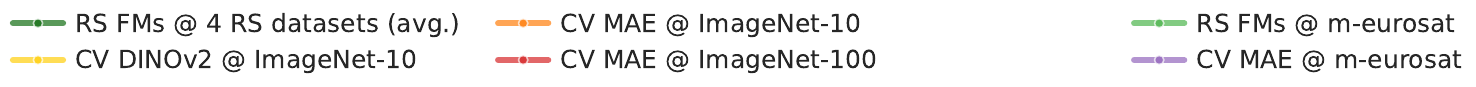}\par\vspace{-2mm}
    \hfill
    \begin{subfigure}[b]{0.46\textwidth}
        \centering        
        \tlabeledimage
            {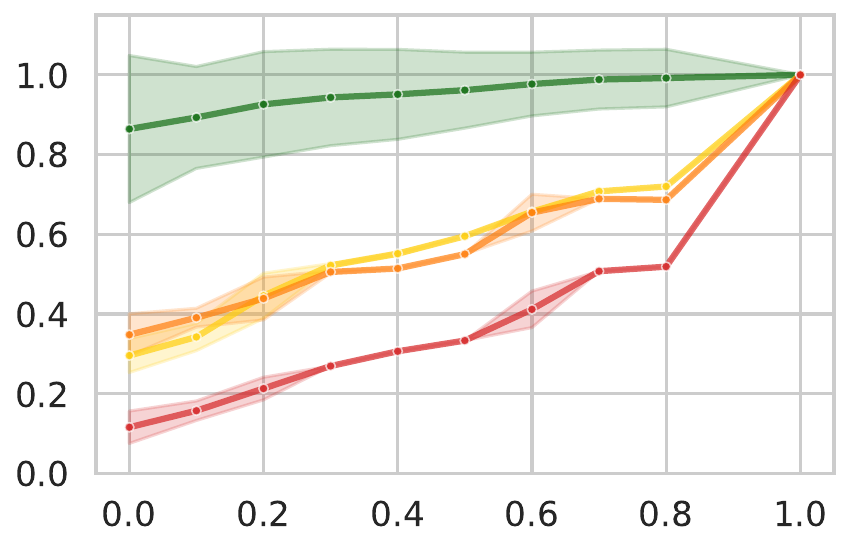}
            {Rel. Retention Rate}
            {Rel. Compute Requirement}
            [\textwidth]
        \vspace*{-3mm}
        \caption{}
        \label{fig:overview-retention}
    \end{subfigure}
    \hfill
    \begin{subfigure}[b]{0.46\textwidth}
        \centering        
        \tlabeledimage
            {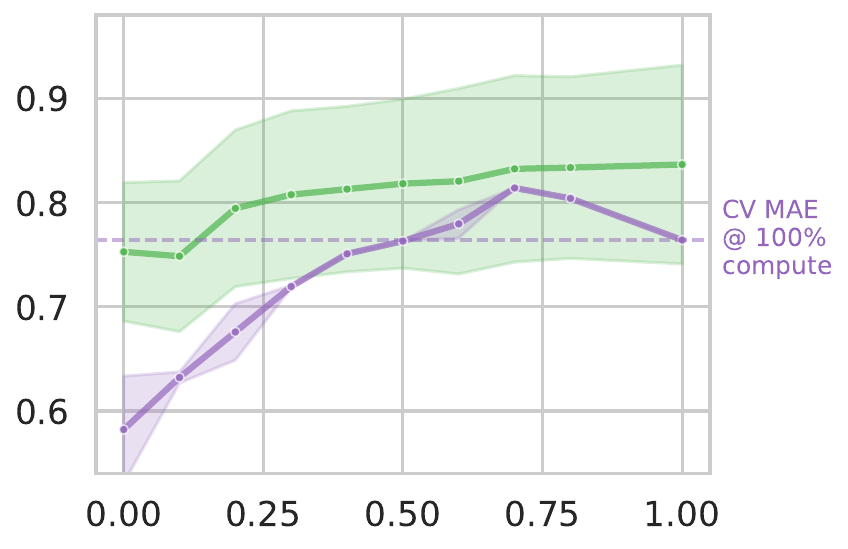}
            {Accuracy}
            {Rel. Compute Requirement}
            [\textwidth]
        \vspace*{-3mm}
        \caption{}
        \label{fig:overview-absolute}
    \end{subfigure}
    \hfill
    
    \caption{Post-hoc slimmability of \ac{RS} \acp{FM} compared with \ac{CV} baselines.
    Legends formatted as: \enquote{model(s) @ evaluation data}.
    a)~Relative retention rate: \ac{RS} \acp{FM} across four \ac{RS} classification datasets maintain on average over \SI{85}{\percent} relative accuracy at \SI{1}{\percent} compute budget, whereas \ac{CV} DINOv2 and \ac{CV} \ac{MAE} retain on average only \SIrange{30}{35}{\percent} on ImageNet-10 and a \ac{CV} \ac{MAE} only \SI{12}{\percent} on ImageNet-100 at the same compute budget.
    b)~Absolute accuracy on m-eurosat: nine \ac{RS} \acp{FM} (green) retain high accuracy even at \SI{1}{\percent} compute budget, matching or exceeding the full-scale \ac{CV} \ac{MAE} (dashed line at \num{0.76}).
    Panopticon is excluded from this comparison because its compute scales differently with width (see Appendix~\ref{sec:slimdims} for details).
    }
    \label{fig:overview}
    \vspace{-3mm}
\end{figure}

To test our hypothesis and quantify this potential, we conduct a comprehensive empirical study across ten \ac{RS} \acp{FM} ranging from \SIrange{23}{631}{\mega\nothing} parameters, evaluating post-hoc slimmability on seven downstream datasets spanning scene classification, semantic segmentation, and change detection.
As shown in \cref{fig:overview-retention}, \ac{RS} \acp{FM} slimmed to \SI{1}{\percent} of their computational budget retain on average over \SI{85}{\percent} relative accuracy across four \ac{RS} classification datasets, whereas \ac{MAE} and DINOv2 pretrained on natural images (denoted as \ac{CV} \ac{MAE} and \ac{CV} DINOv2) evaluated on ImageNet subsets of matched class count retain only \SIrange{30}{35}{\percent} at the same budget.
To isolate the contribution of domain-specific pretraining from dataset characteristics, we evaluate a \ac{CV} \ac{MAE} directly on m-eurosat~\cite{lacoste2023geobench} (\cref{fig:overview-absolute}).
Here, the \ac{CV} \ac{MAE} retains \SI{76}{\percent} relative accuracy at \SI{1}{\percent} compute budget, narrowing but not closing the gap relative to \ac{RS} \acp{FM} on the same dataset, which retain \SIrange{82}{97}{\percent}. % relative accuracy reported, absolute shown!
These results indicate that both dataset characteristics and domain-specific pretraining contribute to slimmability, with dataset characteristics accounting for most of the gap.

Our main contributions are as follows:
\begin{enumerate}[(1)]
    \item We present the first systematic study of slimmability in \ac{RS} \acp{FM}, evaluating ten models across seven datasets spanning scene classification, semantic segmentation, and change detection.
    We show that post-hoc slimming to \SI{1}{\percent} compute budget retains \SIrange{53}{103}{\percent} accuracy across classification datasets, with the majority of models retaining above \SI{85}{\percent}, and that this robustness persists for other tasks.
    Through controlled comparisons with \ac{CV} baselines under the same number of classes and on shared evaluation tasks, we show that the gap reflects both task properties and domain-specific pretraining.
    \item We compare post-hoc and learned slimmability through controlled experiments with \ac{MoCo} and \ac{MAE}, showing that learned slimmable training improves \ac{MoCo}-based models across compute budgets but interacts poorly with \ac{MAE}'s reconstruction objective, particularly on complex multi-label scene classification.
    \item We conduct a representational analysis to give an intuition on why slimming preserves performance. 
    Through intrinsic dimensionality, explained variance, and feature correlation analysis, we show that training paradigms influence how \acp{FM} organize information across scales, indicating that task-relevant variance concentrates in few principal components and is distributed redundantly across different dimensions.
\end{enumerate}

\newPageToggle
\section{Related Work} % ~ 1 page, little less
\textbf{Slimmable Networks and Model Compression.}
Slimmable neural networks enable a single model to operate at multiple widths, facilitating dynamic accuracy-efficiency trade-offs at runtime.
We distinguish two paradigms:
i) \emph{post-hoc slimmability}, where uniform width reduction is applied to pretrained models without retraining; and
ii) \emph{learned slimmability}, where models are trained with explicit width sampling from scratch.
Alternative compression techniques such as training-based structured pruning~\cite{li2017pruning, liu2017learning}, knowledge distillation~\cite{hinton2015distilling}, and quantization~\cite{jacob2018quantization} typically require access to training data for retraining, task-specific optimization, or specialized hardware, limiting their applicability under restricted data access or varying deployment constraints.
Existing work on slimmability has focused almost exclusively on the learned paradigm.
\citet{yu2019slimmable} pioneer this by introducing switchable \ac{BN}, which maintains independent \ac{BN} parameters for each width, achieving competitive ImageNet accuracy through joint training.
\citet{yu2019universally} extend this with Universally Slimmable Networks (US-Nets), which generalize to arbitrary widths via post-statistics \ac{BN} recalibration and the sandwich rule for width sampling during training.
\citet{li2021dynamic} introduce Dynamic Slimmable Networks (DS-Net) with adaptive gates for test-time width selection, achieving 2--4$\times$ computation reduction on ImageNet.
Extending slimmability to \ac{SSL} introduces additional challenges due to the interaction between multi-width optimization and representation learning.
\citet{Cao_2023_CVPR} demonstrate that naive \ac{SSL} application to slimmable networks causes training collapse, and present US3L, which implements temporally consistent guidance and dynamic width sampling to address this.
\citet{zhao2025slimmable} introduce SlimCLR with slow-start training that optimizes full-width models before progressively introducing narrower sub-networks.
However, these works focus exclusively on contrastive methods in \ac{CV}, leaving slimmable \ac{MAE} approaches and applications to \ac{RS} \acp{FM} unexplored.

Post-hoc slimmability, by contrast, has received little systematic attention as a diagnostic tool for representational redundancy.
We use it as such, not as a compression method: a sophisticated compression method that preserves accuracy leaves open whether the result reflects the method or the model, whereas accuracy preserved under simple uniform width reduction is evidence about the model itself.
The closest line of work is training-free structured pruning of pretrained large language models~\cite{men2025shortgpt, zhong2025blockpruner} for deployment, and the recent use of such pruning as a probing methodology in unified multimodal models~\cite{he2025understanding}.
Additionally, in \ac{RS}, scaling has so far only been studied during pretraining: \citet{dionelis2025scaling} pretrain separate models of \SIrange{44}{300}{\mega\nothing} parameters (U-Net, ViT-UPerNet, Mamba) on three dataset scales and find that a 200M-parameter U-Net outperforms larger architectures in low-shot regression, and \citet{burgert2026rank} find that the downstream performance of contrastive pretraining for a ResNet-18 on multispectral Sentinel-2 imagery saturates at roughly \SIrange{100}{200}{\kilo\nothing} images.
Both results are compatible with, but do not test, the redundancy we observe.
We instead take single released checkpoints and ask how much of their width is needed at inference without retraining, so the two views are complementary.
However, no prior work has investigated post-hoc slimmability as a property of pretrained models in \ac{RS}, leaving open whether representational redundancy alone is sufficient to sustain performance under width reduction without any retraining.

\textbf{Remote Sensing Foundation Models.}
\begin{table*}[t]
\centering
\renewcommand{\arraystretch}{0.9}
\setlength\tabcolsep{4pt}

\caption{List of selected \ac{RS} \acp{FM}. Models evaluated in this work are highlighted in gray. \enquote{---} indicates information not publicly available. $^{\text{\ding{59}}}$ multi-modal; $^{\text{\ding{79}}}$ multitemporal. For models with multiple variants, ranges indicate min--max values. Image resolution for forward pass: $224\times224$.}
\label{tab:rsfm_overview}

\begin{tabular}{lrrrr}
\toprule
\textbf{Model} & \makecell{\textbf{\#Pretraining}\\\textbf{Images}} & \makecell{\textbf{Compute}\\Pretraining} & \makecell{\textbf{Parameters}} & \makecell{\textbf{GFLOPs}/\\forward pass} \\
\midrule
SatMAE~\cite{cong2022satmae}              & \SI{713}{\kilo\nothing}                              & \SI{768}{\hour} (V100)                        & \SI{330}{\mega\nothing}                            & --- \\
%RingMo~\cite{sun2022ringmo}               & \SI{2.1}{\mega\nothing}$^\text{\ding{59}}$           & ---                                           & ---                                                & --- \\
%ScaleMAE~\cite{reed2023scale}             & \SI{364}{\kilo\nothing}                              & ---                                           & \SI{323}{\mega\nothing}                            & --- \\
\rowcolor{gray!15}
SSL4EO~\cite{wang2023ssl4eo}              & \SI{251}{\kilo\nothing}$^\text{\ding{59}\ding{79}}$  & \SI{100}{\hour} (A100)                        & \SI{23}{\mega\nothing} -- \SI{87}{\mega\nothing}   & 4.44 -- 17.14 \\
Satlas~\cite{bastani2023satlas}           & \SI{50.4}{\mega\nothing}$^\text{\ding{59}\ding{79}}$ & ---                                           & \SI{86}{\mega\nothing}                             & 17.12 \\
SkySense~\cite{guo2024skysense}           & \SI{21.5}{\mega\nothing}$^\text{\ding{59}}$          & \SI{24.6}{\kilo\hour} (A100)                  & \SI{2.06}{\bel}                                    & --- \\
%SkySense V2~\cite{zhang2025skysenseV2}    & \SI{21.5}{\mega\nothing}$^\text{\ding{59}}$          & \SI{44.5}{\kilo\hour} (A20)                       & \SI{665}{\mega\nothing} (1.26B)                    & --- \\
%Prithvi-EO-1.0~\cite{prithvi_v1}          & \SI{4.2}{\mega\nothing}$^\text{\ding{59}}$           & \SI{846}{\hour} (A100)                        & \SI{86}{\mega\nothing}                             & 16.98 \\
\rowcolor{gray!15}
Prithvi-EO-2.0~\cite{prithvi_v2}          & \SI{4.2}{\mega\nothing}$^\text{\ding{59}\ding{79}}$  & \SI{21}{\kilo\hour} -- \SI{58}{\kilo\hour} (A100) & \SI{304}{\mega\nothing} -- \SI{631}{\mega\nothing} & 59.85 -- 162.18 \\
\rowcolor{gray!15}
DOFA~\cite{xiong2024neural}               & \SI{11.5}{\mega\nothing}$^\text{\ding{59}}$          & \SI{576}{\hour} (L40)                         & \SI{86}{\mega\nothing} -- \SI{330}{\mega\nothing}  & 17.47 -- 60.47 \\
\rowcolor{gray!15}
TerraMind-1.0~\cite{jakubik2025terramind} & \SI{9}{\mega\nothing}$^\text{\ding{59}}$             & \SI{9.2}{\kilo\hour} (A100)                   & \SI{87}{\mega\nothing} -- \SI{305}{\mega\nothing}  & 17.83 -- 61.73 \\
\rowcolor{gray!15}
Panopticon~\cite{waldmann2025panopticon}  & \SI{2.5}{\mega\nothing}$^\text{\ding{59}}$           & ---                                           & \SI{86}{\mega\nothing}                             & 59.16 \\
\rowcolor{gray!15}
CopernicusFM~\cite{wang2025towards}       & \SI{18.7}{\mega\nothing}$^\text{\ding{59}}$          & \SI{512}{\hour} (A100)                        & \SI{86}{\mega\nothing}                             & 17.67 \\
RingMoE~\cite{bi2025ringmoe}              & \SI{400}{\mega\nothing}$^\text{\ding{59}}$           & ---                                           & \SI{14.7}{\bel}                                    & --- \\
%RS-vHeat \cite{hu2025rs}                  & \SI{450}{\kilo\nothing}$^\text{\ding{59}}$           & ---                                           & ---                                                & --- \\
CSMoE~\cite{hackel2025csmoe}              & \SI{1.1}{\mega\nothing}$^\text{\ding{59}}$           & ---                                           & \SI{271}{\mega\nothing} -- \SI{277}{\mega\nothing} &  2.92 -- 13.40 \\
\bottomrule
\end{tabular}
\vspace{-3mm}
\end{table*}

With advances in \ac{SSL} and the increasing availability of large-scale \ac{RS} data, the development of \acp{FM} has gained significant attention for learning general-purpose \ac{RS} representations.
These models leverage \ac{SSL} objectives such as contrastive learning or \ac{MIM} for pretraining by using different data modalities.
As shown in \Cref{tab:rsfm_overview}, widely used \ac{RS} \acp{FM} span multiple orders of magnitude in parameters and computational budgets.
For instance, Prithvi~\cite{prithvi_v1, prithvi_v2} leverages Sentinel-2 and Landsat time-series data, while Satlas~\cite{bastani2023satlas} introduces 300M+ annotations for supervision, and SSL4EO-S12~\cite{wang2023ssl4eo} provides \ac{ViT} encoders pretrained via contrastive objectives on a large-scale multi-seasonal Sentinel-2 dataset.
Beyond scaling, architectural innovations have focused on domain-specific adaptations.
First, temporal and spectral modeling advances through SatMAE~\cite{cong2022satmae} with temporal-spectral encodings and ScaleMAE~\cite{reed2023scale} with resolution-aware embeddings.
Second, modality-agnostic designs have emerged, including DOFA~\cite{xiong2024neural}, Panopticon~\cite{waldmann2025panopticon}, and CopernicusFM~\cite{wang2025towards}, which dynamically adapt to channel configurations, and TerraMind~\cite{jakubik2025terramind}, which enables cross-modal generation.
Third, multi-modal integration is addressed by RingMoE~\cite{bi2025ringmoe}, which employs hierarchical \ac{MoE} with modal-specialized experts.

More recently, efficiency considerations have motivated alternative architectural approaches.
For example, RS-vHeat~\cite{hu2025rs} replaces attention mechanisms with heat conduction operators to reduce complexity, while CSMoE~\cite{hackel2025csmoe} achieves 2$\times$ computational efficiency by utilizing the Soft \ac{MoE} mechanisms.
However, these models are optimized for fixed efficiency-accuracy trade-offs at design time, consequently lacking the runtime adaptability needed for dynamic resource constraints.
Despite these advances, no prior research has investigated whether \ac{RS} \acp{FM} exhibit different slimmability properties compared to \ac{CV} models, nor explored the representational characteristics governing post-hoc slimmability in this domain.

\newPageToggle
\section{Problem Setup and Evaluation Protocol} % ~1 page
\subsection{Slimmable Network Architecture}
We investigate the slimmability~\cite{yu2019universally} of \ac{RS} \acp{FM} by implementing width reduction in transformer blocks. 
For this investigation, we only modify the \ac{FFN} hidden dimension and the per-head dimensions of the \ac{MHSA} layers, so that the computational budget can be varied at inference time.
The embedding dimension stays unchanged, so every normalization layer operates on the same feature dimension at all widths and needs no recalibration (see Appendix~\ref{sec:slimdims} for details).
In addition, we analyze the individual layer types in Appendix~\ref{sec:ffn_vs_attn}.

\textbf{Slimmable \ac{FFN}.} For a standard \ac{FFN} with hidden dimension $d_h$, we implement a slimmable variant that operates on a reduced dimension $d_h' = \lfloor s \cdot d_h \rfloor$, where $s \in (0, 1]$ is the scaling factor. 
Given input $\mathbf{x} \in \mathbb{R}^{N \times d}$, the forward pass becomes:
\begin{align}
\mathbf{h} &= \mathrm{Act}(\mathbf{x} \mathbf{W}_1[:d_h', :]^\top + \mathbf{b}_1[:d_h']), \\
\mathbf{y} &= \mathbf{h} \mathbf{W}_2[:, :d_h']^\top + \mathbf{b}_2,
\end{align}
where $\mathbf{W}_1 \in \mathbb{R}^{d_h \times d}$, $\mathbf{W}_2 \in \mathbb{R}^{d \times d_h}$, 
$\mathbf{b}_1 \in \mathbb{R}^{d_h}$, and $\mathbf{b}_2 \in \mathbb{R}^d$.

\textbf{Slimmable \ac{MHSA}.} For \ac{MHSA} with $H$ heads and head dimension $d_k$, we reduce each head to $d_k' = \lfloor s \cdot d_k \rfloor$ while keeping all $H$ heads. 
For head $h$, we select per-head row indices $\mathcal{I}_h = [h \cdot d_k, h \cdot d_k + d_k')$ and define $\mathcal{I} = \bigcup_{h=1}^{H} \mathcal{I}_h$. 
The query, key, value, and output projections are sliced as:
\begin{align}
\mathbf{Q} &= \mathbf{x} \mathbf{W}_Q[\mathcal{I}, :]^\top + \mathbf{b}_Q[\mathcal{I}], \\
\mathbf{K} &= \mathbf{x} \mathbf{W}_K[\mathcal{I}, :]^\top + \mathbf{b}_K[\mathcal{I}], \\
\mathbf{V} &= \mathbf{x} \mathbf{W}_V[\mathcal{I}, :]^\top + \mathbf{b}_V[\mathcal{I}], \\
\mathbf{x}_{\text{attn}} &= \mathrm{softmax}(\mathbf{Q}\mathbf{K}^\top / \sqrt{d_k'})\mathbf{V}, \\
%\mathbf{x}_{\text{attn}} &= \mathrm{softmax}\left(\frac{\mathbf{Q}\mathbf{K}^\top}{\sqrt{d_k'}}\right)\mathbf{V}, \\
\mathbf{y} &= \mathbf{x}_{\text{attn}} \mathbf{W}_O[:, \mathcal{I}]^\top + \mathbf{b}_O,
\end{align}
where $\mathbf{x} \in \mathbb{R}^{N \times d}$, $\mathbf{W}_{\{Q,K,V\}} \in \mathbb{R}^{(H \cdot d_k) \times d}$, 
$\mathbf{W}_O \in \mathbb{R}^{d \times (H \cdot d_k)}$, and the biases have matching dimensions.
Both slimming operations retain the first $d_h'$ dimensions of the \ac{FFN} hidden layer and the first $d_k'$ dimensions of each attention head. 
We empirically verify in the Appendix \ref{sec:implicit_order} that first, last, and random orderings yield comparable 
performance across models and scales, with any ordering effect small relative to the retention gap between \ac{RS} \acp{FM} and \ac{CV} models.

\textbf{Post-hoc vs.\ Learned Slimmability.} 
We distinguish two paradigms:
i) \emph{post-hoc slimmability} applies uniform width reduction to pretrained encoder transformer blocks while preserving learned weight values, requiring no retraining; and
ii) \emph{learned slimmability} trains models from scratch with slimmable layers and explicit width sampling during training, encouraging weight sharing and robustness across scales.
Details of the learned slimmability training protocol are given in the Appendix \ref{sec:training_from_scratch}.

\subsection{Evaluation Protocol}
\label{sec:eval_protocol}

\textbf{Datasets.}
We evaluate our hypothesis on seven downstream datasets spanning three task types from the geobench benchmark collection~\cite{lacoste2023geobench} and the OSCD dataset~\cite{daudt2018oscd}.
In detail, we select:
i) m-brick-kiln (binary brick-kiln/no-brick-kiln classes), m-eurosat (single-label, 10 \ac{LULC} classes), m-so2sat (single-label, 17 local climate zone classes), and m-bigearthnet (multi-label, 43 \ac{LULC} classes) for scene classification; 
ii) m-cashew-plant (7 classes) and m-SA-crop-type (10 crop-type classes) for semantic segmentation; 
and iii) OSCD (binary change/no-change classes) for \ac{CD}.

\textbf{Feature Extraction and Task Heads.}
For each model at scale $s$, we extract features from the frozen encoder. 
For classification, we use the penultimate layer and train a \ac{KNN} classifier; we report macro-averaged accuracy for single-label datasets and \ac{mAP} for m-bigearthnet.
For dense predictions (semantic segmentation and \ac{CD}), we extract multi-scale features from four equally spaced encoder layers and train a lightweight head that fuses them via concatenation and a linear projection before classifying each spatial position.
Logits are bilinearly upsampled to the input resolution.
For \ac{CD}, both bi-temporal images are processed and the feature difference is fed to the head.
The heads are trained for 30 epochs with AdamW (learning rate $0.01$; weight decay $0.01$ for segmentation, $0.1$ for \ac{CD}) using cross-entropy loss (class-weighted for \ac{CD} to address the severe class imbalance in OSCD).
We report \ac{IoU} for both tasks. 
Details on the \ac{KNN} protocol, segmentation head and linear probing results are provided in the Appendix~\ref{sec:protocol} and~\ref{sec:linear_results}.

\textbf{Metrics.}
We evaluate each model at 31 scales $s$ ranging from \SIrange{0.1}{100}{\percent} (see the Appendix~\ref{subsec:scale_details}).
We report the \emph{relative compute requirement} as the ratio of \acp{FLOP} at scale $s$ to \acp{FLOP} at full scale, and the \emph{relative retention rate} $\text{RetentionRate}(s) = \text{metric}(s) / \text{metric}(1.0)$, which we use when comparing datasets with different absolute performance levels.

\textbf{Representational Analysis.}
To characterize how \ac{RS} \acp{FM} behave under width reduction, we describe three representational properties of the features at each scale (see Appendix~\ref{sec:eff_dim} for details):
i)~the mean feature correlation, i.e., the mean of $|\text{corr}(\mathbf{e}_i,\mathbf{e}_j)|$ over all pairs $i<j$ (without self-correlations), measuring average redundancy between retained dimensions;
ii)~the log-log slope of the \ac{EVR}, i.e., the slope of $\log(\text{EVR})$ against $\log$ of the principal component index, where steeper negative values indicate that variance is concentrated in fewer components (per-scale \ac{EVR} curves are provided in Appendix~\ref{sec:add_evr}); and
iii)~the effective dimensionality $d_{\text{eff}}=(\sum_i\lambda_i)^2/\sum_i\lambda_i^2$, measuring how uniformly variance is spread across dimensions.
We would like to note that these analyses are descriptive only and do not establish why slimming preserves performance.

\newPageToggle
\section{Experiments}
\subsection{Post-Hoc Slimmability of Existing RS FMs}\label{sec:post-hoc}
\begin{figure*}
    \centering
    \noindent\includegraphics[width=0.98\linewidth]{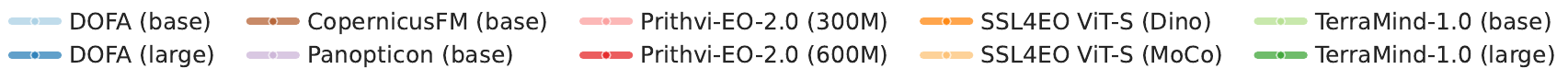}\par\vspace{-2mm}
    \begin{subfigure}[b]{0.40\textwidth}
        \centering        
        \tlabeledimage
            {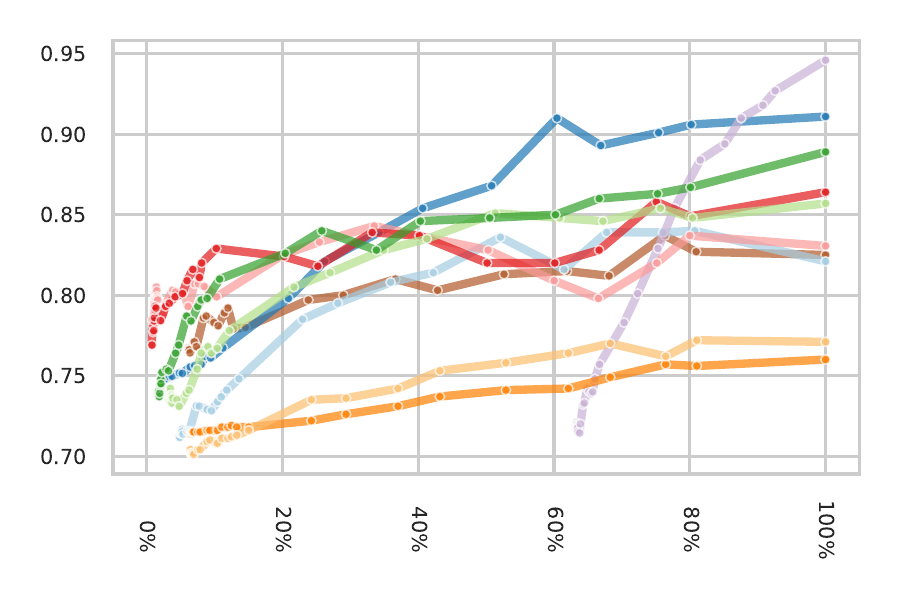}
            {Accuracy}
            {Rel. Compute Requirement}
            [\textwidth]
        \vspace*{-3mm}
        \caption{}
        \label{fig:post-hoc-eurosat}
    \end{subfigure}
    \hspace{5mm}
    \begin{subfigure}[b]{0.40\textwidth}
        \centering        
        \tlabeledimage
            {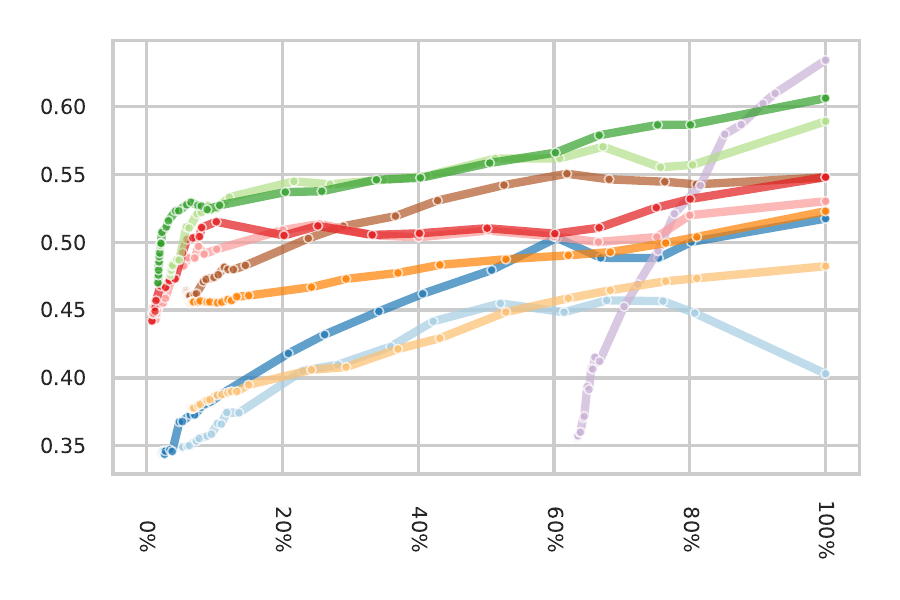}
            {\ac{mAP}}
            {Rel. Compute Requirement}
            [\textwidth]
        \vspace*{-3mm}
        \caption{}
        \label{fig:post-hoc-bigearthnet}
    \end{subfigure}

    \centering
    \begin{subfigure}[b]{0.40\textwidth}
        \centering        
        \tlabeledimage
            {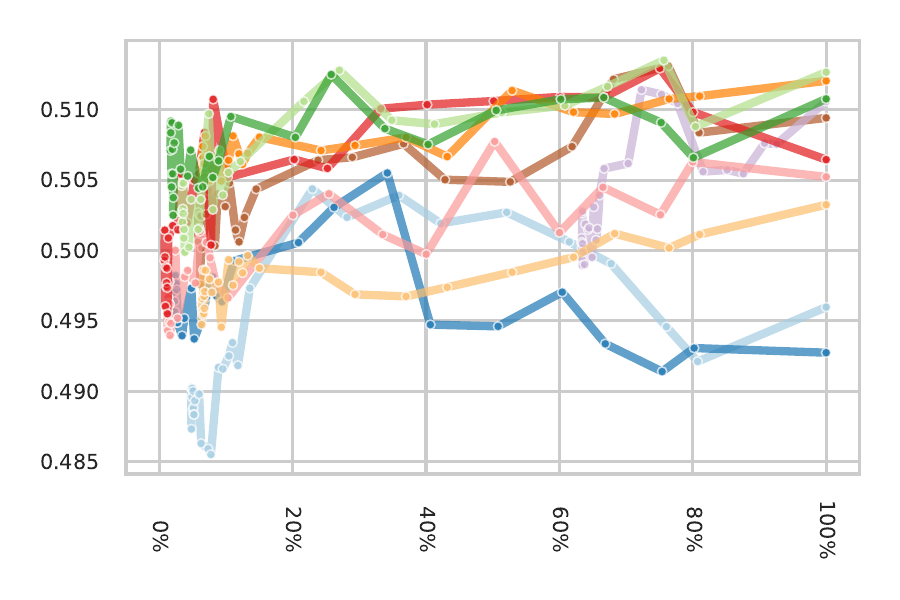}
            {\ac{IoU}}
            {Rel. Compute Requirement}
            [\textwidth]
        \vspace*{-3mm}
        \caption{}
        \label{fig:post-hoc-cashew}
    \end{subfigure}
    \hspace{5mm}
    \begin{subfigure}[b]{0.40\textwidth}
        \centering        
        \tlabeledimage
            {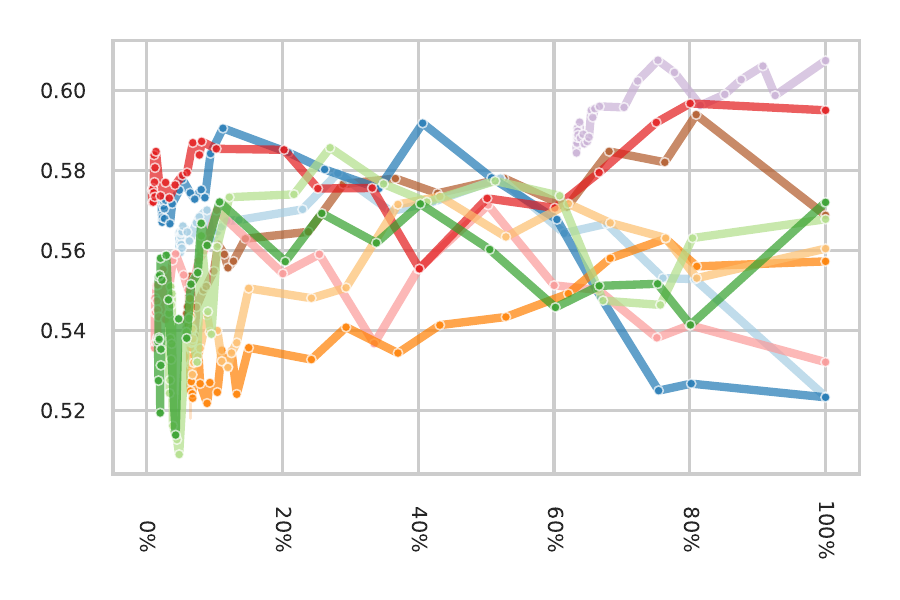}
            {\ac{IoU}}
            {Rel. Compute Requirement}
            [\textwidth]
        \vspace*{-3mm}
        \caption{}
        \label{fig:post-hoc-OSCD}
    \end{subfigure}
    
    \caption{Downstream performance with respect to relative compute requirements of ten pretrained \ac{RS} \acp{FM} on:
        a)~m-eurosat (single-labels scene classification);
        b)~m-bigearthnet (multi-label scene classification);
        c)~m-cashew-plant (semantic segmentation);
        d)~OSCD (change detection).
        Results for m-brick-kiln, m-so2sat, and m-SA-crop-type are provided in Appendix \ref{sec:add_post-hoc}.}
    \label{fig:post-hoc}
\end{figure*}

We evaluate the post-hoc slimmability of ten state-of-the-art \ac{RS} \acp{FM} across four classification, two segmentation, and one change detection datasets.
As shown in \cref{fig:post-hoc}, \ac{RS} \acp{FM} exhibit robust post-hoc slimmability across all tasks, maintaining substantial performance even at extreme slimming.
We show the results obtained by using two classification, one segmentation, and one change detection dataset here, with the results of the remaining two classification and one segmentation dataset provided in the Appendix \ref{sec:add_post-hoc}.

\paragraph{Classification.}
All models with a full compute range maintain \SIrange{82}{97}{\percent} relative retention on m-eurosat at their minimum compute budget (\cref{fig:post-hoc-eurosat}), with gradual degradation beginning below \SI{20}{\percent} compute.
Panopticon is the exception: its unslimmed embedding keeps relative compute above roughly \SI{60}{\percent} (\cref{sec:slimdims}), and it retains \SI{76}{\percent} at its smallest scale.
On the more challenging multi-label m-bigearthnet (\cref{fig:post-hoc-bigearthnet}), eight of nine models retain \SIrange{78}{87}{\percent} at minimum compute budget.
DOFA (large) shows a larger drop to \SI{67}{\percent}, indicating that this model is more sensitive to extreme width reduction on multi-label tasks, and Panopticon retains \SI{57}{\percent} at its smallest scale.
Both datasets exhibit non-monotonic behavior where intermediate scales frequently outperform the respective full-scale models, except for Panopticon, whose performance increases monotonically with compute.
This behavior is consistent with full-scale models encoding redundant features.
The critical compute threshold below which performance drops sharply appears to track dataset complexity.
In detail, minimum compute suffices for stable performance on m-eurosat, while m-bigearthnet requires \SIrange{2}{10}{\percent} compute.
The exception is Panopticon which cannot be assessed here, as it does not reach this range.
Results for m-brick-kiln and m-so2sat follow the same pattern and are provided in the Appendix \ref{sec:add_post-hoc}.

\paragraph{Dense Prediction.}
Slimmability extends to dense prediction with even higher robustness than classification.
On m-cashew-plant (\cref{fig:post-hoc-cashew}), all \ac{RS} \acp{FM} retain \SIrange{97}{101}{\percent} of full-scale \ac{IoU} at their respective smallest scale, including Panopticon at its smallest scale (relative compute above roughly \SI{60}{\percent}, see Appendix~\ref{sec:slimdims}), with curves effectively indistinguishable from full scale across the entire compute range.
The more complex 10-class m-SA-crop-type shows reduced but still robust retention of \SIrange{78}{94}{\percent} at minimum compute budget, with Panopticon retaining \SI{84}{\percent} at its smallest scale (Appendix~\ref{sec:add_post-hoc}).
On OSCD (\cref{fig:post-hoc-OSCD}), all models retain almost full performance of full-scale \ac{IoU} at minimum compute budget. 
Both DOFA models, CopernicusFM, and Prithvi-EO-2.0 (300M) exceed full scale, while Prithvi-EO-2.0 (600M), Panopticon, both TerraMind models, and both SSL4EO models retain more than \SI{90}{\percent}.
The non-monotonic pattern mirrors that observed across classification tasks, which shows that slimmability is not specific to a single task type.

\paragraph{Model scale and architecture effects.}
Larger models within each architecture (DOFA (large), Prithvi-EO-2.0 (600M), TerraMind-1.0 (large)) generally show smoother degradation curves than their smaller counterparts across classification tasks. 
DOFA exhibits the most pronounced non-monotonic peaks across all evaluated tasks, and full-scale performance on m-bigearthnet falls below the maximum for DOFA (base). 
Prithvi maintains high retention at extreme slimming on classification, while CopernicusFM and both SSL4EO models degrade smoothly with little overshoot above full scale.
Panopticon increases monotonically with compute over the range it covers and reaches the highest full-scale performance on m-eurosat and m-bigearthnet, but also shows the steepest decline.

\paragraph{Comparison with CV models.}
\ac{RS} \acp{FM} degrade gradually across the compute range, retaining \SIrange{82}{97}{\percent} on m-eurosat down to their minimum compute budget.
In contrast, as shown in \cref{fig:overview-retention}, \ac{CV} \ac{MAE} and DINOv2 on ImageNet at matched class count (10 classes) degrade sharply over the same range, retaining less than \SI{35}{\percent} at \SI{1}{\percent} compute budget; \ac{CV} \ac{MAE} on ImageNet-100 drops to \SI{12}{\percent}.
Evaluating the \ac{CV} \ac{MAE} directly on m-eurosat (\cref{fig:overview-absolute}) flattens its curve relative to ImageNet but leaves it steeper than the \ac{RS} \acp{FM}, with retention improving to \SI{76}{\percent} at \SI{1}{\percent} compute budget.
The flattening when the \ac{CV} model is moved to an \ac{RS} evaluation domain indicates that task properties contribute to slimmability; the residual gap to the \ac{RS} \acp{FM} indicates that domain-specific pretraining contributes additionally.
The contrast lies in the shape of the slimming behavior rather than in any single retention value.

\newPageToggle
\subsection{Learned vs Post-Hoc Slimmability} % ~1 page, little more
\begin{figure*}[t]
    \centering    
    \noindent\includegraphics[width=0.98\linewidth]{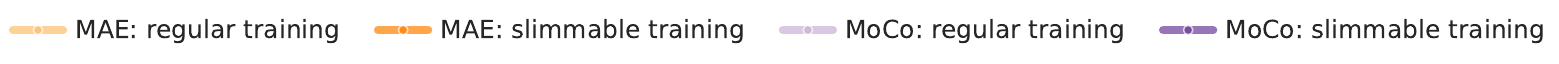}\par\vspace{-2mm}
    \begin{subfigure}[b]{0.40\textwidth}
        \centering        
        \tlabeledimage
            {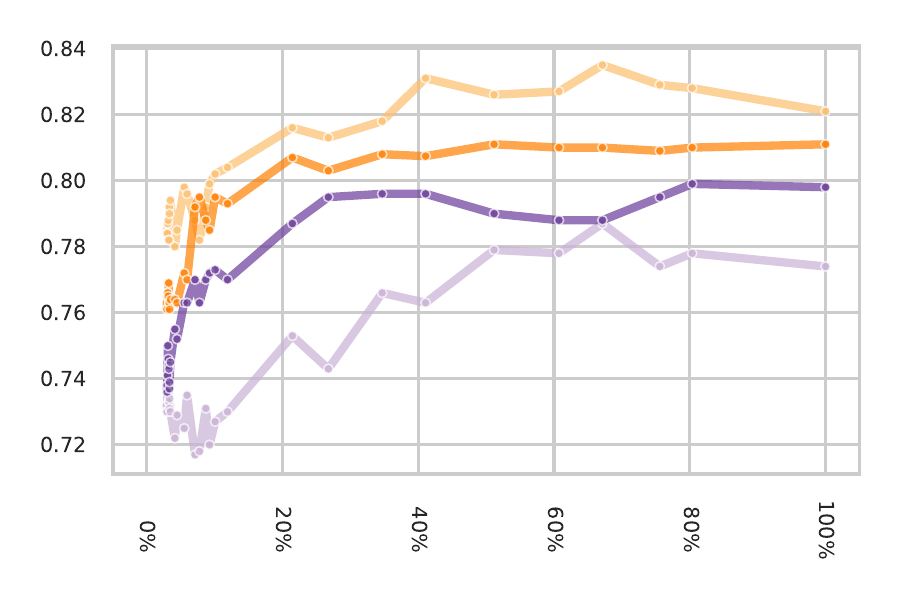}
            {Accuracy}
            {Rel. Compute Requirement}
            [\textwidth]
        \vspace*{-3mm}
        \caption{}
        \label{fig:train-son-eurosat}
    \end{subfigure}
    \hspace{5mm}
    \begin{subfigure}[b]{0.40\textwidth}
        \centering        
        \tlabeledimage
            {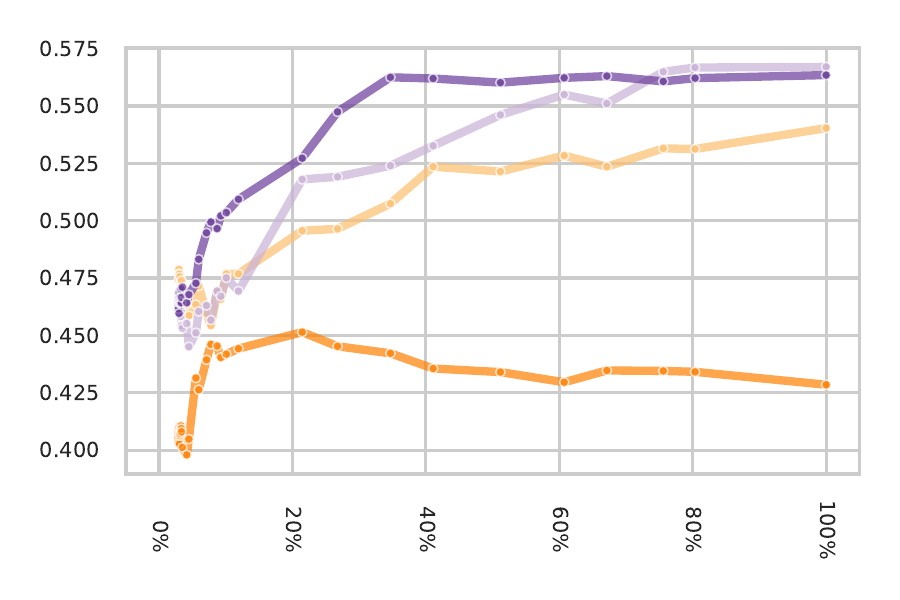}
            {\ac{mAP}}
            {Rel. Compute Requirement}
            [\textwidth]
        \vspace*{-3mm}
        \caption{}
        \label{fig:train-son-ben}
    \end{subfigure}
    
    \caption{\ac{KNN} classification performance with respect to their relative compute requirements of a \ac{MAE}-based and a \ac{MoCo}-based trained model from scratch, each once with and once without slimmable training on:
    a)~m-eurosat; and b)~m-bigearthnet.}
    \label{fig:train-son}
\end{figure*}
To investigate whether explicit slimmable training improves over post-hoc slimmability under controlled conditions, we train \ac{ViT}-Base models from scratch under both \ac{MoCo} and \ac{MAE} pretraining objectives, comparing learned slimmable training against regular training.
All models are trained on a curated \ac{RS} dataset following \cite{hackel2025csmoe}; an ablation on pretraining length is provided in the Appendix \ref{sec:training_from_scratch}.
Results are shown in \cref{fig:train-son}; results for the remaining datasets are provided in the Appendix \ref{sec:add_learned_v_post-hoc}.

\paragraph{MoCo benefits consistently from learned slimmability; MAE does not.}
\ac{MoCo}-based learned slimmable training consistently outperforms regular training at low compute budgets while maintaining equivalent or superior performance at higher scales.
On m-eurosat, the advantage of slimmable \ac{MoCo} over regular \ac{MoCo} is modest at minimum compute budget and increases at intermediate compute budgets.
On the more challenging multi-label m-bigearthnet (\cref{fig:train-son-ben}), both \ac{MoCo} variants reach similar full-scale performance, but the slimmable variant achieves near-peak accuracy at substantially lower compute budgets.
This pattern suggests that explicit multi-scale optimization structures \ac{MoCo} representations to be more robust under width reduction without sacrificing peak performance.
\ac{MAE}-based learned slimmable training shows no such benefit.
On m-eurosat, regular \ac{MAE} outperforms learned slimmable \ac{MAE} across nearly all scales.
On m-bigearthnet, the slimmable \ac{MAE} additionally degrades at larger scales. 
The advantage of \ac{MoCo} over \ac{MAE} on m-bigearthnet holds for both slimmable and regular training, consistent with prior findings that contrastive objectives promote greater linear separability on multi-label tasks \citep{he2022masked, kang2026contrastive}, and is therefore not specific to the slimmability protocol.
To our knowledge, no established slimmable training procedure exists for reconstruction-based objectives, and the current protocol does not appear to transfer from contrastive to reconstructive pretraining.

\newPageToggle
\subsection{Representational Analysis of Slimmability}
\begin{figure*}[ht]
    \centering
    \noindent\includegraphics[width=0.98\linewidth]{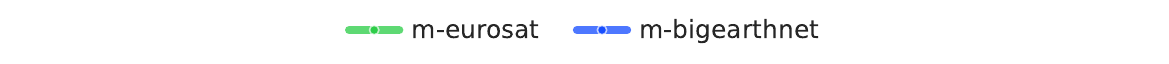}\par\vspace{-2mm}
    \begin{subfigure}[b]{0.32\textwidth}
        \centering        
        \tlabeledimage
            {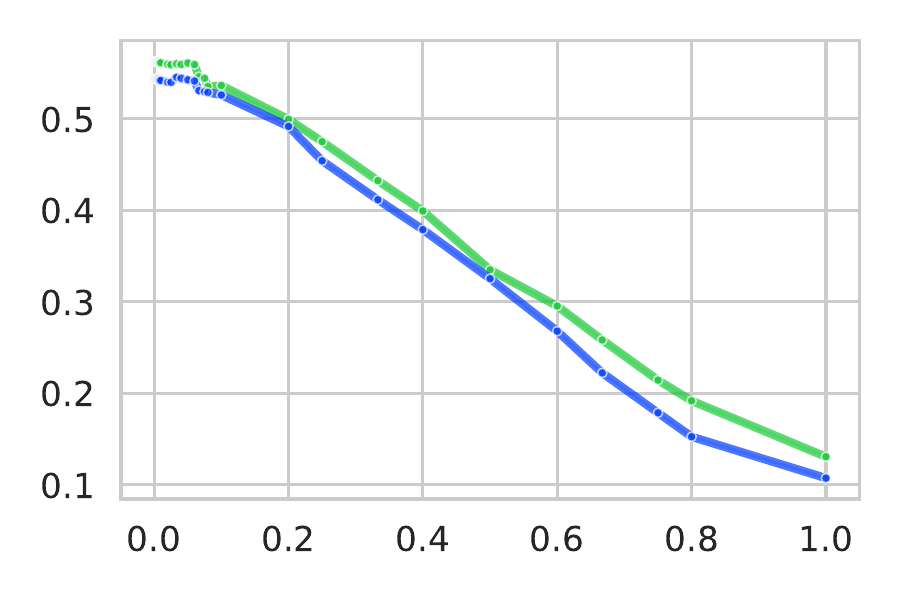}
            {Mean Feat. Corr.}
            {\phantom{Rel. Compute Requirement}}
            [\textwidth]
        \vspace*{-2mm}
        \caption{}
        \label{fig:corr_large_dofa_large}
    \end{subfigure}
    \begin{subfigure}[b]{0.32\textwidth}
        \centering        
        \tlabeledimage
            {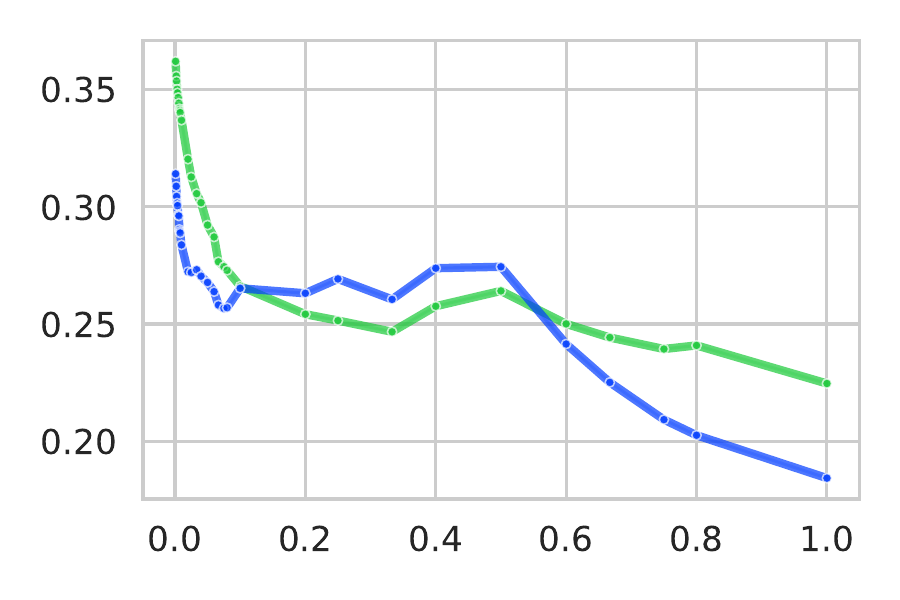}
            {\phantom{Mean Feat. Corr.}}
            {Rel. Compute Requirement}
            [\textwidth]
        \vspace*{-2mm}
        \caption{}
        \label{fig:corr_large_terramind_large}
    \end{subfigure}
    \begin{subfigure}[b]{0.32\textwidth}
        \centering        
        \tlabeledimage
            {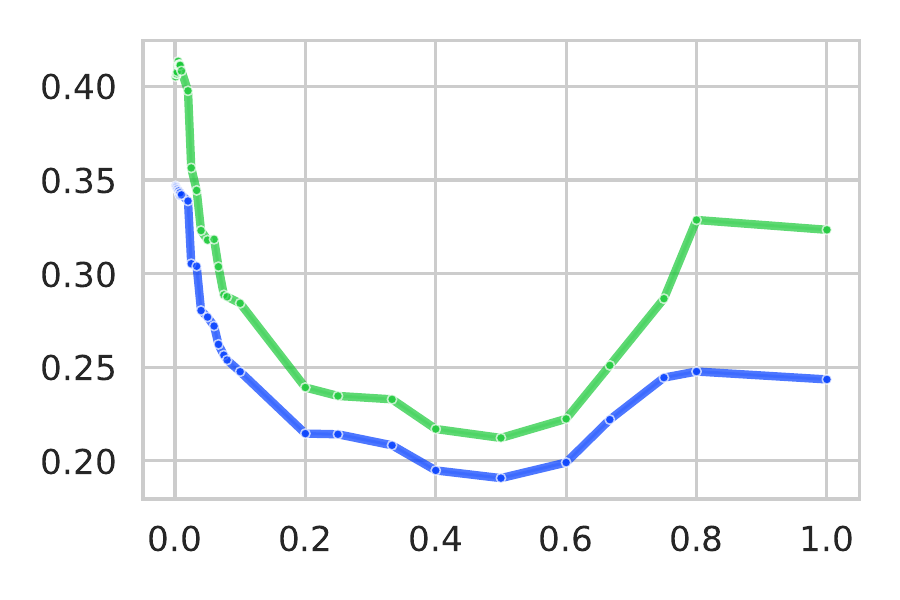}
            {\phantom{Mean Feat. Corr.}}
            {\phantom{Rel. Compute Requirement}}
            [\textwidth]
        \vspace*{-2mm}
        \caption{}
        \label{fig:corr_large_prithvi_600m}
    \end{subfigure}
    
    \caption{Mean feature correlation of the test split of m-eurosat and m-bigearthnet from the geobench benchmark collection with respect to their relative compute requirements on pretrained \ac{RS} \acp{FM}:
    a)~DOFA (large); b)~TerraMind-1.0 (large); and c)~Prithvi-EO-2.0 (600M).}
    \label{fig:corr_large}
\end{figure*}
\begin{figure*}[ht]
    \centering
    \noindent\includegraphics[width=0.98\linewidth]{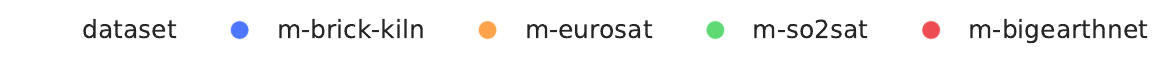}\par\vspace{-2mm}
    \noindent\includegraphics[width=0.98\linewidth]{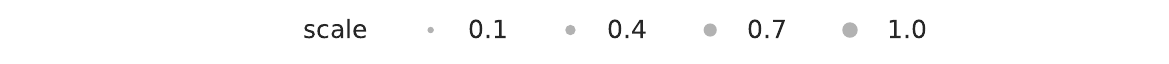}\par\vspace{-3mm}
    \begin{subfigure}[t]{0.32\textwidth}
        \centering
        \tlabeledimage
            {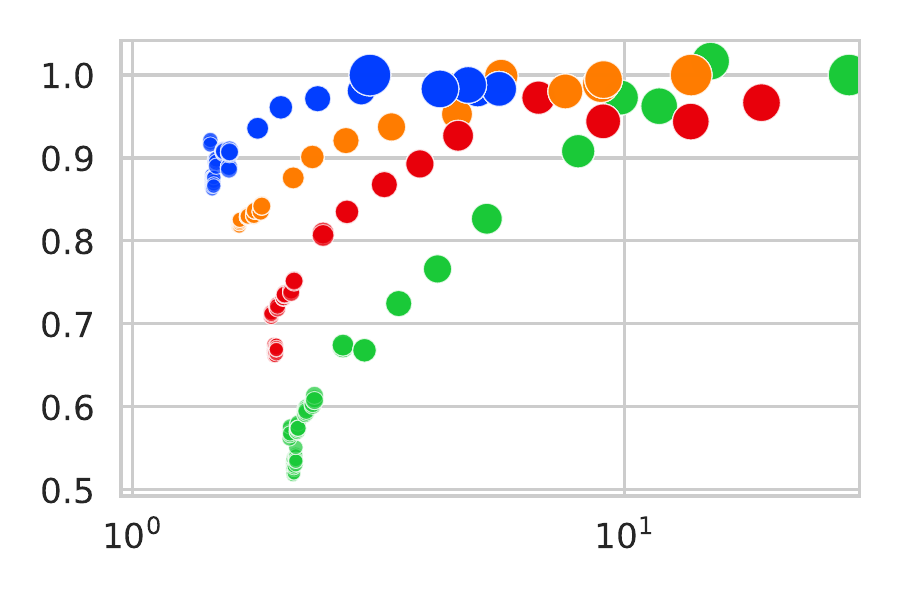}
            {\hspace{-3mm}Rel. Retention Rate}
            {\phantom{Effective Dimension}}
            [\textwidth]
        \vspace*{-3mm}
        \caption{}
        \label{fig:effDim_large-dofa-large}
    \end{subfigure}%
    \begin{subfigure}[t]{0.32\textwidth}
        \centering
        \tlabeledimage
            {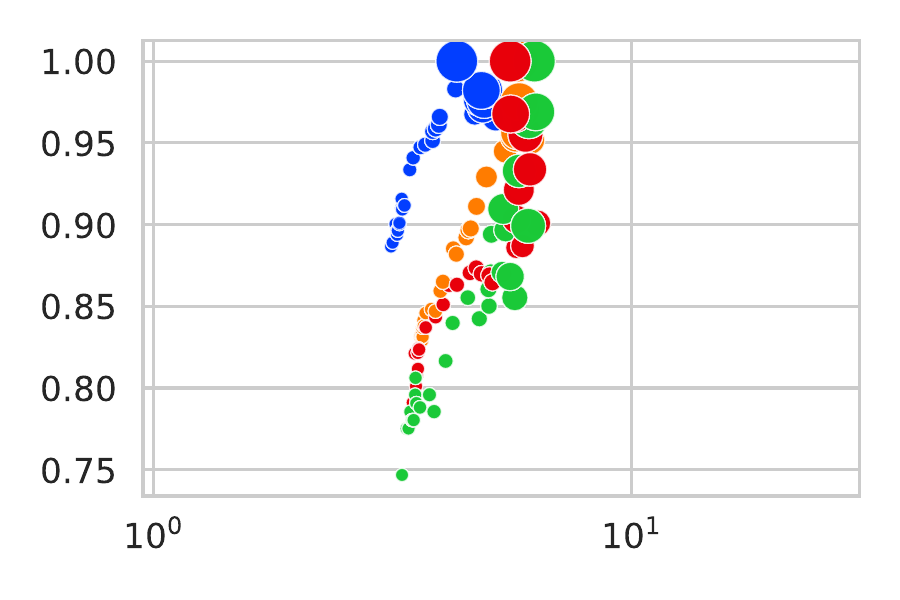}
            {\phantom{Rel. Retention Rate}}
            {Effective Dimension}
            [\textwidth]
            [1mm]
            [1mm]
        \vspace*{-3mm}
        \caption{}
        \label{fig:effDim_large-terramind-large}
    \end{subfigure}%
    \begin{subfigure}[t]{0.32\textwidth}
        \centering
        \tlabeledimage
            {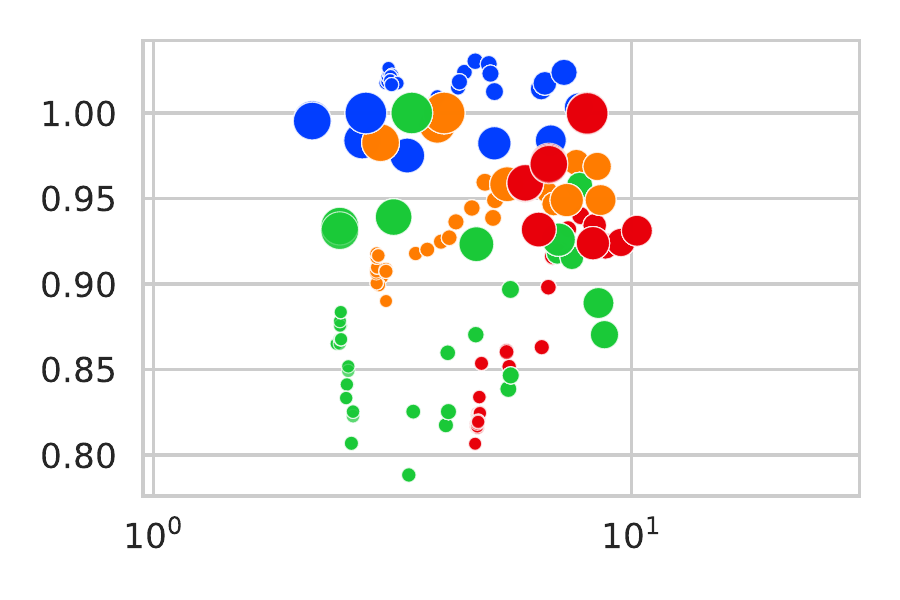}
            {\phantom{Rel. Retention Rate}}
            {\phantom{Effective Dimension}}
            [\textwidth]
        \vspace*{-3mm}
        \caption{}
        \label{fig:effDim_large-prithvi600}
    \end{subfigure}%

    \caption{Effective dimension of the test split of four classification datasets from the geobench benchmark collection with respect to their relative retention rate on pretrained \ac{RS} \acp{FM}:
    a)~DOFA (large); 
    b)~TerraMind-1.0 (large);and
    c)~Prithvi-EO-2.0 (600M).}
    \label{fig:effDim_large}
    \vspace*{-3mm}
\end{figure*}

To characterize how \ac{RS} \acp{FM} exhibit robust post-hoc slimmability and why architectures degrade differently under width reduction, we analyze three complementary representational properties: 
i)~mean feature correlation, measuring average redundancy between retained dimensions; 
ii)~the log-log slope of the \ac{EVR}, characterizing how concentrated task-relevant variance is across principal components, where steeper negative values indicate greater concentration (see Appendix \ref{sec:add_evr} for per-scale \ac{EVR} curves); and 
iii)~effective dimensionality, measuring how many dimensions contribute independent useful information at each scale.
Results for the three large models are shown in \cref{fig:corr_large,fig:loglog_slope_evr,fig:effDim_large}; results for smaller models are provided in the Appendix \ref{sec:add_evr}, \ref{sec:eff_dim}, and \ref{sec:feat_corr_model_scales} and support the same narrative to a lesser degree.
At minimum compute budget, all three large \ac{RS} \acp{FM} converge toward high feature correlation, steep \ac{EVR} slopes, and low effective dimensionality.
This is consistent with slimmed representations concentrating variance in few, highly correlated dimensions while retaining most downstream performance.
\begin{wrapfigure}{r}{0.55\linewidth}  % r = right side; 0.45 = figure width
%\begin{figure}[ht]
    \centering    
    \noindent\includegraphics[width=0.98\linewidth]{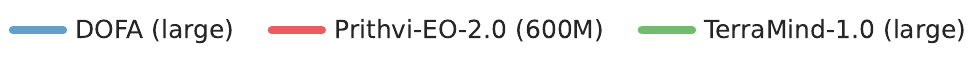}\par\vspace{-2mm}
    \tlabeledimage
        {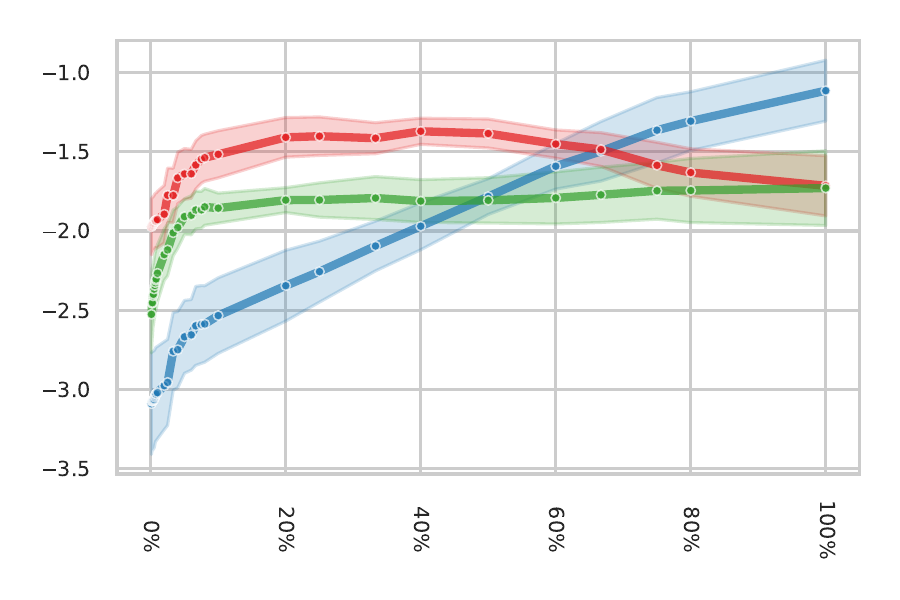}
        {Slope of the EVR}
        {Rel. Compute Requirement}
        [5.5cm]
    
    \caption{Log-Log-Slopes of the \ac{EVR} with relative compute requirements for three different \ac{RS} \acp{FM} averaged across four \ac{RS} classifications datasets.}
    \label{fig:loglog_slope_evr}
    \vspace{-8mm}
%\end{figure}
\end{wrapfigure}

\paragraph{DOFA (large).}
At minimum compute budget, DOFA (large) shows high feature correlation ($\sim$\num{0.56} on m-eurosat; \cref{fig:corr_large_dofa_large}), a steep \ac{EVR} slope ($\sim$\num{-3.1}; \cref{fig:loglog_slope_evr}), and effective dimensionality below \num{2} (\cref{fig:effDim_large-dofa-large}).
As compute budget increases, all three measures change monotonically: correlation drops to near-zero, the slope rises to $\sim$\num{-1.1}, and effective dimensionality expands to \num{34} on m-bigearthnet at full scale.
Retention and effective dimensionality are strongly correlated across scales, indicating that DOFA (large) requires many independent dimensions for full-scale performance.

\paragraph{TerraMind-1.0 (large).}
TerraMind-1.0 (large) shows a saturating rather than monotonic structure.
Feature correlation declines from moderate values ($\sim$\num{0.36} on m-eurosat) through \SI{10}{\percent} compute budget before stabilizing, then decreases gradually to full scale (\cref{fig:corr_large_terramind_large}).
The \ac{EVR} slope remains essentially flat across most of the compute range before steepening at minimum compute budget (\cref{fig:loglog_slope_evr}). 
Effective dimensionality rises from $\sim$\num{3.6} at minimum compute to $\sim$\num{5.7} by \SI{40}{\percent} compute budget, then plateaus through full scale (\cref{fig:effDim_large-terramind-large}).
On m-bigearthnet it peaks mildly at intermediate scales before declining.
The early plateau of effective dimensionality, together with the stable \ac{EVR} slope, is consistent with TerraMind-1.0 (large) retaining a stable representational profile once a moderate compute budget is reached, as observed in \cref{sec:post-hoc}.

\paragraph{Prithvi-EO-2.0 (600M).}
Prithvi-EO-2.0 (600M) exhibits a qualitatively distinct structure across all three analyses.
Feature correlation follows a non-monotonic trajectory across the compute range (\cref{fig:corr_large_prithvi_600m}), and the \ac{EVR} slope shows a corresponding non-monotonic pattern (\cref{fig:loglog_slope_evr}).
Effective dimensionality rises to a peak of $\sim$\num{8.6} on m-eurosat (and $\sim$\num{10.3} on m-bigearthnet) around \SIrange{40}{50}{\percent} compute budget, then drops sharply to $\sim$\num{3} before recovering partially toward full scale (\cref{fig:effDim_large-prithvi600}).
Retention, in contrast, remains within \SIrange{79}{103}{\percent} of full-scale across the entire compute range. 
This decoupling indicates that Prithvi-EO-2.0 (600M) encodes task-relevant information robustly across a wide range of active dimensions, including configurations where effective dimensionality differs by more than \num{2.5}$\times$ at near-identical retention.\\
Together, these analyses show that the models differ in their representational profiles under width reduction, and that no single pattern in the analyzed properties consistently tracks retention.

\newPageToggle
\section{Conclusion}
We have presented the first systematic study of post-hoc slimmability in \ac{RS} \acp{FM}, evaluating ten state-of-the-art models across classification, semantic segmentation, and change detection.
\ac{RS} \acp{FM} degrade gradually under width reduction, retaining on average over \SI{85}{\percent} relative accuracy across four classification datasets at minimum compute budget (except Panopticon, see Appendix~\ref{sec:slimdims}).
\ac{CV} \ac{MAE} and DINOv2 degrade sharply over the same range, retaining only \SIrange{30}{35}{\percent} on ImageNet subsets of matched class count.
On m-eurosat, where \ac{RS} \acp{FM} retain \SIrange{82}{97}{\percent}, a \ac{CV} \ac{MAE} evaluated on it retains \SI{76}{\percent}, narrowing but not closing the gap.
The shape of the slimming behavior, gradual for \ac{RS} \acp{FM} and steep for \ac{CV} \acp{FM}, distinguishes the two model groups clearly.
Our feature correlation, \ac{EVR} slope, and effective dimensionality analyses are consistent with redundant encoding in slimmed \ac{RS} \acp{FM}, but their relation to retention is model-dependent, and we cannot establish a cause for slimmability.
We see three plausible, non-exclusive explanations for why \ac{RS} \acp{FM} differ from \ac{CV} \acp{FM}.
First, \ac{RS} models may rely predominantly on low- and mid-level visual features that fewer independent dimensions can represent~\cite{burgert2025imagenettrained}, whereas \ac{CV} recognition draws on higher-level semantic features.
Second, \ac{RS} images are acquired under comparatively standardized viewing geometries and sensors, so fewer invariances may need to be learned than for natural images.
Third, current \ac{RS} benchmarks may not yet stress the capacity of modern \acp{FM}; richer temporal, multi-modal, or geospatial benchmarks may reveal larger benefits of scale.
Class cardinality alone does not explain the gap, since \ac{CV} models remain far less slimmable on ImageNet subsets matched in class count.
Learned slimmable training improves over post-hoc slimmability for contrastive objectives, while reconstruction-based objectives do not benefit from current protocols.

These findings have direct practical implications.
Post-hoc slimming from \SIrange{5}{20}{\percent} of the full compute budget frequently maintains over \SI{80}{\percent} relative accuracy without retraining, and intermediate-scale deployment often matches or exceeds full-scale performance.
Computational budget can therefore be treated as a continuous deployment variable.
In our experiments, the critical compute threshold tracks label-space complexity: binary m-brick-kiln tolerates the smallest budgets, 10-class m-eurosat remains effective at about \SI{1}{\percent} compute, and 17-class m-so2sat and 43-label m-bigearthnet require roughly \SIrange{2}{10}{\percent}.
%This gives a starting point for choosing the deployment width, which we recommend to confirm by 5-fold cross-validation on the labeled training data, since we have no principled method to predict the optimal width.

\paragraph{Limitations and Future Work.}
Our findings characterize post-hoc width redundancy in the evaluated checkpoints, not a general scaling law for \ac{RS} \acp{FM}.
Relative retention must be interpreted alongside absolute performance, which differs substantially across models and datasets.
Our protocol slims only the \ac{FFN} and \ac{MHSA} layers of the encoder blocks and leaves the patch embedding unchanged, so the attainable compute reduction depends on their share of the computation.
For example, in Panopticon the embedding accounts for more than \SI{60}{\percent} of the relative compute.
Our main results use \ac{KNN}, but linear probing and fine-tuning yield the same trends (Appendix~\ref{sec:linear_results}).
Retention decreases under geographic distribution shift, although substantial redundancy remains, and aggregate metrics can mask class-wise degradation (Appendix~\ref{sec:failure}).
Our study is limited to \ac{ViT}-based \ac{RS} \acp{FM} evaluated on optical Sentinel-2 imagery.
As future work, we plan to cover other modalities such as \acl{SAR}, non-\ac{ViT} backbones (convolutional, hybrid, state-space), and generative \ac{RS} models such as DiffusionSat, for which retention would have to be assessed through generation quality or learned representations rather than downstream accuracy.
Finally, learned slimmable training under reconstruction-based objectives remains an open problem, as existing protocols are designed for contrastive pretraining and do not suit \ac{MAE}-style objectives.
Stabilizing it would extend the benefits of learned slimmability to a broader range of \ac{RS} \acp{FM}.

\section*{Impact Statement}
This paper presents work whose goal is to advance the field of Machine Learning.
Post-hoc slimming reduces the computational cost of deploying remote sensing 
foundation models, which could lower barriers to large-scale Earth observation 
in sensitive contexts such as surveillance. We encourage responsible use in 
accordance with applicable legal and ethical frameworks.

% Acknowledgements can already be on page 11 as per "Details about camera ready submission" email
\section*{Acknowledgements}
This work is supported by the European Research Council Project Agent-BigEarth (No. 101292498) and the European Union Horizon Project PROTEUS (No. 101296397).

\newPageToggle

%\section*{References}
\bibliographystyle{plainnat}
\bibliography{sections/99a-references}

%%%%%%%%%%%%%%%%%%%%%%%%%%%%%%%%%%%%%%%%%%%%%%%%%%%%%%%%%%%%
\newpage
\appendix

\section*{Technical appendices and supplementary material}
\subsection*{Appendix Overview}
\begin{enumerate}
  \setlength{\itemsep}{0pt}
  \setlength{\parskip}{0pt}

    \item[\ref{sec:slimdims}] \hyperref[sec:slimdims]{Details on Layer Slimming}
    \item[\ref{sec:ffn_vs_attn}] \hyperref[sec:ffn_vs_attn]{Slimmability of Different Layer Types}
    \item[\ref{sec:implicit_order}] \hyperref[sec:implicit_order]{Implicit Dimension Ordering}
    \item[\ref{sec:training_from_scratch}] \hyperref[sec:training_from_scratch]{Details on Training from Scratch}
    \item[\ref{sec:protocol}] \hyperref[sec:protocol]{Details on the Evaluation Protocol}
    \item[\ref{sec:linear_results}] \hyperref[sec:linear_results]{Linear Evaluation and Finetuning}
    \item[\ref{sec:eff_dim}] \hyperref[sec:eff_dim]{Effective Dimensionality Analysis}
    \item[\ref{sec:add_post-hoc}] \hyperref[sec:add_post-hoc]{Additional Experiments on Post-Hoc Slimmability}
    \item[\ref{sec:add_learned_v_post-hoc}] \hyperref[sec:add_learned_v_post-hoc]{Additional Experiments on Learned vs Post-Hoc }Slimmability
    \item[\ref{sec:feat_corr_model_scales}] \hyperref[sec:feat_corr_model_scales]{Feature Correlation at different Model Scales}
    \item[\ref{sec:add_evr}] \hyperref[sec:add_evr]{Additional Experiments on Explained Variance}
    \item[\ref{sec:failure}] \hyperref[sec:failure]{Failure Modes and Edge Cases}
    \item[\ref{sec:compute}] \hyperref[sec:compute]{Computational Resources}
    \item[\ref{sec:licenses}] \hyperref[sec:licenses]{Licenses and Assets}
    
\end{enumerate}
\FloatBarrier

\section{Details on Layer Slimming}\label{sec:slimdims}

\paragraph{Slimmed dimensions.}
Post-hoc slimming reduces only the \ac{FFN} hidden dimension and the per-head dimensions of the \ac{MHSA} layers in each encoder transformer block. 
The transformer embedding dimension is never changed, nor is the patch embedding layer. 
Consequently, every normalization layer operates on the same feature dimension at all widths and requires neither recalibration nor switchable parameters, unlike batch normalization in learned slimmable networks of other works (e.g., \cite{yu2019slimmable, yu2019universally, li2021dynamic, Cao_2023_CVPR, zhao2025slimmable}.

\paragraph{Panopticon.}
For all other models, the patch embedding accounts for only about \SI{5}{\percent} of the parameters, so leaving it unslimmed has little influence on the reported retention.
Panopticon is an exception: its patch embedding accounts for more than \SI{66}{\percent} of the parameters.
Since our protocol slims only the \ac{MHSA} and \ac{FFN}, the patch embedding layer is never reduced.
The figures plot total relative compute rather than the slimming scale, i.e., the width fraction applied to the two layers. 
For most models, both axes nearly coincide as their patch embeddings are small compared to the rest of the model.
For Panopticon, even the smallest scale corresponds to a relative compute of more than \SI{60}{\percent} (e.g., \cref{fig:post-hoc}), so its curves cannot reach the low-compute regime covered by the other models.
We keep the same protocol to remain comparable, as slimming the patch embedding for a single model would break comparability.
Panopticon is therefore excluded from the main comparisons that plot retention against relative compute.
\section{Slimmability of Different Layer Types}\label{sec:ffn_vs_attn}
\begin{figure*}
    \centering
    \noindent\includegraphics[width=0.98\linewidth]{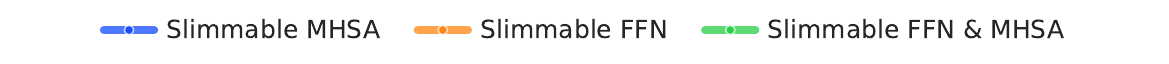}\par\vspace{-2mm}
    % Row 1
    %\rowgrouplabel[10mm]{m-eurosat}\hspace{1mm}
    \begin{subfigure}[t]{0.3\textwidth}
        \centering
        \tlabeledimage
            {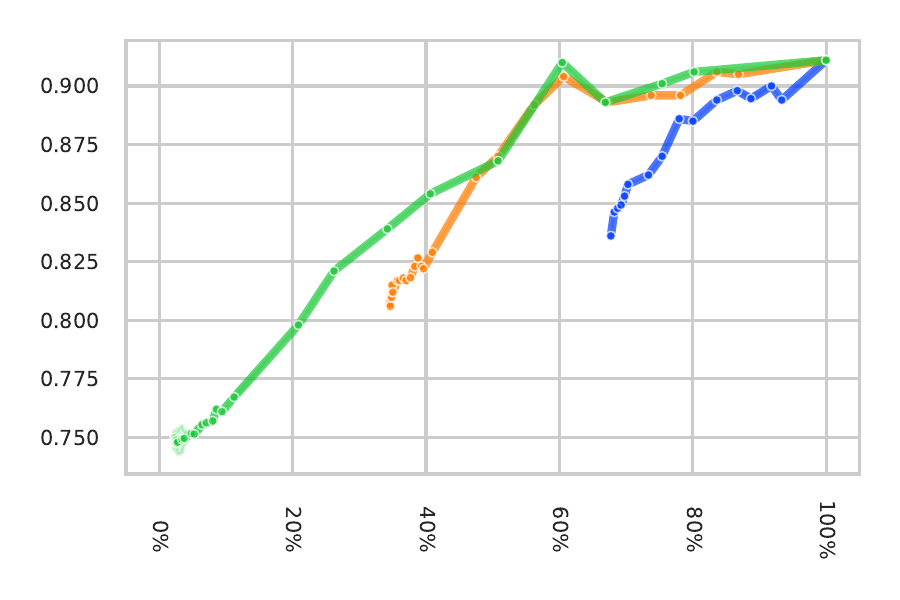}
            {Accuracy}
            {\phantom{Rel. Compute Requirement}}
        \vspace*{-6mm}
        \caption{}
        \label{fig:ffn_v_attn_dofa_euro}
    \end{subfigure}%
    \begin{subfigure}[t]{0.3\textwidth}
        \centering
        \tlabeledimage
            {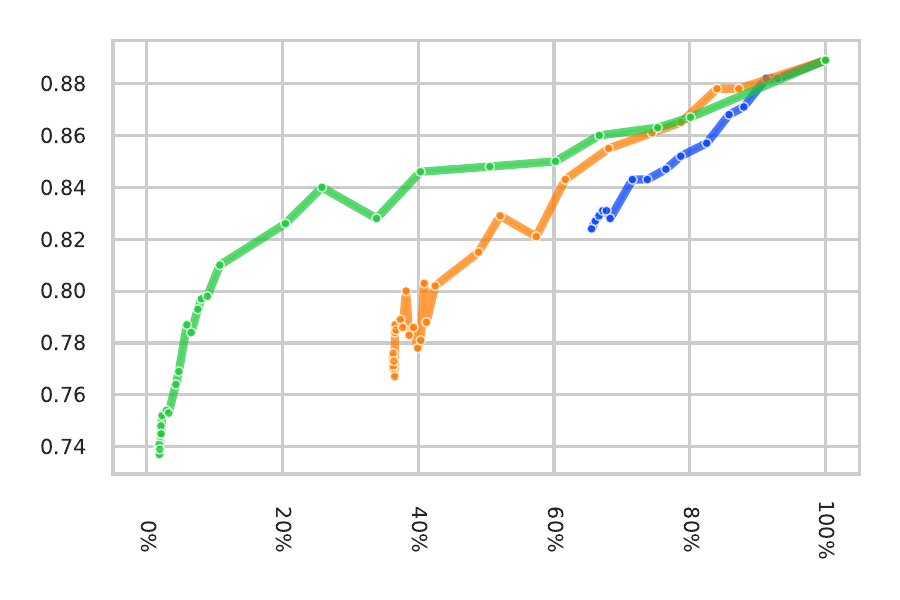}
            {\phantom{Accuracy}}
            {\phantom{Rel. Compute Requirement}}
        \vspace*{-6mm}
        \caption{}
        \label{fig:ffn_v_attn_terramind_euro}
    \end{subfigure}%
    \begin{subfigure}[t]{0.3\textwidth}
        \centering
        \tlabeledimage
            {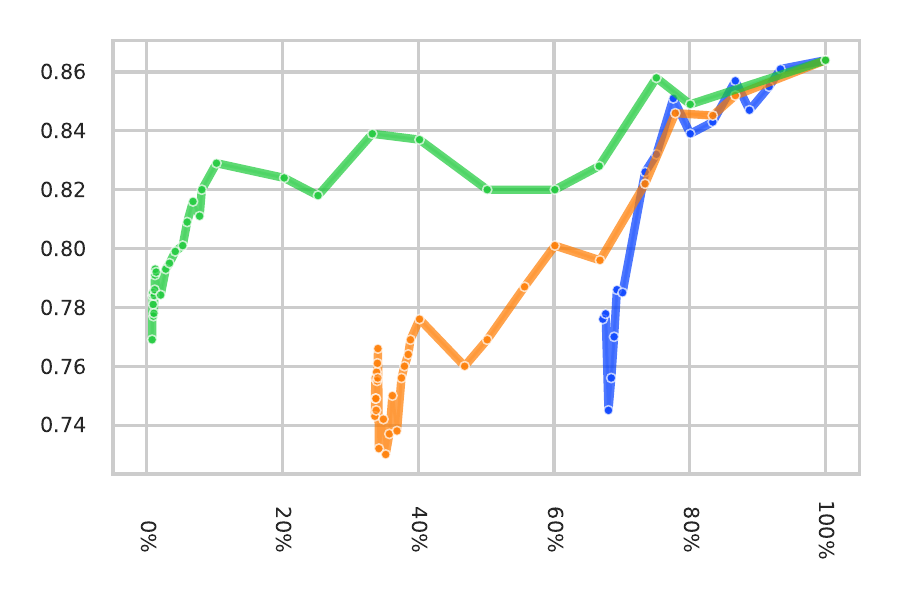}
            {\phantom{Accuracy}}
            {\phantom{Rel. Compute Requirement}}
        \vspace*{-6mm}
        \caption{}
        \label{fig:ffn_v_attn_prithvi_euro}
    \end{subfigure}%

    % Row 2
    %\rowgrouplabel[6mm]{m-bigearthnet}\hspace{2mm}
    \begin{subfigure}[t]{0.3\textwidth}
        \centering
        \tlabeledimage
            {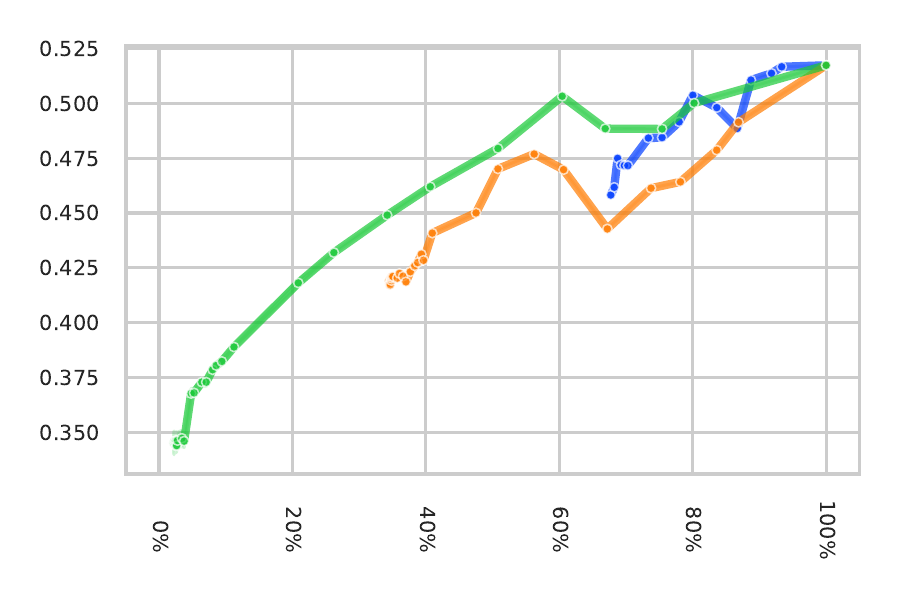}
            {\ac{mAP}}
            {\phantom{Rel. Compute Requirement}}
        \vspace*{-3mm}
        \caption{}
        \label{fig:ffn_v_attn_dofa_ben}
    \end{subfigure}%
    \begin{subfigure}[t]{0.3\textwidth}
        \centering
        \tlabeledimage
            {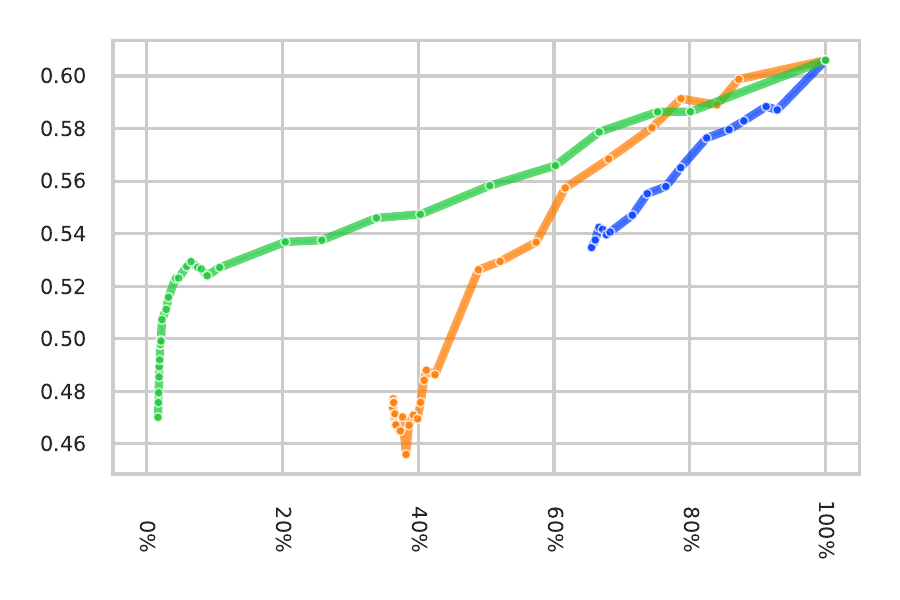}
            {\phantom{\ac{mAP}}}
            {Rel. Compute Requirement}
        \vspace*{-3mm}
        \caption{}
        \label{fig:ffn_v_attn_terramind_ben}
    \end{subfigure}%
    \begin{subfigure}[t]{0.3\textwidth}
        \centering
        \tlabeledimage
            {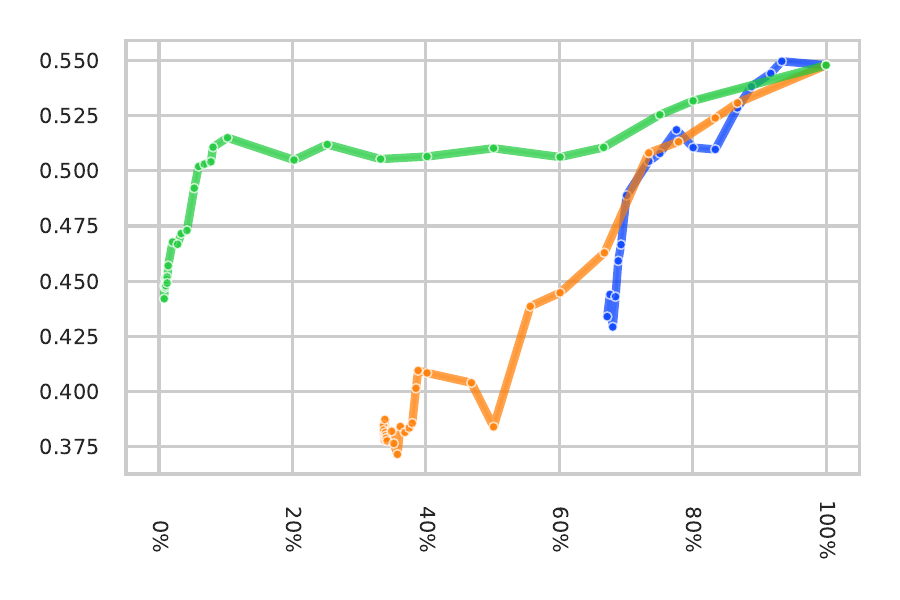}
            {\phantom{\ac{mAP}}}
            {\phantom{Rel. Compute Requirement}}
        \vspace*{-3mm}
        \caption{}
        \label{fig:ffn_v_attn_prithvi_ben}
    \end{subfigure}
    
    \caption{\ac{KNN} accuracy and \ac{mAP} with respect to their relative compute requirements for slimming only the \ac{FFN} layer, only the \ac{MHSA} layer, or both layers simultaneously of:
    a)~DOFA (large); 
    b)~TerraMind-1.0 (large); and
    c)~Prithvi-EO-2.0 (600M) on m-eurosat; and 
    d)~DOFA (large); 
    e)~TerraMind-1.0 (large); and
    f)~Prithvi-EO-2.0 (600M) on m-bigearthnet.}
    \label{fig:ffn_v_attn}
\end{figure*}

To understand the computational contributions of different architectural components, we compare three slimming strategies: 
i) slimming only \ac{FFN} layers while keeping \ac{MHSA} at full width; 
ii) slimming only \ac{MHSA} layers while keeping \ac{FFN} at full width; and 
iii) slimming both layer types simultaneously. 
The results of these experiments are shown in \cref{fig:ffn_v_attn}.

As one can see from the figure, our analysis reveals that \ac{MHSA} and \ac{FFN} layers contribute asymmetrically to both computational cost and representational capacity. 
Models slimmed exclusively through \ac{MHSA} width reduction (blue lines) cannot achieve relative compute requirements below \SI{60}{\percent}, as the fixed-width \ac{FFN} layers dominate the computational budget. 
Moreover, \ac{MHSA}-only slimming exhibits steeper performance degradation relative to compute reduction compared to \ac{FFN}-only approaches, suggesting that attention mechanisms encode more task-critical information per compute. 
In contrast, \ac{FFN}-only slimming (orange lines) enables compression down to \SI{35}{\percent} relative compute while maintaining more gradual performance decay.
This indicates that \ac{FFN} layers contribute disproportionately to \acp{FLOP} but contain more redundant parameters that can be removed with minimal performance impact.

However, simultaneous slimming of both layer types (green lines) proves superior across all evaluated models, enabling compression to \SI{1}{\percent} relative compute while exhibiting the most gradual performance degradation curves. 
This synergistic effect suggests that maintaining architectural balance, where \ac{MHSA} and \ac{FFN} widths scale proportionally, preserves the learned feature processing pipeline more effectively than asymmetric compression. 
The ability to slim both components simultaneously also provides access to the full computational range, from \num{1}--\SI{100}{\percent} of original \acp{FLOP}, enabling fine-grained deployment optimization. 
From a practical deployment perspective, practitioners should prioritize simultaneous slimming of both layer types to maximize compression ratios while maintaining performance, rather than attempting to preserve specific architectural components at full capacity.
\section{Implicit Dimension Ordering}\label{sec:implicit_order}

\begin{figure}[ht]
    \centering
    \noindent\includegraphics[width=0.98\linewidth]{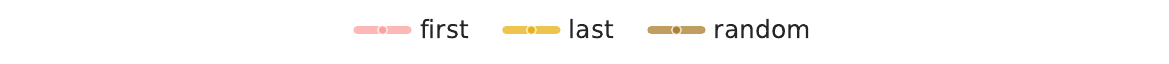}\par\vspace{-2mm}
    \begin{subfigure}[t]{0.3\textwidth}
        \centering
        \tlabeledimage
            {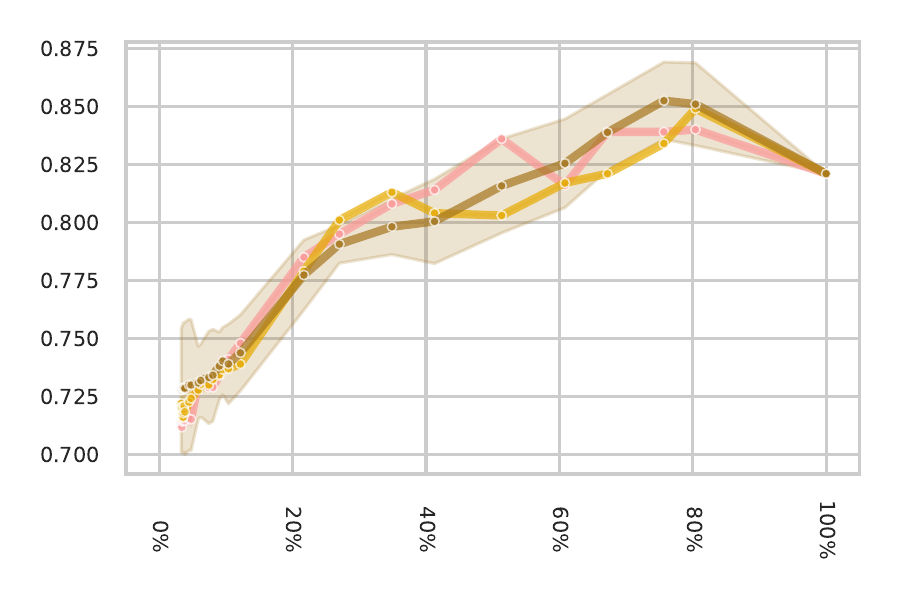}
            {Accuracy}
            {\phantom{Rel. Compute Requirement}}
        \vspace*{-3mm}
        \caption{}
        \label{fig:impl_order_dofa}
    \end{subfigure}%
    \begin{subfigure}[t]{0.3\textwidth}
        \centering
        \tlabeledimage
            {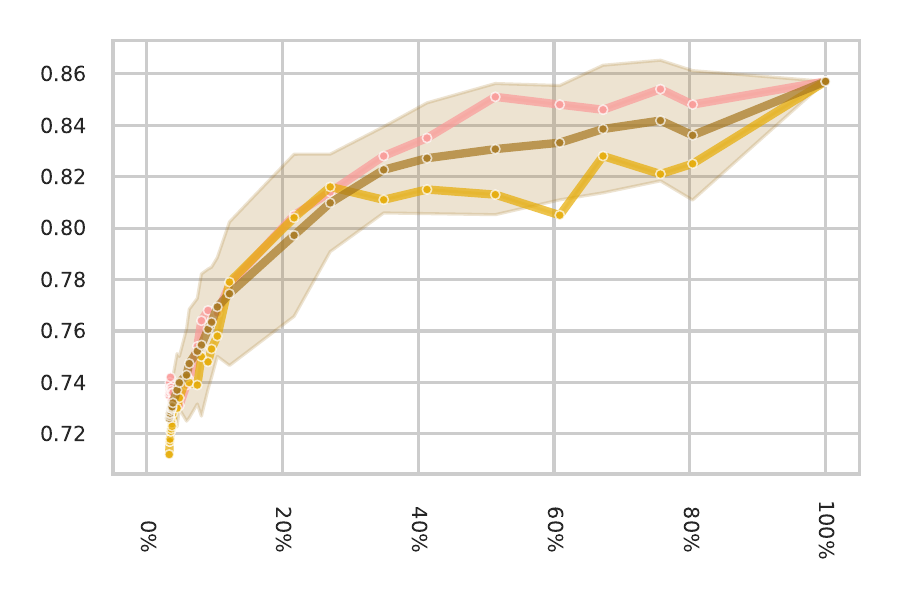}
            {\phantom{Accuracy}}
            {Rel. Compute Requirement}
        \vspace*{-3mm}
        \caption{}
        \label{fig:impl_order_terramind}
    \end{subfigure}%
    \begin{subfigure}[t]{0.3\textwidth}
        \centering
        \tlabeledimage
            {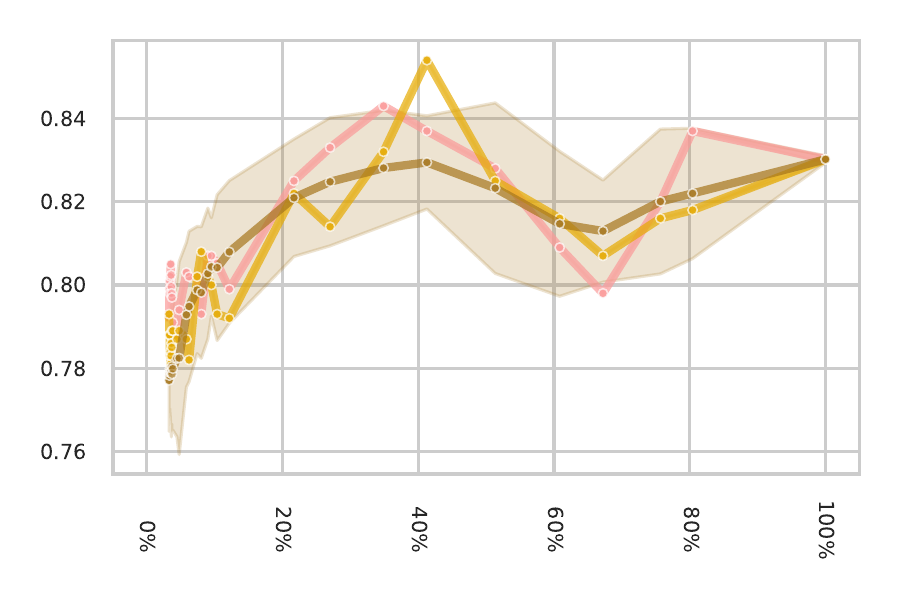}
            {\phantom{Accuracy}}
            {\phantom{Rel. Compute Requirement}}
        \vspace*{-3mm}
        \caption{}
        \label{fig:impl_order_prithvi}
    \end{subfigure}%
    \caption{KNN classification performance with respect to relative compute requirements on 
    the test split of m-eurosat when selecting the first, last, or a random subset of neurons 
    for slimming. Random order experiments are repeated 20 times:
    a)~DOFA (base); 
    b)~TerraMind-1.0 (base); and
    c)~Prithvi-EO-2.0 (300M).}
    \label{fig:impl_order}
    \vspace{-3mm}
\end{figure}

\cref{fig:impl_order} shows \ac{KNN} classification performance on m-eurosat for pretrained \ac{RS} \acp{FM} when slimming retains the first $d'$ dimensions, the last $d'$ dimensions, or a randomly selected subset of $d'$ dimensions, where random selection was repeated 20 times.
Error bars denote 2-sigma intervals, consistent with the visualization convention used throughout this paper.
For TerraMind-1.0 (base), first retention lies marginally above and last retention marginally below the 2-sigma interval of random at compute scales below \SI{1}{\percent}, with all three conditions converging above this threshold.
For DOFA (base), first and last retention fall within the 2-sigma interval of random retention across all compute scales, indicating that dimension order has no detectable effect beyond the variability of random subset selection.
For Prithvi-EO-2.0 (300M), first and last retention are nearly identical across all compute scales and both lie slightly above the random 2-sigma interval only at the lowest compute scales, with no detectable ordering between them.
Across all three models, the deviations from random at very low compute are inconsistent in structure: for TerraMind, first and last bracket the random interval from opposite sides, whereas for Prithvi, first and last deviate jointly in the same direction.
This cross-model inconsistency rules out a shared implicit dimension ordering in the pretrained weights as an explanation, and confirms that the deviations reflect model-specific noise rather than a structural property of the slicing convention.
The findings reported in \cref{sec:post-hoc} are therefore not an artifact of the first-dimension slicing convention.
\section{Details on Training from Scratch}\label{sec:training_from_scratch}
To investigate learned slimmability (Contribution C2), we train both \ac{MoCo} and \ac{MAE} architectures with and without slimmable training on a curated \ac{RS} dataset. 
We use the dataset from~\citet{hackel2025csmoe}, which applies entropy-maximizing stratified sampling to Major Tom Core (MTC)~\cite{francis2024major} based on Köppen-Geiger climate zones and ESA WorldCover classes, yielding approximately 100 spatially diverse samples per joined stratum and a total of $\sim \SI{1}{\mega\nothing}$ training patches at $120\times120$ pixels. 
All models use a \ac{ViT}-Base backbone with 10 input channels (\SI{10}{\meter} and \SI{20}{\meter} Sentinel-2 bands) and image size $224\times224$ pixels.

\textbf{Training Protocol.} Regular (non-slimmable) training is conducted for 300 epochs with batch size of 384 per GPU and AdamW optimizer ($\beta_1=0.9$, $\beta_2=0.95$, weight decay 0.05). 
The learning rate follows a linear warmup for $1\%$ of total steps, followed by cosine annealing from base learning rate $\text{lr}_{\text{base}} \times \frac{\text{batch size}}{256}$ to $\frac{\text{lr}_{\text{base}}}{100}$, where $\text{lr}_{\text{base}}=10^{-4}$.
For \ac{MoCo}, we use momentum encoder update with cosine schedule from 0.996 to 1.0 over 10 epochs, projection head dimension 128, and \ac{NTXent} loss with memory bank size 4096. 
For \ac{MAE}, we use mask ratio 0.75, decoder dimension 512, decoder depth 1, and 16 decoder heads with \ac{MSE} reconstruction loss.
All pretraining experiments were run on 4 NVIDIA A100-80GB GPUs or 4 NVIDIA H200-141GB GPUs.

\textbf{Slimmable Pretraining.} Learned slimmable training extends the regular protocol to 800 epochs to account for the increased optimization difficulty. 
During each training step, we perform manual optimization with multiple forward-backward passes: 
i) the full network at scale $s=1.0$ with regular task loss; 
ii) the minimal network at scale $s_{\text{min}}$ with task loss plus knowledge distillation from the full network;
and iii) three randomly sampled intermediate scales with the same combined loss. 
The minimal scale $s_{\text{min}}$ linearly decreases from 0.5 to 0.01 over the first \SI{50}{\percent} of training to encourage robust feature learning across scales following \cite{zhao2025slimmable}. 
Knowledge distillation uses temperature-scaled KL divergence with $T=1.0$ between features passed through a two-layer MLP distillation head ($d \rightarrow 2d \rightarrow d$ with GELU activation). 
This multi-scale training encourages weight sharing and robustness across the entire range of network widths.

\begin{figure*}
    \centering
    \noindent\includegraphics[width=0.98\linewidth]{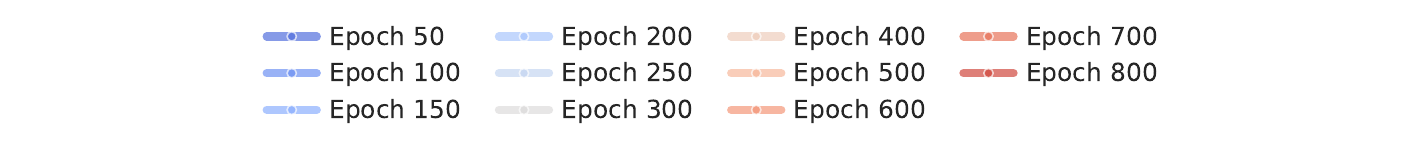}\par\vspace{-2mm}
    % Column headers
    \hspace{8mm}% offset for row group label width
    %\makebox[0.40\textwidth]{\textsf{\textit{\underline{\textbf{Regular Training}}}}}%
    %\makebox[0.40\textwidth]{\textsf{\textit{\underline{\textbf{Slimmable Training}}}}}

    % Row 1: MAE
    %\rowgrouplabel[14.5mm]{MAE}%
    \begin{subfigure}[t]{0.40\textwidth}
        \centering
        \tlabeledimage
            {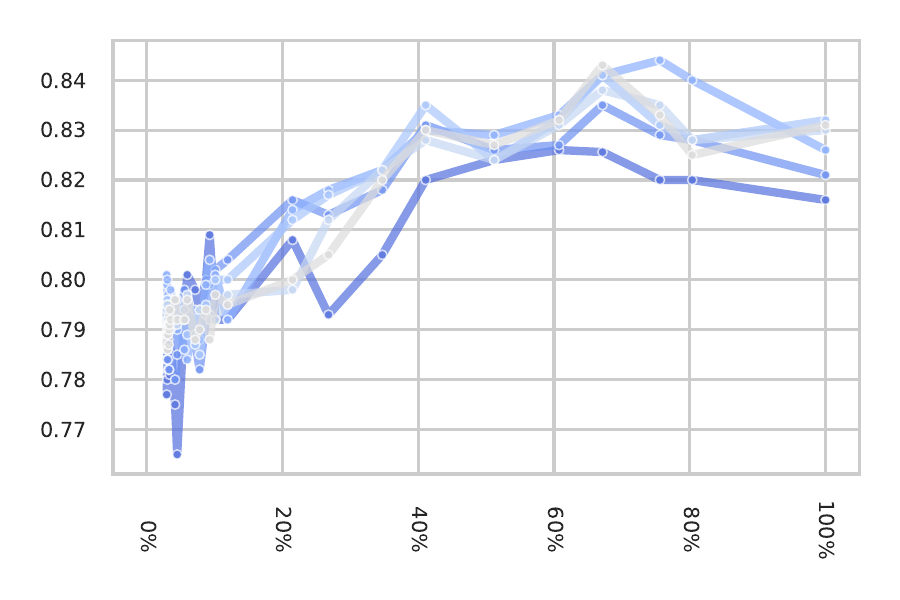}
            {Accuracy}
            {\phantom{Rel. Compute Requirement}}
        \vspace*{-6mm}
        \caption{}
        \label{fig:training_MAE}
    \end{subfigure}%
    \begin{subfigure}[t]{0.40\textwidth}
        \centering
        \tlabeledimage
            {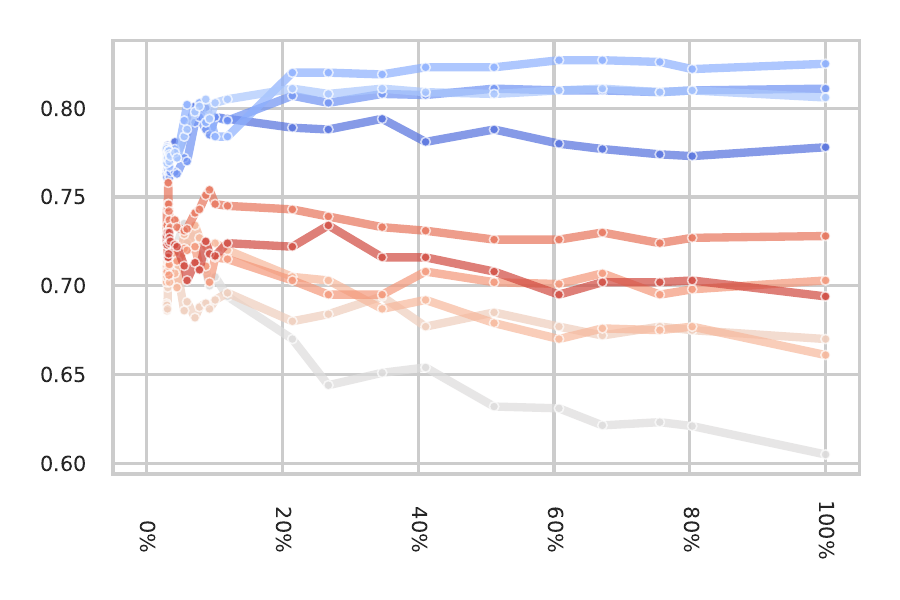}
            {\phantom{Accuracy}}
            {\phantom{Rel. Compute Requirement}}
        \vspace*{-6mm}
        \caption{}
        \label{fig:training_MAE_slim}
    \end{subfigure}%

    % Row 2: MoCo
    %\rowgrouplabel[13mm]{MoCo}%
    \begin{subfigure}[t]{0.40\textwidth}
        \centering
        \tlabeledimage
            {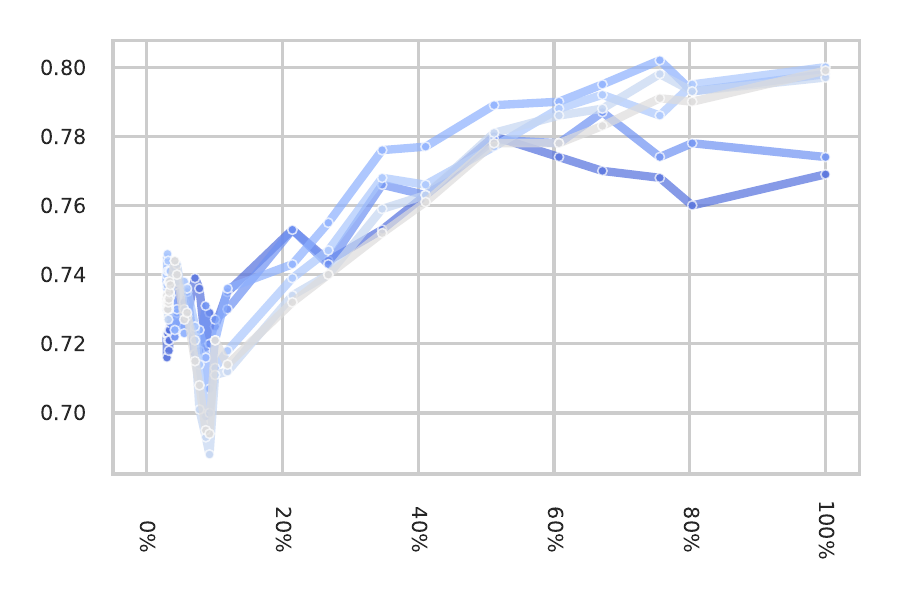}
            {Accuracy}
            {Rel. Compute Requirement}
        \vspace*{-3mm}
        \caption{}
        \label{fig:training_MoCo}
    \end{subfigure}%
    \begin{subfigure}[t]{0.40\textwidth}
        \centering
        \tlabeledimage
            {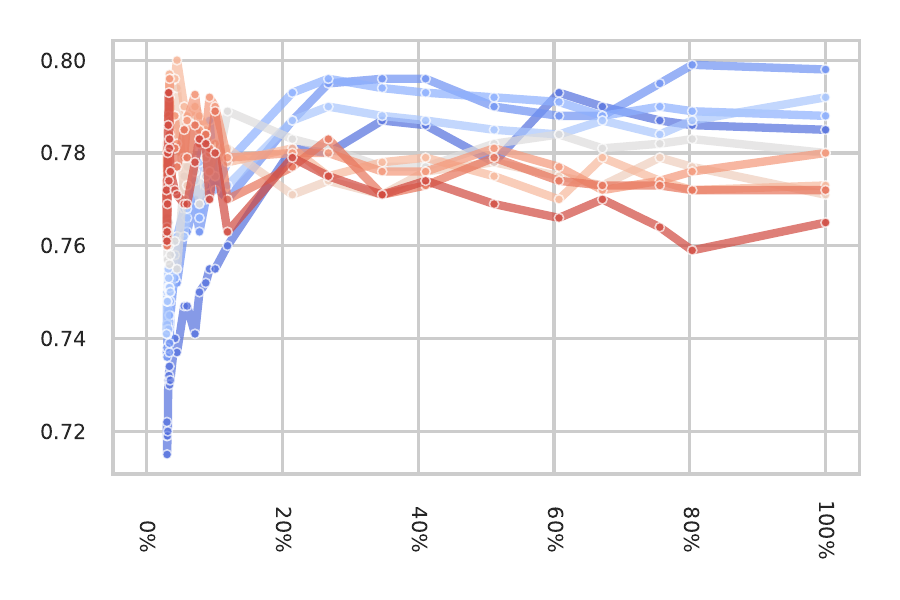}
            {\phantom{Accuracy}}
            {Rel. Compute Requirement}
        \vspace*{-3mm}
        \caption{}
        \label{fig:training_MoCo_slim}
    \end{subfigure}%

    \caption{\ac{KNN} classification performance with respect to relative compute requirements of a \ac{MAE}-based and a \ac{MoCo}-based trained model from scratch, each once with and once without slimmable training on m-eurosat for different training durations: a)~regular \ac{MAE}; b)~slimmable \ac{MAE}; c)~regular \ac{MoCo}; d)~slimmable \ac{MoCo}.}
    \label{fig:training_mae_moco_epochs}
\end{figure*}

\textbf{Evaluation.} We evaluate all resulting models at multiple intermediate checkpoints (every 50 epochs for the first 300 epochs, then every 100 epochs) on the m-eurosat dataset from the geobench benchmark collections using our \ac{KNN} evaluation protocol.
The results of this evaluation can be seen in \cref{fig:training_mae_moco_epochs}.
As one can see from the figures, for regular training the accuracy of the full network increases with additional training for both \ac{MAE} and \ac{MoCo} architectures (\cref{fig:training_MAE} and \ref{fig:training_MoCo}) although only small or marginal improvements can be seen after 100 epochs.
For smaller scales (with respective lower compute requirements), the accuracy is not monotonously increasing with longer training.

For slimmable training, in the case of \ac{MAE} (\cref{fig:training_MAE_slim}) the accuracy increases across all scales for the first 200 epochs, then sharply declines by almost \SI{20}{\percent} and then slowly increases over the following 600 epochs but never fully recovers.
The slimmably trained \ac{MoCo} model decreases accuracy for higher compute requirements (larger than \SI{20}{\percent}) but increases for smaller ones. 
Overall the accuracy equalizes for this model across all compute requirements with longer training.
\section{Details on the Evaluation Protocol}\label{sec:protocol}
\subsection{Scale Evaluation}\label{subsec:scale_details}
During \ac{KNN} and linear evaluation, we evaluate each model on 31 scales: \{0.001, 0.002, 0.0025, 0.00333, 0.004, 0.005, 0.006, 0.00667, 0.0075, 0.008, 0.01, 0.02, 0.025, 0.0333, 0.04, 0.05, 0.06, 0.0667, 0.075, 0.08, 0.1, 0.2, 0.25, 0.333, 0.4, 0.5, 0.6, 0.667, 0.75, 0.8, 1.0\}. The reduced dimensions $d_k' = \lfloor s \cdot d_k \rfloor$ and $d_h' = \lfloor s \cdot d_h \rfloor$ are at minimum 1, meaning multiple scales can result in the same reduced dimension for small values of $s$. For finetuning only the following five scales are evaluated due to computational constraints: \{0.001, 0.01, 0.1, 0.5, 1.0\}.

\subsection{Feature Extraction and Downstream Tasks}\label{subsec:linear_probing_protocol}
\subsubsection{Classification}
\textbf{Feature Extraction.} For all models and scales, we extract features from the penultimate layer (before the final classification head or pooling operation) on both training and test sets of each downstream dataset. 
Features are saved as NumPy arrays along with their corresponding labels to enable efficient repeated evaluation without recomputing features.

\textbf{\ac{KNN} Evaluation.} We use $k=5$ neighbors and temperature parameter $t=0.9$ for the weighted KNN classifier. Classification is performed using cosine similarity between test features and the training feature bank. 
For the binary dataset (m-brick-kiln) and multi-class datasets (m-eurosat, m-so2sat), we compute macro-averaged accuracy. For the multi-label m-bigearthnet dataset, we report the \ac{mAP}.

\textbf{Linear Probing.} Linear classifiers are trained for 50 epochs using AdamW optimizer with learning rate 0.05, batch size 64, and no weight decay. For multi-class tasks, we use cross-entropy loss; for multi-label m-bigearthnet, we use binary cross-entropy with logits. We report the same metrics as for \ac{KNN} evaluation.

\textbf{Finetuning.} We finetune \acp{FM} for 40 epochs using AdamW optimizer with learning rate 3e-3. All other parameters are the same as for linear probing.

\subsubsection{Dense Prediction}
\textbf{Semantic Segmentation.} Following common practice for dense prediction with frozen \ac{ViT} backbones, we extract multi-scale features from four equally spaced encoder transformer blocks rather than the penultimate layer alone (indices $\mathrm{round}(i \cdot (n-1)/3)$ for $i \in \{0,1,2,3\}$, where $n$ is the total number of blocks). 
Class tokens are removed and the resulting tokens are reshaped into spatial feature maps of shape $(B, 4, D, H_p, W_p)$, where $H_p = W_p = 224/p$ and $p$ is the patch size of the respective backbone. 
Features are L2-normalized along the feature dimension. 
We train a SegFormer-style \ac{MLP} head that projects each of the four feature maps to a shared embedding dimension of 256, fuses them by concatenation followed by a linear projection, classifies per patch, and bilinearly upsamples to the input resolution of $224 \times 224$. 
Training uses AdamW with learning rate 0.01, weight decay 0.01, cosine annealing, batch size 32, and 30 epochs, with cross-entropy loss ignoring index 255. 
We report the \ac{IoU} on the test set.

\textbf{Change Detection.} We extract features from both timestamps independently using the same four-block selection and reshaping procedure as for semantic segmentation, yielding tensors of shape $(N, 2, 4, D, H_p, W_p)$. 
Per-pixel feature differences $f_{t_2} - f_{t_1}$ are computed along the timestamp dimension, producing inputs of shape $(N, 4, D, H_p, W_p)$ that match the semantic segmentation input format. 
We apply a head with the same architecture as for semantic segmentation but with embedding dimension 128 and binary output. 
Training uses the same protocol as for semantic segmentation, with weight decay 0.1 and class weights $[1, n_{\text{neg}}/n_{\text{pos}}]$ in the cross-entropy loss to address the strong class imbalance of OSCD. We report the \ac{IoU} averaged over the change and no-change classes on the test set.

\subsubsection{Experimental Repeatability} 
Linear probing experiment is performed five times with different seeds and averaged results are reported to account for variability in both feature extraction (at different scales) and downstream evaluation.
\ac{KNN} experiments are deterministic and don't require different seeds.
Finetuning experments are repeated three times with different seeds.
All downstream evaluations were run on single NVIDIA A40-48GB GPUs.
\section{Linear Evaluation and Finetuning}\label{sec:linear_results}
\begin{figure*}
    \centering    
    \noindent\includegraphics[width=0.98\linewidth]{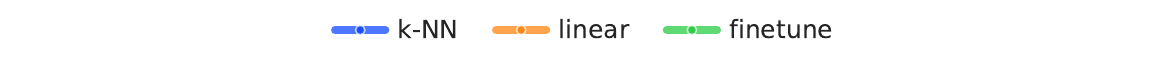}\par\vspace{-2mm}
    % Row 1
    %\rowgrouplabel[10mm]{m-eurosat}\hspace{1mm}
    \begin{subfigure}[t]{0.3\textwidth}
        \centering
        \tlabeledimage
            {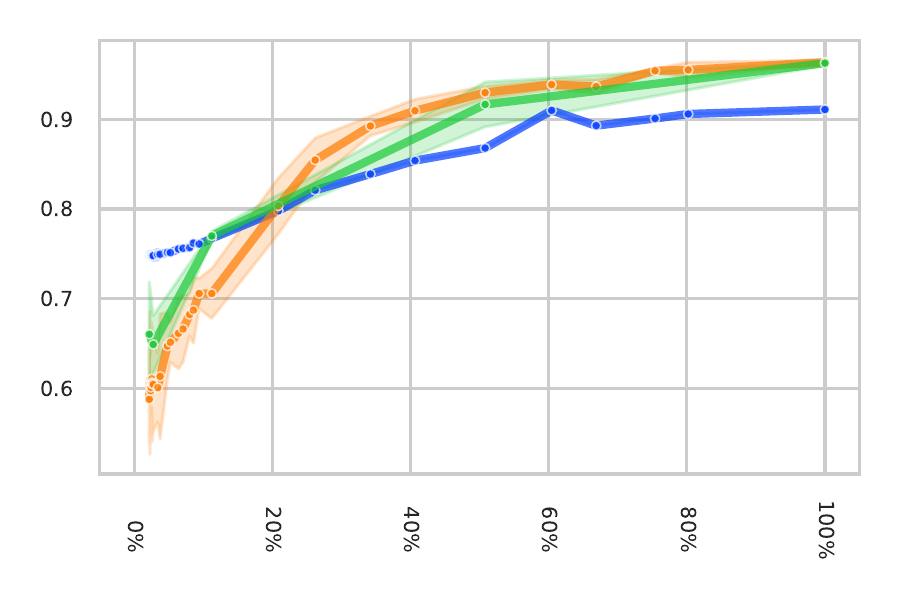}
            {Accuracy}
            {\phantom{Rel. Compute Requirement}}
        \vspace*{-6mm}
        \caption{}
        \label{fig:knn_v_lin_dofa_euro}
    \end{subfigure}%
    \begin{subfigure}[t]{0.3\textwidth}
        \centering
        \tlabeledimage
            {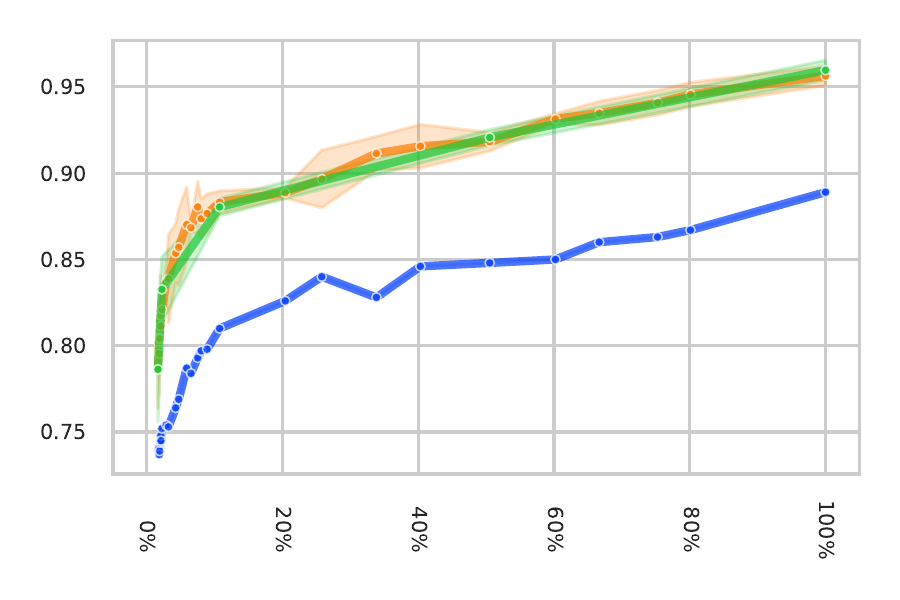}
            {\phantom{Accuracy}}
            {\phantom{Rel. Compute Requirement}}
        \vspace*{-6mm}
        \caption{}
        \label{fig:knn_v_lin_terramind_euro}
    \end{subfigure}%
    \begin{subfigure}[t]{0.3\textwidth}
        \centering
        \tlabeledimage
            {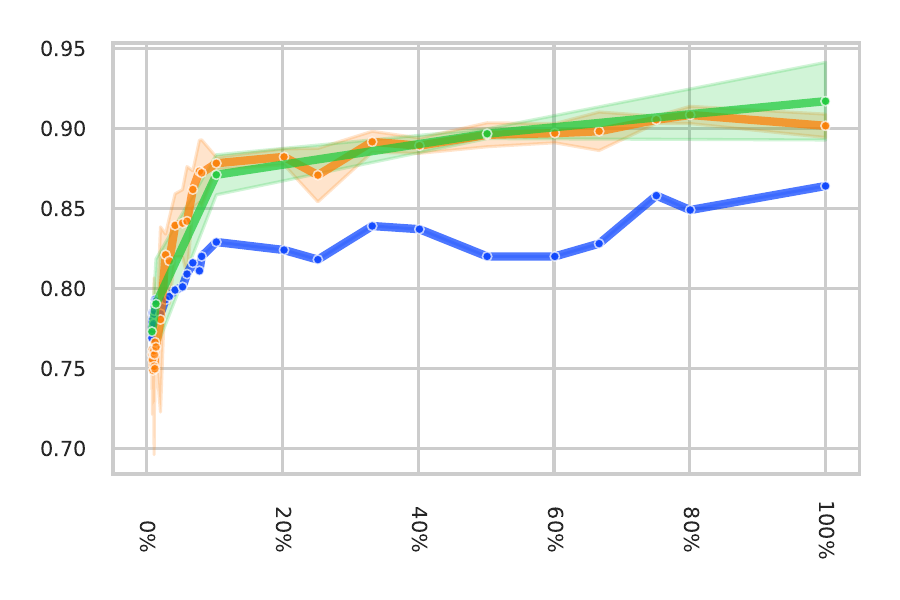}
            {\phantom{Accuracy}}
            {\phantom{Rel. Compute Requirement}}
        \vspace*{-6mm}
        \caption{}
        \label{fig:knn_v_lin_prithvi_euro}
    \end{subfigure}%
    
    % Row 2
    %\rowgrouplabel[6mm]{m-bigearthnet}\hspace{2mm}
    \begin{subfigure}[t]{0.3\textwidth}
        \centering
        \tlabeledimage
            {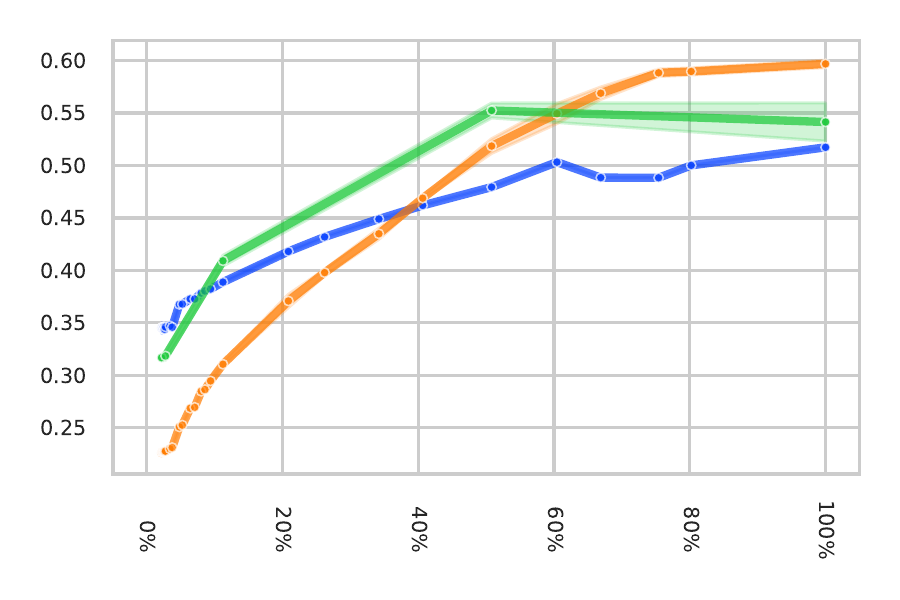}
            {\ac{mAP}}
            {\phantom{Rel. Compute Requirement}}
        \vspace*{-3mm}
        \caption{}
        \label{fig:knn_v_lin_dofa_ben}
    \end{subfigure}%
    \begin{subfigure}[t]{0.3\textwidth}
        \centering
        \tlabeledimage
            {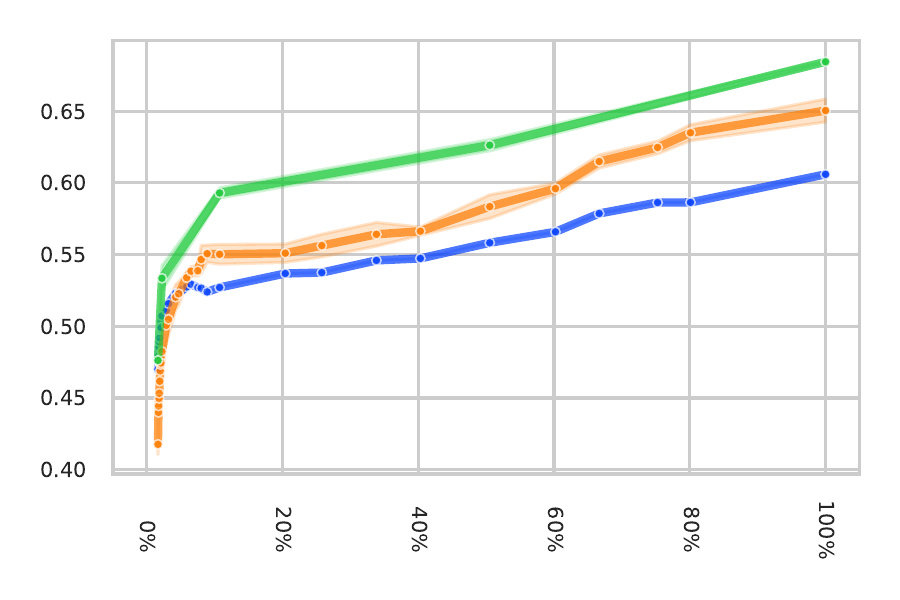}
            {\phantom{\ac{mAP}}}
            {Rel. Compute Requirement}
        \vspace*{-3mm}
        \caption{}
        \label{fig:knn_v_lin_terramind_ben}
    \end{subfigure}%
    \begin{subfigure}[t]{0.3\textwidth}
        \centering
        \tlabeledimage
            {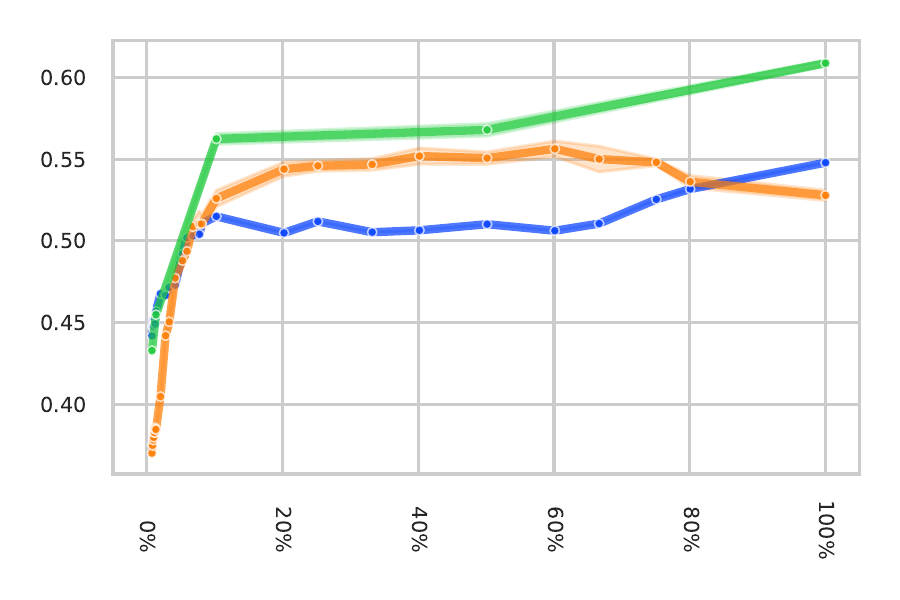}
            {\phantom{\ac{mAP}}}
            {\phantom{Rel. Compute Requirement}}
        \vspace*{-3mm}
        \caption{}
        \label{fig:knn_v_lin_prithvi_ben}
    \end{subfigure}

    \caption{\ac{KNN}, linear, and finetuned classification accuracy with respect to their relative compute requirements of: 
    a)~DOFA (large); 
    b)~TerraMind-1.0 (large); and
    c)~Prithvi-EO-2.0 (600M) on m-eurosat; and 
    d)~DOFA (large); 
    e)~TerraMind-1.0 (large); and
    f)~Prithvi-EO-2.0 (600M) on m-bigearthnet.}
    \label{fig:knn_v_lin}
\end{figure*}
We evaluate DOFA (large), TerraMind-1.0 (large), and Prithvi-EO-2.0 (600M) on m-eurosat and m-bigearthnet on \ac{KNN}, linear probing, and finetuning using the evaluation protocol described in \cref{subsec:linear_probing_protocol}.
The results of this evaluation are shown in \cref{fig:knn_v_lin}.
As one can see from the figure, for all evaluations except for m-bigearthnet on Prithvi-EO-2.0 (600M) the linear classification accuracy is higher for larger scales (with larger compute requirements).
Still, for m-bigearthnet on Prithvi-EO-2.0 (600M) the linear classification accuracy is higher for $0.0667 \leq s \leq 0.8$.
Additionally, for most cases there is a specific size $\Tilde{s}$ where the \ac{KNN} accuracy is higher if $s < \Tilde{s}$ and lower if $s > \Tilde{s}$.
Only for m-eurosat on Terramind-1.0 (large) the linear classification accuracy is higher than the \ac{KNN} accuracy for all sizes and for m-bigearthnet on Prithvi-EO-2.0 (600M) there are three sizes (\SI{5}{\percent}, \SI{6}{\percent}, and  \SI{100}{\percent}) that do not follow this monotonicity.

For finetuning, in both datasets and all models the accuracy matches or exceeds the accuracy of the linear probing.
Only for m-bigearthnet and DOFA (large) the linear probing model significantly outperforms the finetuned model.
Across all evaluated dataset-model combinations, the finetuned models show the same accuracy drop for very small scales as was seen in \ac{KNN} and linear probing experiments.

To summarize, in general pretrained models benefit from linear probing only for larger sizes whereas \ac{KNN} evaluation is stronger for models which were more rigorously slimmed.
However, the size $\Tilde{s}$ where smaller sizes benefit from \ac{KNN} evaluation is not consistent for a specific model or dataset.
Finetuned models also have a decrease in accuracy at small scales but are not as effected as models using linear probing.
\section{Effective Dimensionality Analysis}\label{sec:eff_dim}
\begin{figure*}
    \centering
    \noindent\includegraphics[width=0.98\linewidth]{figures/plots/legends/eff_dim_ds_legend_only.pdf}\par\vspace{-2mm}
    \noindent\includegraphics[width=0.98\linewidth]{figures/plots/legends/eff_dim_scale_legend_only.pdf}\par\vspace{-3mm}
    % Row 1
    \begin{subfigure}[t]{0.3\textwidth}
        \centering
        \tlabeledimage
            {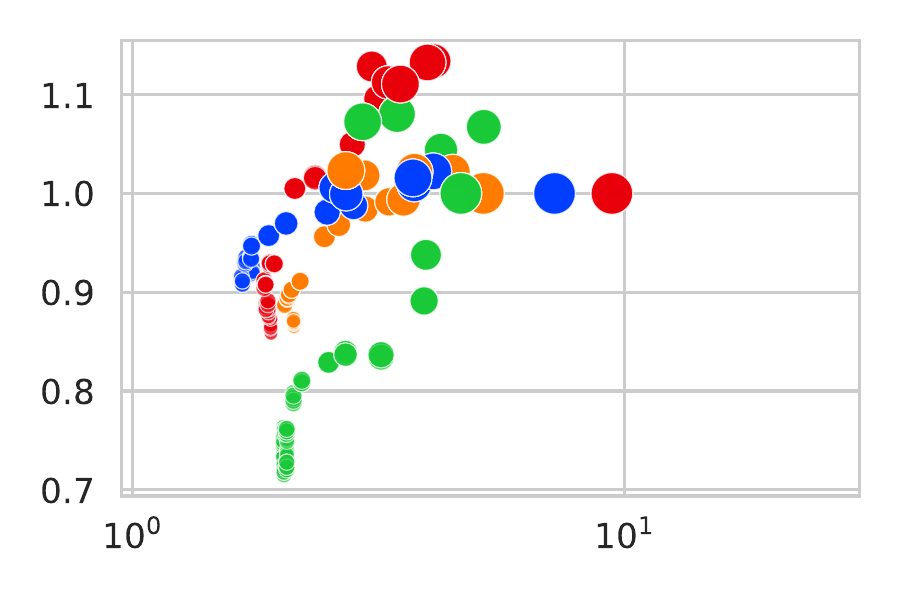}
            {Rel. Retention Rate}
            {\phantom{Effective Dimension}}
            [4cm]
            [1mm]
            [1mm]
        \vspace*{-6mm}
        \caption{}
        \label{fig:effDim-dofa}
    \end{subfigure}%
    \begin{subfigure}[t]{0.3\textwidth}
        \centering
        \tlabeledimage
            {figures/plots/explain/effective_dim/eff_dim_dofa_large_nolegend.pdf}
            {\phantom{Rel. Retention Rate}}
            {\phantom{Effective Dimension}}
        \vspace*{-6mm}
        \caption{}
        \label{fig:effDim-dofa-large}
    \end{subfigure}%
    \begin{subfigure}[t]{0.3\textwidth}
        \centering
        \tlabeledimage
            {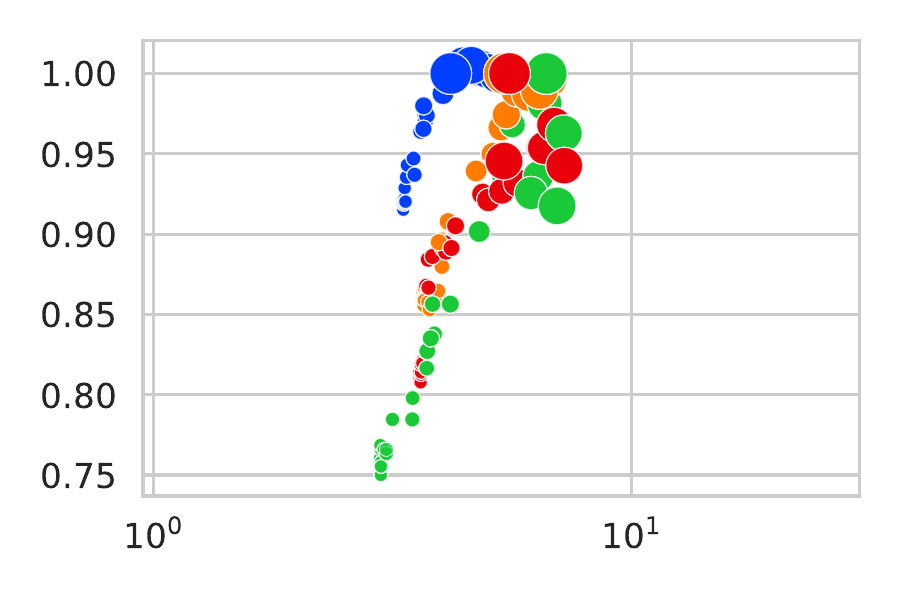}
            {\phantom{Rel. Retention Rate}}
            {\phantom{Effective Dimension}}
        \vspace*{-6mm}
        \caption{}
        \label{fig:effDim-terramind}
    \end{subfigure}%
    
    % Row 2
    \begin{subfigure}[t]{0.3\textwidth}
        \centering
        \tlabeledimage
            {figures/plots/explain/effective_dim/eff_dim_terramind_large_nolegend.pdf}
            {Rel. Retention Rate}
            {\phantom{Effective Dimension}}
            [4cm]
            [1mm]
            [1mm]
        \vspace*{-3mm}
        \caption{}
        \label{fig:effDim-terramind-large}
    \end{subfigure}%
    \begin{subfigure}[t]{0.3\textwidth}
        \centering
        \tlabeledimage
            {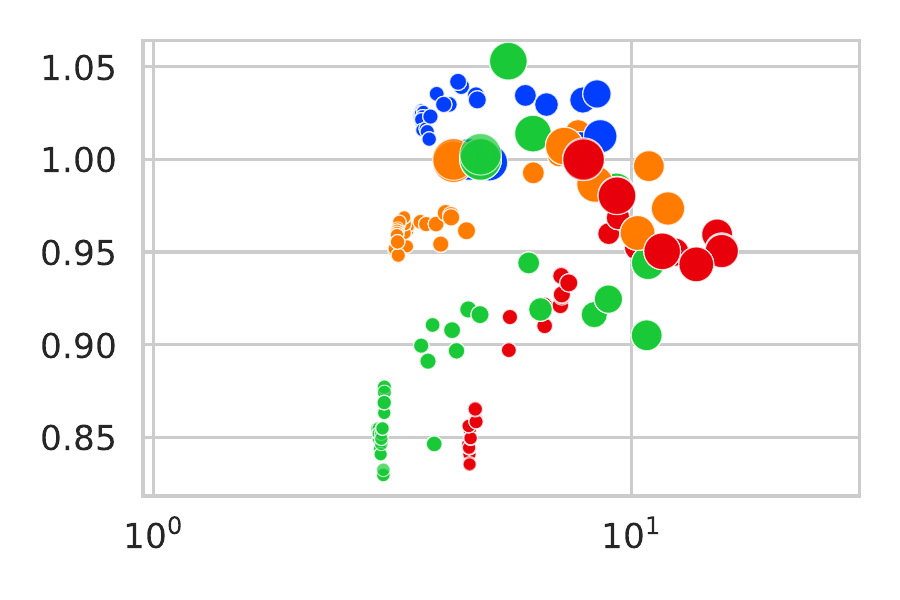}
            {\phantom{Rel. Retention Rate}}
            {Effective Dimension}
        \vspace*{-3mm}
        \caption{}
        \label{fig:effDim-prithvi300}
    \end{subfigure}%
    \begin{subfigure}[t]{0.3\textwidth}
        \centering
        \tlabeledimage
            {figures/plots/explain/effective_dim/eff_dim_prithvi2_600_nolegend.pdf}
            {\phantom{Rel. Retention Rate}}
            {\phantom{Effective Dimension}}
        \vspace*{-3mm}
        \caption{}
        \label{fig:effDim-prithvi600}
    \end{subfigure}

    % Row 3
    \begin{subfigure}[t]{0.3\textwidth}
        \centering
        \tlabeledimage
            {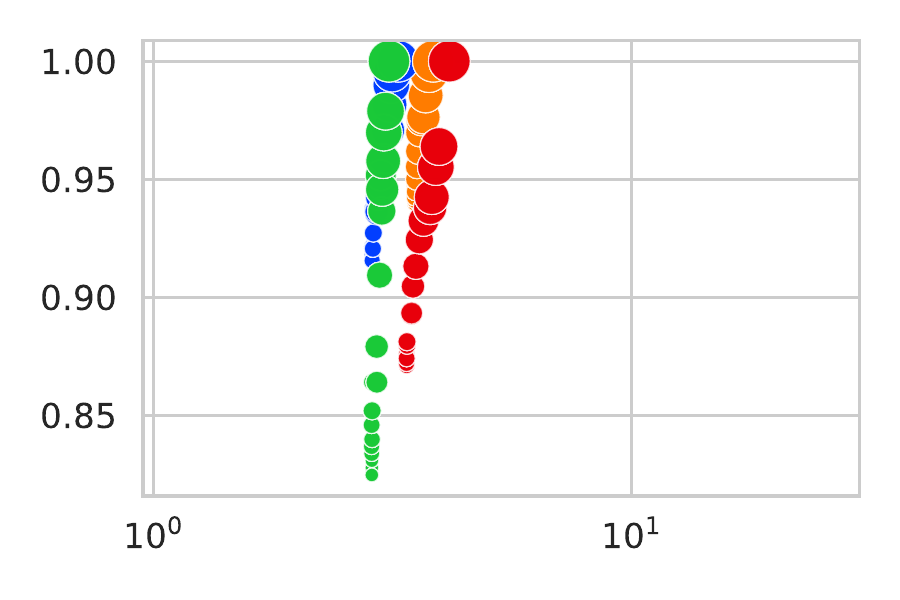}
            {Rel. Retention Rate}
            {\phantom{Effective Dimension}}
        \vspace*{-3mm}
        \caption{}
        \label{fig:effDim-ssl4eoDINO}
    \end{subfigure}%
    \begin{subfigure}[t]{0.3\textwidth}
        \centering
        \tlabeledimage
            {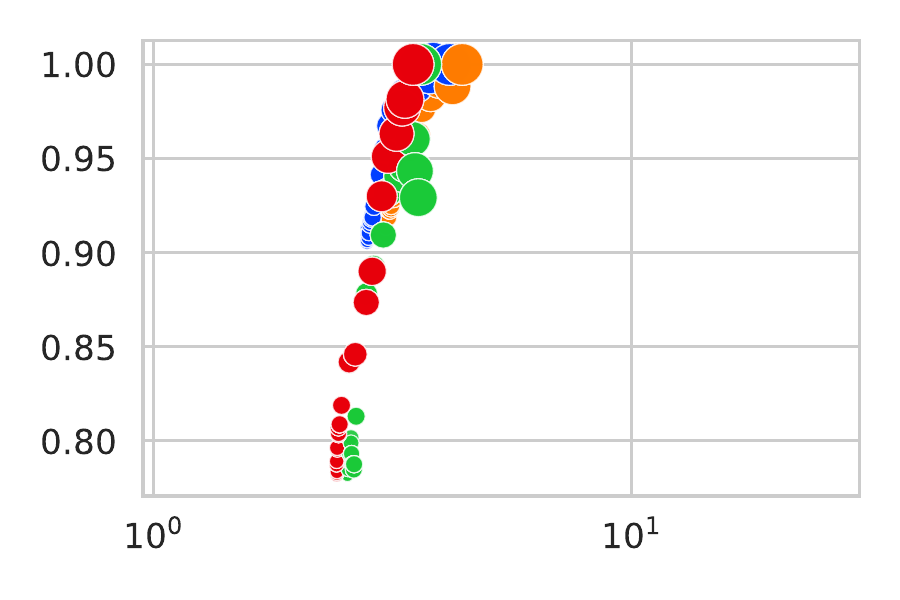}
            {\phantom{Rel. Retention Rate}}
            {\phantom{Effective Dimension}}
        \vspace*{-3mm}
        \caption{}
        \label{fig:effDim-ssl4eoMOCO}
    \end{subfigure}%

    % Row 3
    \begin{subfigure}[t]{0.3\textwidth}
        \centering
        \tlabeledimage
            {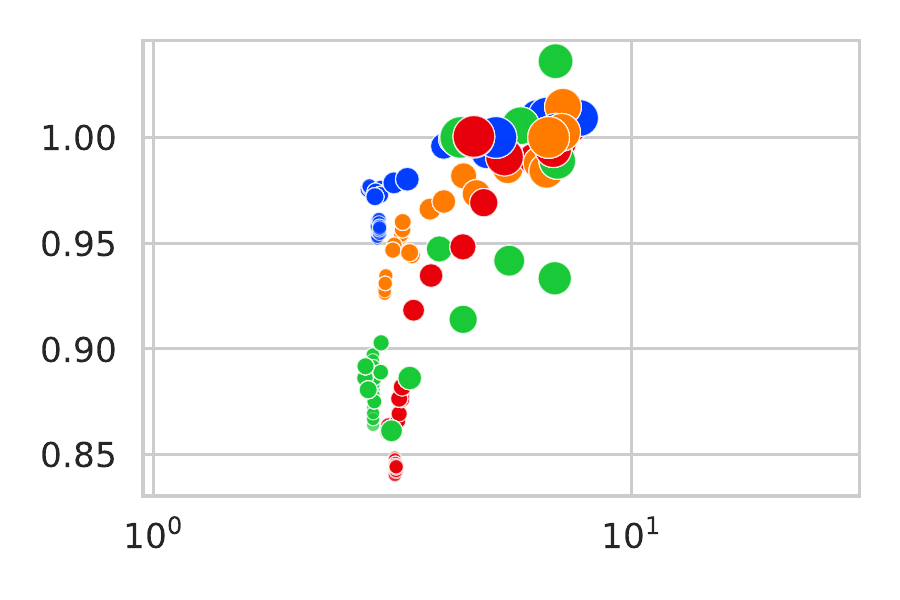}
            {Rel. Retention Rate}
            {\phantom{Effective Dimension}}
        \vspace*{-3mm}
        \caption{}
        \label{fig:effDim-copernicusFM}
    \end{subfigure}%
    \begin{subfigure}[t]{0.3\textwidth}
        \centering
        \tlabeledimage
            {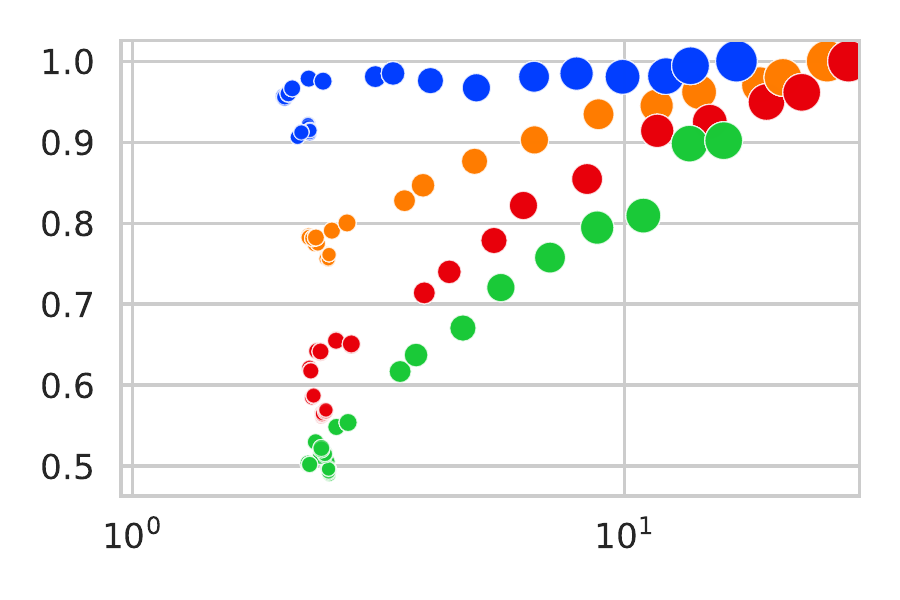}
            {\phantom{Rel. Retention Rate}}
            {\phantom{Effective Dimension}}
        \vspace*{-3mm}
        \caption{}
        \label{fig:effDim-panopticon}
    \end{subfigure}%

    \caption{Effective Dimension of the test split of four classification datasets from the geobench benchmark collection with respect to their relative retention rate on pretrained \ac{RS} \acp{FM}:
    a)~DOFA (base); 
    b)~DOFA (large); 
    c)~TerraMind-1.0 (base); 
    d)~TerraMind-1.0 (large);
    e)~Prithvi-EO-2.0 (300M);
    f)~Prithvi-EO-2.0 (600M);
    g)~SSL4EO ViT-S (Dino);
    h)~SSL4EO ViT-S (MoCo);
    i)~CopernicusFM; and
    j)~Panopticon.}
    \label{fig:effDim}
\end{figure*}

\cref{fig:effDim} relates effective dimensionality to relative retention across the evaluated \ac{RS} \acp{FM}.
The analysis is descriptive: it shows how variance is distributed across feature dimensions, not which dimensions carry task-relevant information.

\paragraph{DOFA.}
DOFA models (\cref{fig:effDim-dofa,fig:effDim-dofa-large}) concentrate variance in few dimensions: effective dimensionality is between \num{1.4} and \num{2.0} at \SI{1}{\percent} compute, while \SIrange{53}{92}{\percent} of full-scale performance is retained.
Reaching full performance goes along with a much larger effective dimensionality, up to \num{34} for the large model on the multi-label m-bigearthnet (roughly $16$--$24\times$ the minimum-compute value).

\paragraph{SSL4EO.}
The SSL4EO ViT-S models (\cref{fig:effDim-ssl4eoDINO,fig:effDim-ssl4eoMOCO}) have the lowest variance of effective dimensionality of all evaluated models, within \numrange{2.4}{4.4} across datasets and compute scales, with monotonic trajectories. %CHECK: "monotonic" for every dataset and seed
On the simpler m-eurosat and m-brick-kiln, both models retain \SIrange{91}{94}{\percent} at \SI{1}{\percent} compute with effective dimensionalities of \numrange{2.8}{3.5}, which increase to \numrange{3.2}{4.4} at full scale.
On the more complex m-bigearthnet and m-so2sat, retention at low compute drops to \SIrange{78}{87}{\percent}, while full-scale effective dimensionality stays at \numrange{3.1}{4.2}.
DINO shows slightly higher effective dimensionality and retention on the harder datasets than MoCo. 

\paragraph{Prithvi-EO-2.0 (600M).}
At minimum compute, Prithvi-EO-2.0 (600M) (\cref{fig:effDim-prithvi600}) retains \SIrange{81}{103}{\percent} with effective dimensionality between \num{2.4} and \num{4.7}, comparable to SSL4EO.
The trajectory is strongly non-monotonic: effective dimensionality peaks at \numrange{7.7}{10.3} around \SIrange{40}{50}{\percent} compute and decreases to \numrange{2.8}{8.1} at full scale, mirroring the non-monotonic \ac{EVR} slope (\cref{fig:loglog_slope_evr}).
Across this range, retention varies by less than 15 percentage points while effective dimensionality differs by more than $2.5\times$.

\paragraph{TerraMind-1.0.}
TerraMind-1.0 base and large (\cref{fig:effDim-terramind,fig:effDim-terramind-large}) start at slightly higher minimum-compute effective dimensionality than DOFA (\numrange{3.0}{3.7} for base, \numrange{3.1}{3.6} for large), with retention at \SI{1}{\percent} compute of \SIrange{75}{92}{\percent} (base) and \SIrange{75}{89}{\percent} (large), lower on harder datasets.
Full-scale effective dimensionality reaches \numrange{4.2}{6.6} (base) and \numrange{4.3}{6.3} (large), an increase of roughly $1.5$--$2.2\times$ from minimum to full scale.
At high compute, effective dimensionality typically peaks at intermediate scales before plateauing or decreasing.

\paragraph{Copernicus-FM.}
Copernicus-FM (\cref{fig:effDim-copernicusFM}) closely resembles TerraMind-1.0 base: at the smallest scale, effective dimensionality is \numrange{2.7}{3.3} with a retention of \SIrange{84}{98}{\percent}, lower on the harder datasets, and it increases by a factor of about $1.5$--$2.3\times$ to \numrange{4.5}{7} at full scale.

\paragraph{Panopticon.}
Panopticon (\cref{fig:effDim-panopticon}) behaves similarly to DOFA (large): at the smallest scale, effective dimensionality is low (\numrange{2.2}{2.7}) with a retention of \SIrange{49}{91}{\percent}, lowest on m-so2sat and highest on m-brick-kiln. 
Effective dimensionality grows by more than an order of magnitude toward full scale, accompanied by retention gains of up to about 45 percentage points.
Note that Panopticon's compute never falls below roughly \SI{60}{\percent} because its patch embedding is not slimmed (\cref{sec:slimdims}), so its smallest scale does not correspond to the minimum compute of the other models.

\paragraph{Summary.}
The relation between effective dimensionality and retention differs across models: DOFA and Panopticon show a large dimensional expansion alongside sizeable retention gains, whereas Prithvi-EO-2.0 and TerraMind-1.0 retain performance across scales with markedly different effective dimensionality.
The non-monotonic pattern shared by Prithvi-EO-2.0 and TerraMind-1.0 (and absent in SSL4EO) is not explained by these analyses; incomplete convergence during pretraining is one untested possibility.
\section{Additional Experiments on Post-Hoc Slimmability}\label{sec:add_post-hoc}

\subsection{Additional Experiments on Classification}
\begin{figure*}
    \centering
    \noindent\includegraphics[width=0.98\linewidth]{figures/plots/legends/post_hoc_slimming_existing_fms_legend_only.pdf}\par\vspace{-2mm}
    \begin{subfigure}[t]{0.40\textwidth}
        \centering
        \tlabeledimage
            {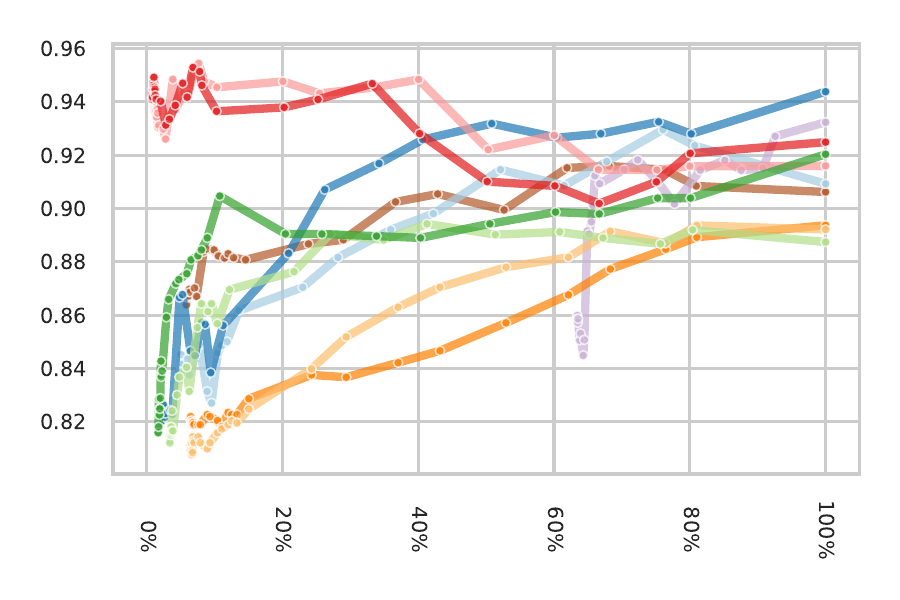}
            {Accuracy}
            {Rel. Compute Requirement}
        \vspace*{-3mm}
        \caption{}
        \label{fig:post-hoc-brick}
    \end{subfigure}%
    \begin{subfigure}[t]{0.40\textwidth}
        \centering
        \tlabeledimage
            {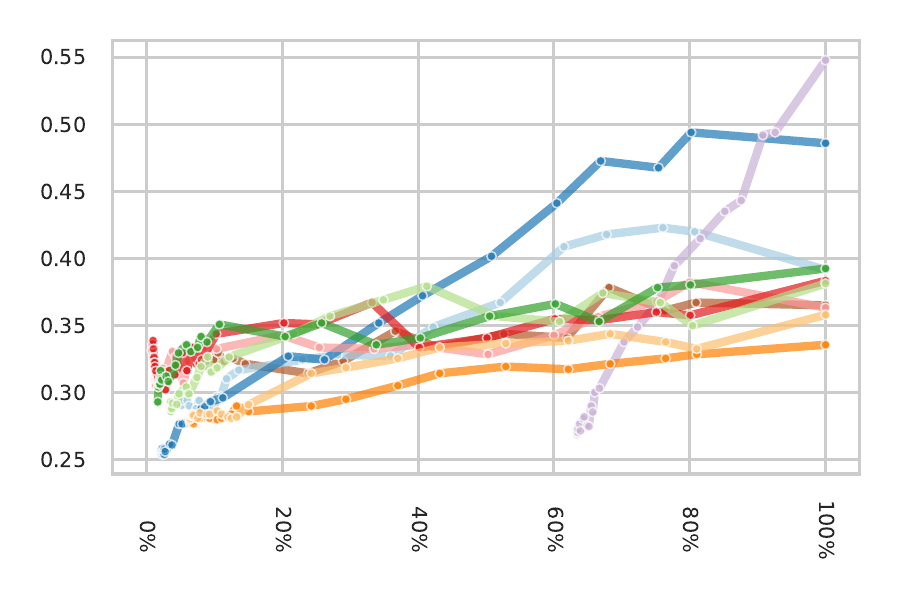}
            {\phantom{Accuracy}}
            {Rel. Compute Requirement}
        \vspace*{-3mm}
        \caption{}
        \label{fig:post-hoc-so2sat}
    \end{subfigure}%
    
    \caption{\ac{KNN} classification performance with respect to their relative compute requirements of ten pretrained \ac{RS} \acp{FM} on:
    a)~m-brick-kiln; and b)~m-so2sat.}
    \label{fig:post-hoc-part2}
\end{figure*}

\paragraph{m-brick-kiln.}
As shown in \cref{fig:post-hoc-brick}, all \ac{RS} \acp{FM} exhibit high retention at minimum compute, consistent with the low task complexity of binary scene classification.
Both Prithvi-EO-2.0 models retain more than full-scale performance across the majority of the compute range, with Prithvi-EO-2.0 (300M) reaching \SI{102}{\percent} and Prithvi-EO-2.0 (600M) reaching \SI{103}{\percent} relative retention even at minimum compute budget.
Performance remains above full scale through approximately \SI{40}{\percent} compute before converging.
TerraMind-1.0 (base) similarly exceeds full-scale performance at intermediate scales (\SIrange{27}{80}{\percent} compute), while TerraMind-1.0 (large) reaches full scale without surpassing it, apart from a small dip between \SIrange{10}{40}{\percent} compute.
DOFA (base) retains \SI{92}{\percent} at minimum compute and exceeds full scale by up to \SI{2}{\percent} at intermediate scales before converging; DOFA (large) retains \SI{87}{\percent} at minimum compute and increases monotonically.
CopernicusFM retains \SI{97}{\percent} at minimum compute and exceeds full scale by up to \SI{1}{\percent} between \SIrange{60}{80}{\percent} compute.
Both SSL4EO models reach their minimum compute at approximately \SI{6}{\percent} of full budget due to their smaller architecture, retaining about \SI{91}{\percent} and increasing monotonically to full scale.
Panopticon, whose relative compute does not fall below roughly \SI{60}{\percent} (see Appendix~\ref{sec:slimdims}), retains \SI{91}{\percent} at its smallest scale and increases to full scale.
The non-monotonic behavior, where intermediate-scale models outperform full-scale models, is consistent with the pattern observed on m-eurosat.

\paragraph{m-so2sat.}
As shown in \cref{fig:post-hoc-so2sat}, in this 17-class single-label classification task, retention at minimum compute is substantially lower than on m-brick-kiln, reflecting the higher task complexity.
DOFA (base) retains \SI{75}{\percent} at minimum compute and exhibits pronounced non-monotonic behavior, with performance at \SI{76}{\percent} and \SI{81}{\percent} compute exceeding full scale by \SI{8}{\percent} and \SI{7}{\percent} respectively before falling back at full scale.
DOFA (large) shows the lowest minimum-compute retention of any model across all evaluated tasks (except for Panopticon, excluded due to embedding layer differences as noted in Appendix~\ref{sec:slimdims}), retaining only \SI{53}{\percent} at minimum compute, with gradual recovery to full scale.
This is consistent with the pronounced non-monotonic peak on m-bigearthnet observed in \cref{sec:post-hoc} and suggests that DOFA (large) is particularly sensitive to width reduction on complex multi-label tasks.
Prithvi-EO-2.0 (300M) retains \SI{85}{\percent} at minimum compute and peaks at \SI{5}{\percent} above full scale near \SI{80}{\percent} compute before converging.
Prithvi-EO-2.0 (600M) retains \SI{87}{\percent} at minimum compute but exhibits an irregular trajectory across the compute range, with retention varying between \SI{79}{\percent} and \SI{96}{\percent} at intermediate scales before reaching full scale.
Both SSL4EO models retain \SIrange{78}{83}{\percent} at their minimum compute budget and increase monotonically, with MoCo showing a slight non-monotonic dip near \SI{81}{\percent} compute.
Both TerraMind models retain about \SI{75}{\percent} at minimum compute, with TerraMind-1.0 (base) peaking near full-scale performance at \SI{41}{\percent} compute before a non-monotonic dip at \SIrange{50}{61}{\percent} compute and recovery to full scale.
Across all models, the critical compute threshold below which performance drops sharply is higher on m-so2sat than on any classification task reported in \cref{sec:post-hoc}, indicating that task complexity is the primary determinant of the effective lower bound on compute budget.

\subsection{Additional Experiment on Segmentation}
\begin{figure*}
    \centering   
    \noindent\includegraphics[width=0.98\linewidth]{figures/plots/legends/post_hoc_slimming_existing_fms_legend_only.pdf}\par\vspace{-2mm}
    \tlabeledimage
        {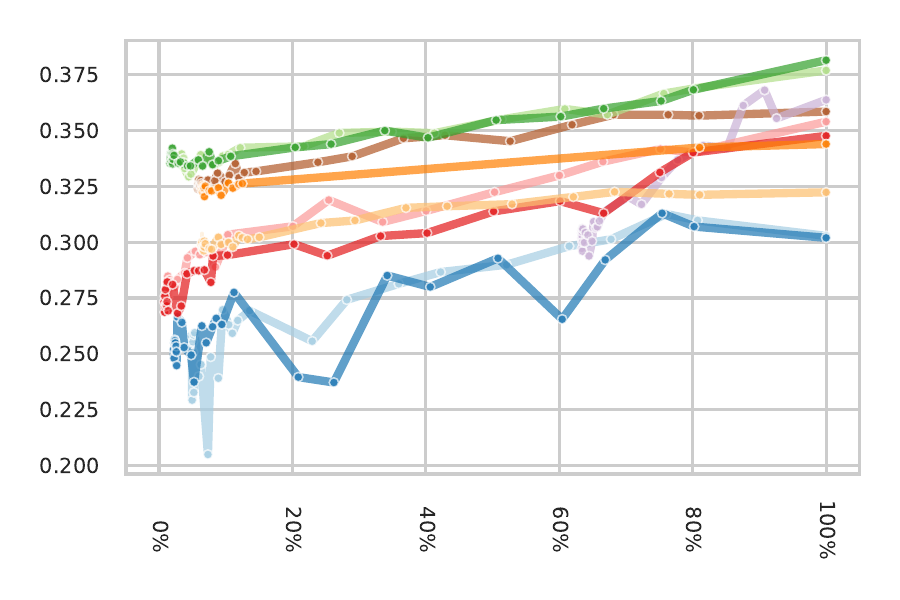}
        {\ac{IoU}}
        {Rel. Compute Requirement}
    \vspace*{-3mm}
    
    \caption{\ac{IoU} with respect to relative compute requirements of ten pretrained \ac{RS} \acp{FM} on m-SA-crop-type (10-class semantic segmentation).}
    \label{fig:post-hoc-SAcrop}
\end{figure*}

\paragraph{m-SA-crop-type.}
On this 10-class semantic segmentation task, slimmability is robust across all models, with retention at minimum compute ranging from \SIrange{78}{94}{\percent}.
Both SSL4EO ViT-S models exhibit the highest retention at minimum compute of all evaluated models on this task, with the DINO model retaining \SI{94}{\percent} and the MoCo model retaining \SI{93}{\percent} at approximately \SI{6}{\percent} compute.
Both curves are remarkably flat across the full compute range, with DINO retention varying by less than \SI{2}{\percent} between minimum compute and \SI{81}{\percent} compute before converging to full scale, and MoCo remaining similarly stable before marginally exceeding full scale by \SI{0.1}{\percent} at approximately \SI{68}{\percent} compute.
TerraMind-1.0 models also exhibit stable trajectories.
TerraMind-1.0 (base) retains \SI{89}{\percent} at minimum compute with a gradual, monotonically increasing curve through full scale.
TerraMind-1.0 (large) retains \SI{88}{\percent} at minimum compute and shows the flattest curve of the large-scale models, increasing gradually and monotonically to full scale without non-monotonic deviations.
CopernicusFM retains \SI{91}{\percent} at minimum compute and increases gradually and almost monotonically to full scale.
Panopticon retains \SI{84}{\percent} at its smallest scale and slightly exceeds full scale by about \SI{1}{\percent} near \SI{90}{\percent} compute.
DOFA (base) retains \SI{82}{\percent} at minimum compute but exhibits the most irregular trajectory of any model on this task, including a pronounced dip to \SI{68}{\percent} retention near \SI{7}{\percent} compute followed by recovery to \SI{89}{\percent} at \SI{10}{\percent} compute.
Both DOFA models exceed full scale by \SIrange{2}{4}{\percent} at approximately \SI{77}{\percent} compute, consistent with the non-monotonic pattern observed across other tasks.
DOFA (large) retains \SI{84}{\percent} at minimum compute with a similarly irregular mid-range trajectory, including a secondary dip to \SI{79}{\percent} near \SI{26}{\percent} compute.
Both Prithvi-EO-2.0 models retain \SIrange{78}{79}{\percent} at minimum compute and show generally increasing but irregular trajectories across the full compute range, never exceeding full-scale performance.
Prithvi-EO-2.0 (300M) shows a notable non-monotonic dip at \SI{33}{\percent} compute after peaking at \SI{25}{\percent} compute (\SI{90}{\percent} retention), while Prithvi-EO-2.0 (600M) varies irregularly between \SI{77}{\percent} and \SI{85}{\percent} retention through the moderate compute regime before converging.
The task-complexity-dependent pattern holds: minimum-compute retention on m-SA-crop-type is lower than on the simpler m-cashew-plant task reported in \cref{sec:post-hoc}, consistent with the 10-class label structure requiring greater representational capacity at small widths.
\section{Additional Experiments on Learned vs Post-Hoc Slimmability}\label{sec:add_learned_v_post-hoc}
\begin{figure*}
    \centering
    \noindent\includegraphics[width=0.98\linewidth]{figures/plots/legends/native_vs_posthoc_legend_only.pdf}\par\vspace{-2mm}
    \begin{subfigure}[t]{0.35\textwidth}
        \centering
        \tlabeledimage
            {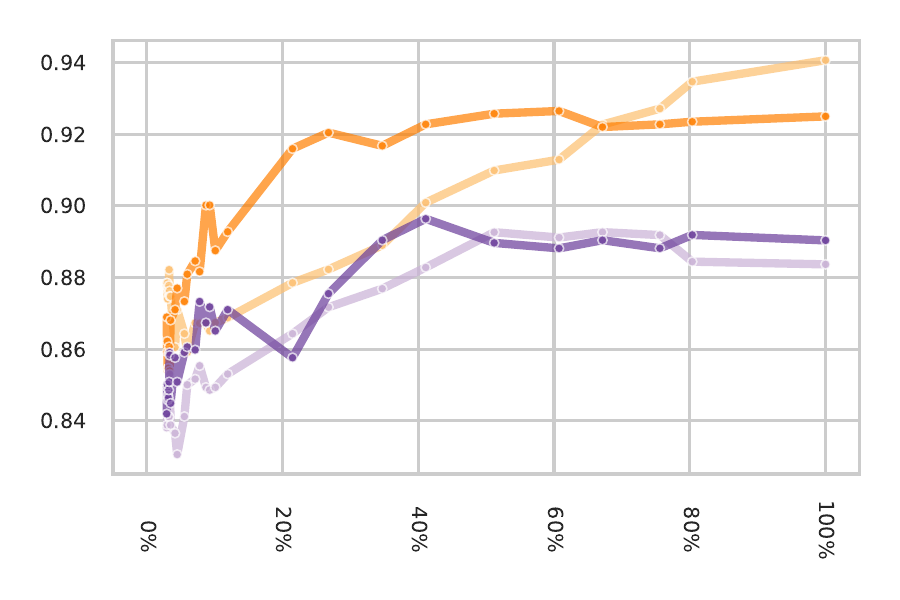}
            {Accuracy}
            {Rel. Compute Requirement}
        \vspace*{-3mm}
        \caption{}
        \label{fig:train-son-brick}
    \end{subfigure}%
    \begin{subfigure}[t]{0.35\textwidth}
        \centering
        \tlabeledimage
            {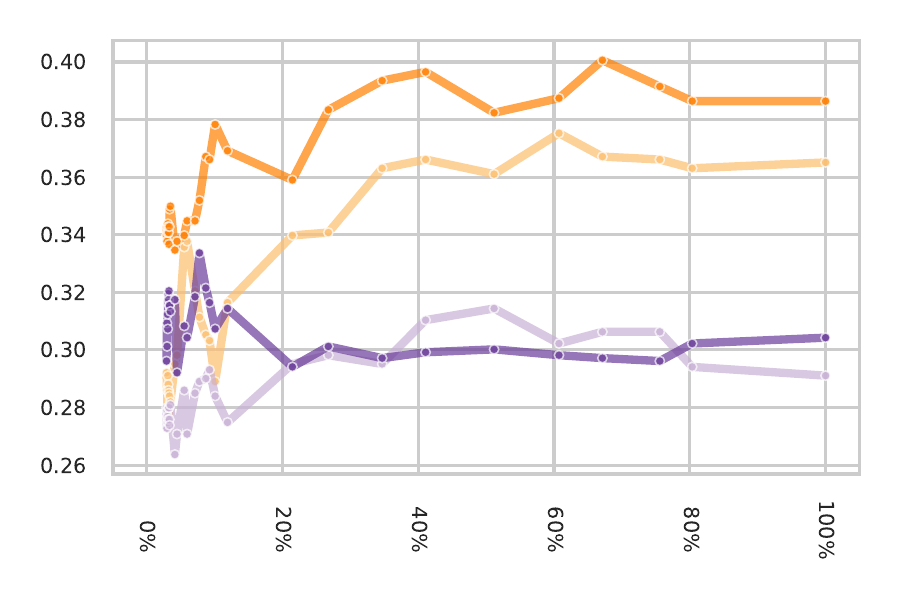}
            {\phantom{Accuracy}}
            {Rel. Compute Requirement}
        \vspace*{-3mm}
        \caption{}
        \label{fig:train-son-so2sat}
    \end{subfigure}%
    
    \caption{\ac{KNN} classification performance with respect to their relative compute requirements of a \ac{MAE}-based and a \ac{MoCo}-based trained model from scratch, each once with and once without slimmable training on:
    a)~m-brick-kiln; and b)~m-so2sat.}
    \label{fig:train-son-part2}
\end{figure*}
We provide additional results for the remaining two datasets (m-brick-kiln and m-so2sat) to complement the main analysis of learned versus post-hoc slimmability.
These additional evaluations further validate the training paradigm-dependent patterns observed on m-eurosat and m-bigearthnet, while revealing nuanced task-complexity effects.

\textbf{Binary and fine-grained classification results.}
For the binary m-brick-kiln dataset (\cref{fig:train-son-brick}), slimmable \ac{MoCo} achieves marginal improvements at low computational budgets (\num{0.842} vs \num{0.838} accuracy at \SI{3}{\percent} compute) and peaks at \num{0.896} at \SI{41}{\percent} compute compared to \num{0.893} for regular training at \SI{51}{\percent} compute.
Both trained models converge near \num{0.89} at moderate scales, confirming that learned slimmability maintains performance parity at full capacity while improving low-scale robustness.
For the learned slimmable \ac{MAE}, the pattern shows disadvantages at very low scales (\num{0.869} vs \num{0.878} at \SI{3}{\percent}) that inverse at intermediate budgets (\num{0.916} vs \num{0.879} at \SI{21}{\percent}), though regular \ac{MAE} ultimately dominates at maximum scale (\num{0.940} vs \num{0.925}).

The fine-grained m-so2sat dataset (\cref{fig:train-son-so2sat}) reveals more pronounced differences.
Slimmable \ac{MoCo} substantially outperforms regular training, achieving \num{0.296} accuracy at \SI{3}{\percent} compute versus \num{0.280} and reaching a peak of \num{0.334} at \SI{8}{\percent} compute, thereby surpassing regular \ac{MoCo}'s maximum of \num{0.314} at \SI{51}{\percent} compute by \SI{6.4}{\percent} relative accuracy while requiring less than \SI{20}{\percent} of the compute.
Remarkably, the learned slimmable \ac{MAE} completely reverses the degradation pattern observed on m-bigearthnet, dominating across all evaluated scales with \num{0.340} accuracy at \SI{3}{\percent} compute versus \num{0.292} for regular \ac{MAE}, maintaining this advantage through full scale (\num{0.386} vs \num{0.365}).

\textbf{Unclear relation betwee task complexity and MAE slimmability.}
These additional results establish no clear relationship between task complexity and \ac{MAE} slimmability performance.
Learned slimmable \ac{MAE} exhibits severe degradation on the complex multi-label m-bigearthnet task, marginal disadvantages on the binary m-brick-kiln task, underperformance on the intermediate m-eurosat task, but strong advantages on the fine-grained single-label m-so2sat task.
The benefit of learned slimmability for \ac{MAE} does not follow a clear pattern across tasks. 
It is strongly positive on m-so2sat, mixed on m-brick-kiln, negative on m-eurosat, and severely negative on m-bigearthnet. 
We do not have an explanation for this variation, and it does not track class count monotonically.
In contrast, \ac{MoCo}'s instance discrimination objective demonstrates consistent slimmability improvements across all task complexities, confirming that contrastive learning and slimmable training can be successfully combined to develop \acp{FM} for flexible deployment across diverse downstream applications.
\section{Feature Correlation at different Model Scales}\label{sec:feat_corr_model_scales}
\begin{figure*}
    \centering
    \noindent\includegraphics[width=0.98\linewidth]{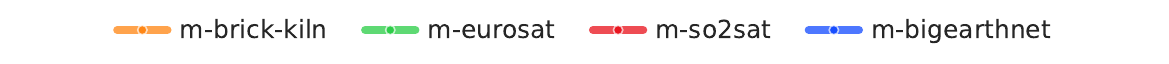}\par\vspace{-2mm}
    % Row 1
    \begin{subfigure}[b]{0.3\textwidth}
        \centering
        \tlabeledimage
            {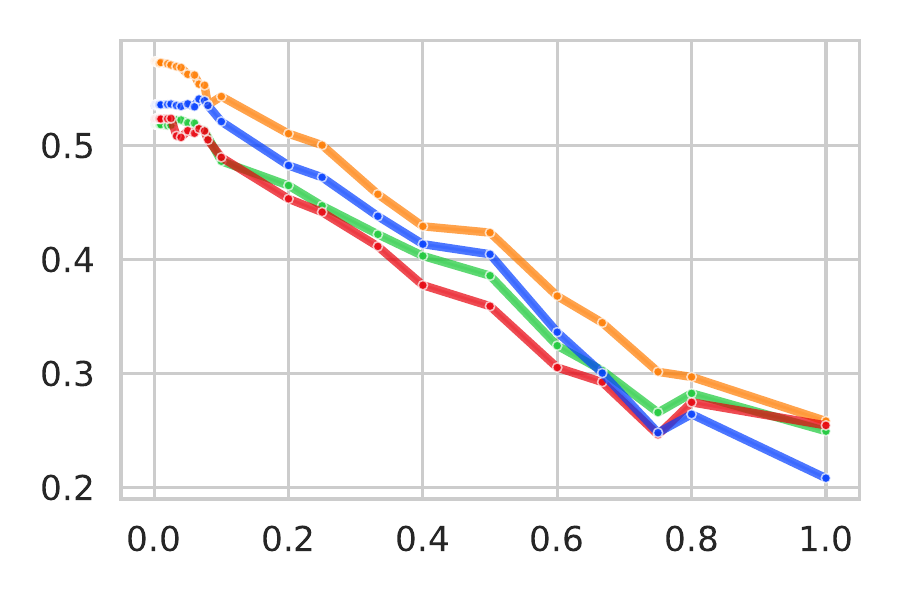}
            {Mean Feat. Corr.}
            {\phantom{Rel. Compute Requirement}}
        \vspace*{-5mm}
        \caption{}
        \label{fig:corr_full-dofa}
    \end{subfigure}%
    \begin{subfigure}[b]{0.3\textwidth}
        \centering
        \tlabeledimage
            {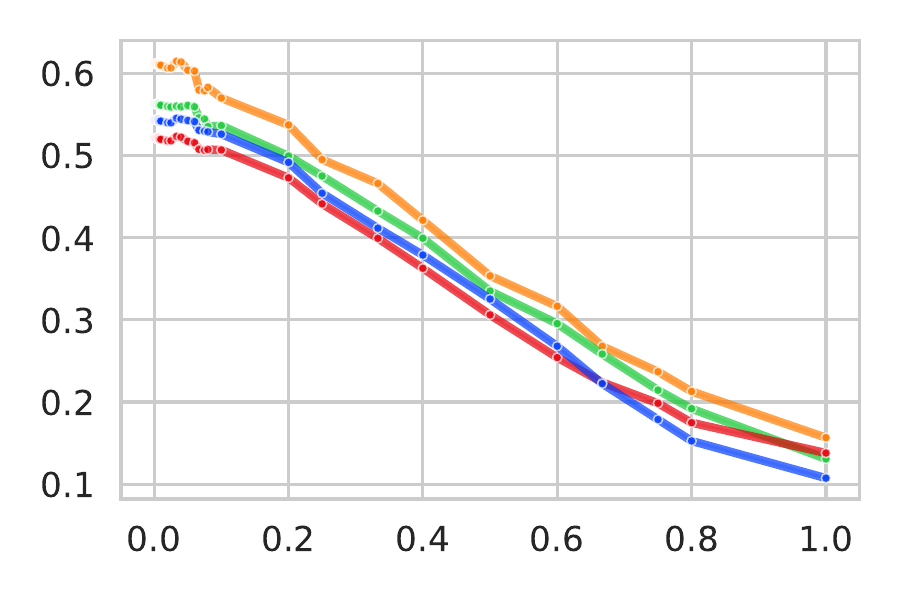}
            {\phantom{Mean Feat. Corr.}}
            {\phantom{Rel. Compute Requirement}}
        \vspace*{-5mm}
        \caption{}
        \label{fig:corr_full-dofa-large}
    \end{subfigure}%
    \begin{subfigure}[b]{0.3\textwidth}
        \centering
        \tlabeledimage
            {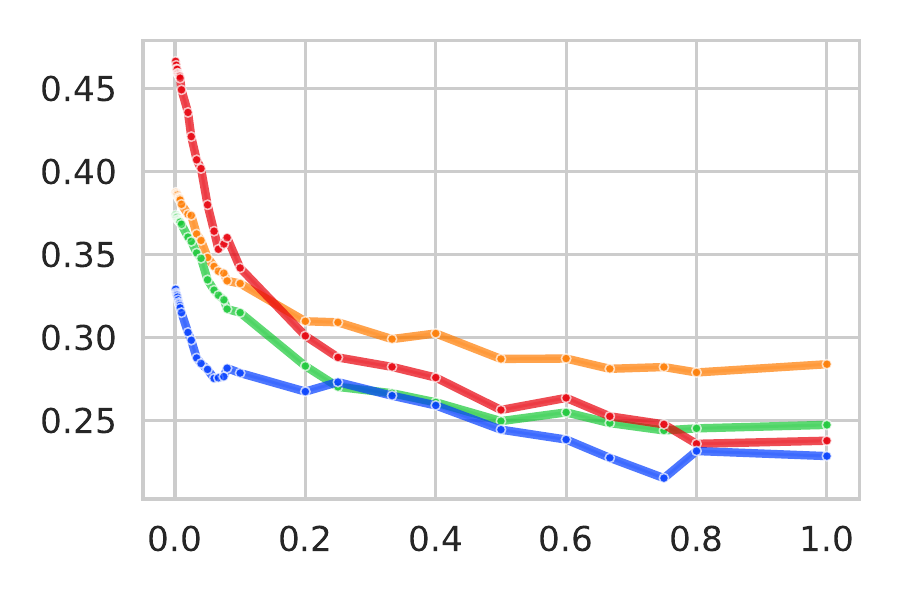}
            {\phantom{Mean Feat. Corr.}}
            {\phantom{Rel. Compute Requirement}}
        \vspace*{-5mm}
        \caption{}
        \label{fig:corr_full-terramind}
    \end{subfigure}%
    
    % Row 2
    \begin{subfigure}[b]{0.3\textwidth}
        \centering
        \tlabeledimage
            {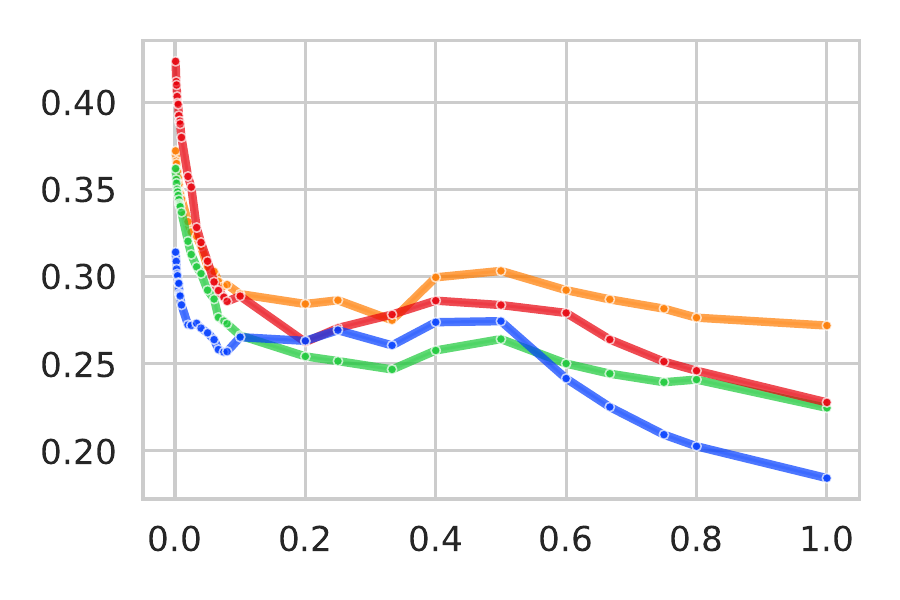}
            {Mean Feat. Corr.}
            {\phantom{Rel. Compute Requirement}}
        \vspace*{-2mm}
        \caption{}
        \label{fig:corr_full-terramind-large}
    \end{subfigure}%
    \begin{subfigure}[b]{0.3\textwidth}
        \centering
        \tlabeledimage
            {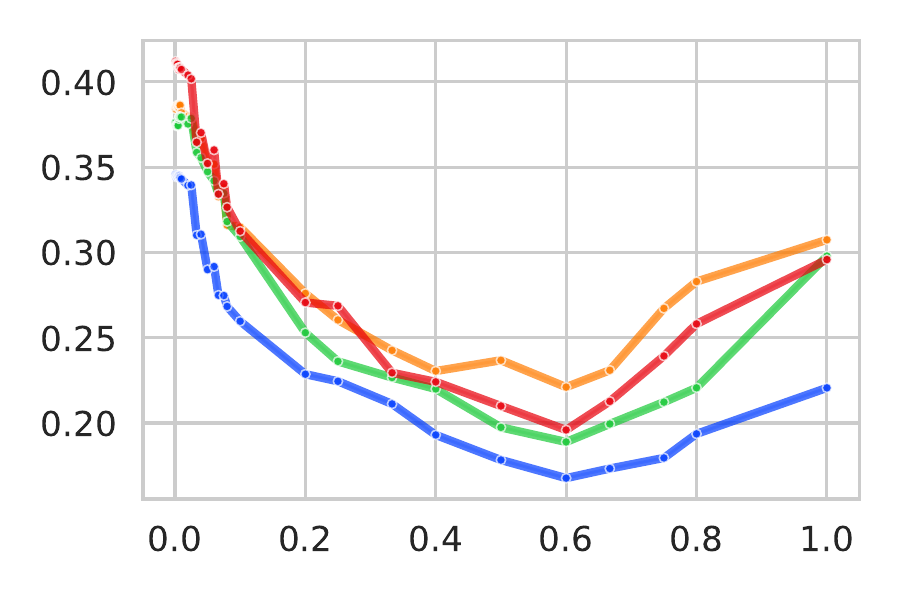}
            {\phantom{Mean Feat. Corr.}}
            {Model Scale}
        \vspace*{-2mm}
        \caption{}
        \label{fig:corr_full-prithvi300}
    \end{subfigure}%
    \begin{subfigure}[b]{0.3\textwidth}
        \centering
        \tlabeledimage
            {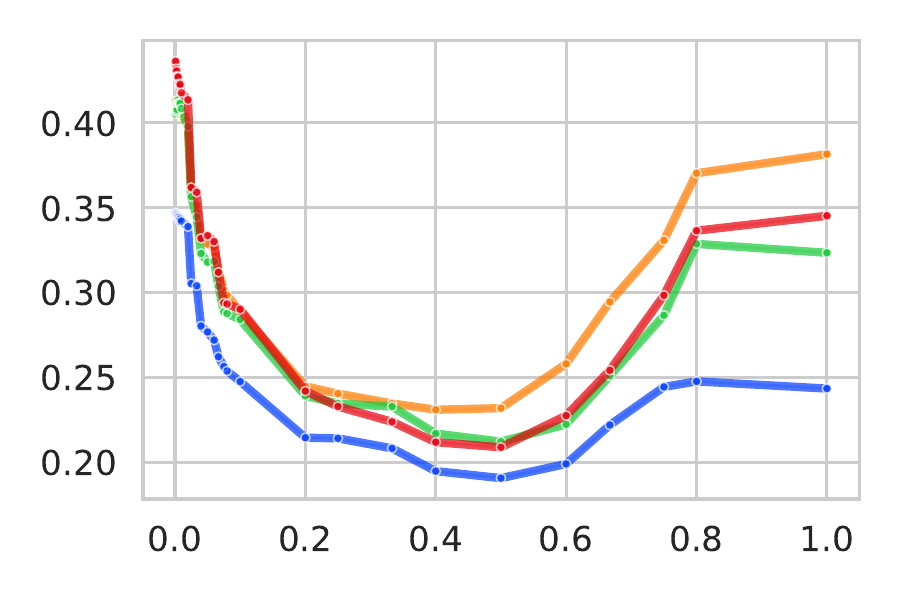}
            {\phantom{Mean Feat. Corr.}}
            {\phantom{Rel. Compute Requirement}}
        \vspace*{-2mm}
        \caption{}
        \label{fig:corr_full-prithvi600}
    \end{subfigure}%
    
    % Row 3
    \begin{subfigure}[b]{0.3\textwidth}
        \centering
        \tlabeledimage
            {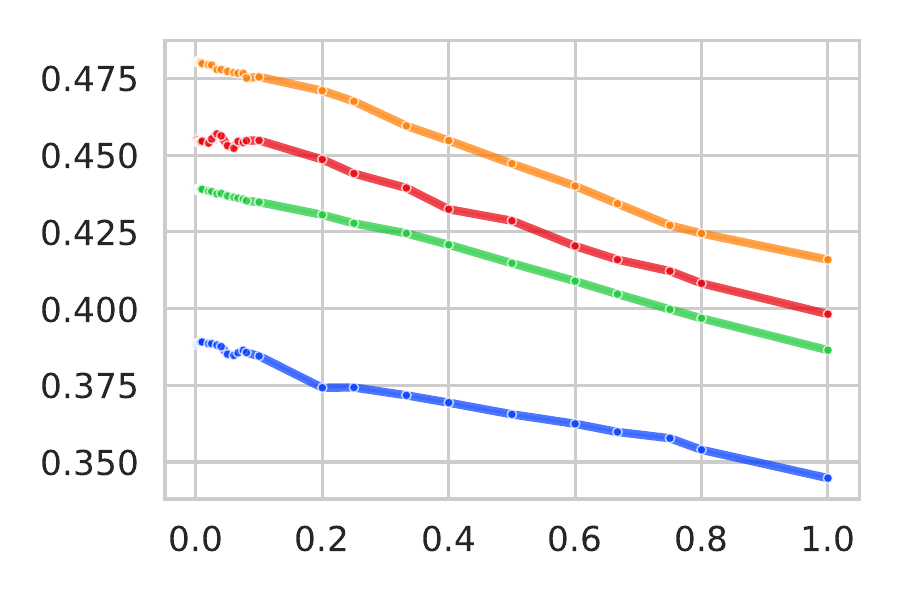}
            {Mean Feat. Corr.}
            {\phantom{Rel. Compute Requirement}}
        \vspace*{-2mm}
        \caption{}
        \label{fig:corr_full-ssl4eo_dino}
    \end{subfigure}%
    \begin{subfigure}[b]{0.3\textwidth}
        \centering
        \tlabeledimage
            {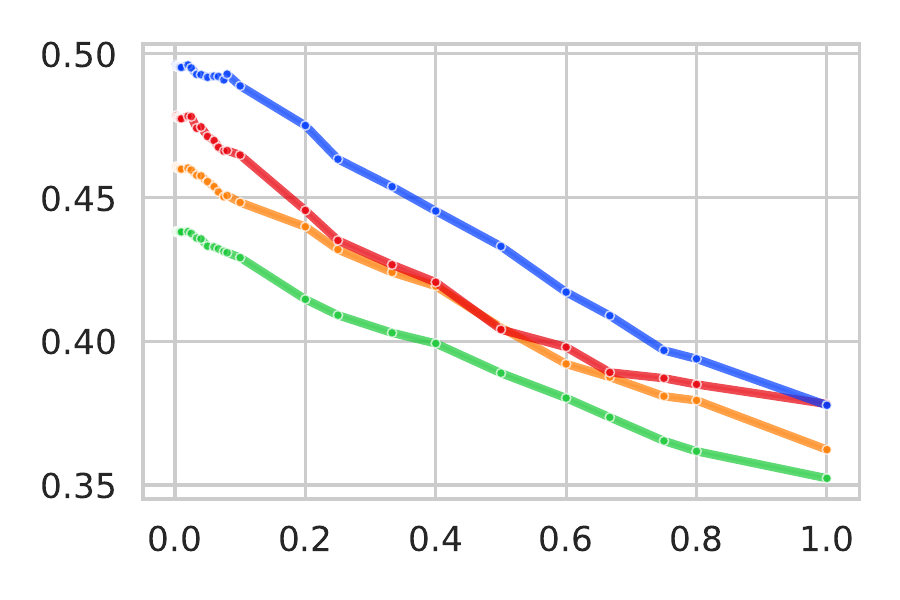}
            {\phantom{Mean Feat. Corr.}}
            {\phantom{Rel. Compute Requirement}}
        \vspace*{-2mm}
        \caption{}
        \label{fig:corr_full-ssl4eo_moco}
    \end{subfigure}%
    
    % Row 4
    \begin{subfigure}[b]{0.3\textwidth}
        \centering
        \tlabeledimage
            {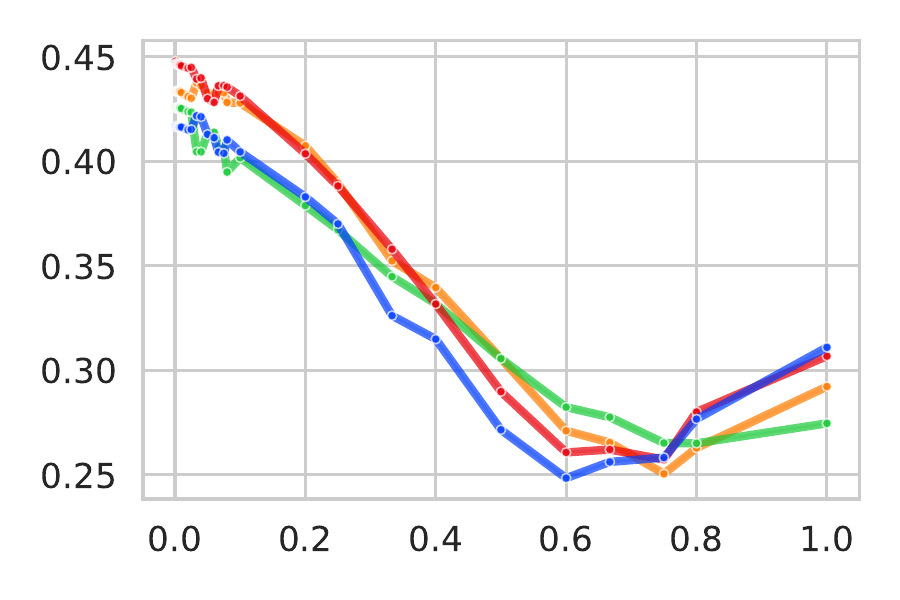}
            {Mean Feat. Corr.}
            {\phantom{Rel. Compute Requirement}}
        \vspace*{-2mm}
        \caption{}
        \label{fig:corr_full-copernicusfm}
    \end{subfigure}%
    \begin{subfigure}[b]{0.3\textwidth}
        \centering
        \tlabeledimage
            {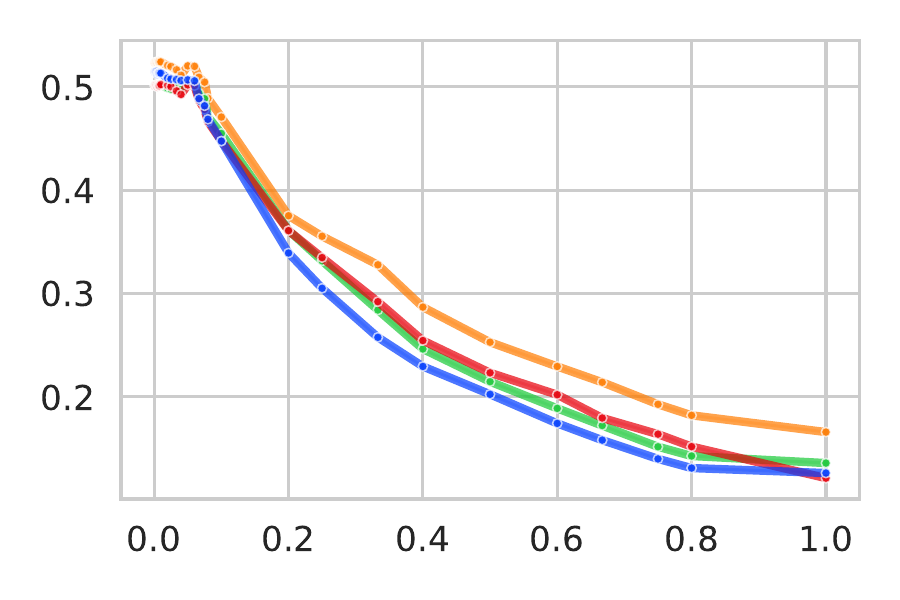}
            {\phantom{Mean Feat. Corr.}}
            {\phantom{Rel. Compute Requirement}}
        \vspace*{-2mm}
        \caption{}
        \label{fig:corr_full-panopticon}
    \end{subfigure}%

    \caption{Mean Feature Correlation of the test split of four classification datasets from the geobench benchmark collection with respect to the model scale on pretrained \ac{RS} \acp{FM}:
    a)~DOFA (base); 
    b)~DOFA (large); 
    c)~TerraMind-1.0 (base); 
    d)~TerraMind-1.0 (large); 
    e)~Prithvi-EO-2.0 (300M); 
    f)~Prithvi-EO-2.0 (600M); 
    g)~SSL4EO ViT-S (Dino);
    h)~SSL4EO ViT-S (MoCo). 
    i)~CopernicusFM; and
    j)~Panopticon.\\
    Figure partially overlaps with \cref{fig:corr_large}, but we visualize all data for ease of comparison.}
    \label{fig:corr_full}
\end{figure*}
We provide an extended feature correlation analysis across ten \ac{RS} \acp{FM} and four datasets in \cref{fig:corr_full}.
The analysis reveals that distinct correlation shapes remain consistent across both model sizes and datasets, indicating that these patterns are fundamental characteristics of training paradigms rather than task-specific adaptations.

The DOFA models (\cref{fig:corr_full-dofa,fig:corr_full-dofa-large}) exhibit a characteristic correlation shape across all datasets: highly organized feature correlations at small scales transitioning to near-independence at full scale.
Notably, the DOFA (base) model shows less correlated features at small scales but more at large scales compared to the DOFA (large), resulting in a less steep transition slope.
Despite this difference in slope steepness, both models share the same fundamental correlation shape that remains consistent across binary (m-brick-kiln), intermediate (m-eurosat), fine-grained (m-so2sat), and multi-label (m-bigearthnet) tasks.
This consistent pattern, combined with largely monotonic performance improvement across scales, confirms that additional uncorrelated dimensions at full scale contribute useful orthogonal information despite the correlation collapse.

The TerraMind-1.0 models (\cref{fig:corr_full-terramind,fig:corr_full-terramind-large}) demonstrate identical correlation shapes across both base and large sizes, characterized by smooth and mostly monotonic decrease from moderate to low correlations.
The TerraMind-1.0 (base) model maintains higher correlation values across all scales compared to TerraMind-1.0 (large), and exhibits a slightly smoother slope, but the fundamental shape remains unchanged.
This shape consistency across model sizes and datasets indicates that TerraMind-1.0's training paradigm induces a stable feature organization strategy where core discriminative features are efficiently encoded in early dimensions, with larger models simply distributing this information more independently.

Prithvi-EO-2.0 models (\cref{fig:corr_full-prithvi300,fig:corr_full-prithvi600}) exhibit the distinctive non-monotonic U-shaped correlation pattern across both Prithvi-EO-2.0 (300M) and Prithvi-EO-2.0 (600M) and all datasets.
While the Prithvi-EO-2.0 (300M) model shows a less pronounced rebound at large scales, not reaching as high correlations as the Prithvi-EO-2.0 (600M) variant, the U-shaped character remains clearly visible.
This consistent U-shape, characterized by moderate initial correlations, drops at low-to-intermediate scales, and subsequent rebounds at full scale, suggests that Prithvi's temporal training objective may induce systematic scale-dependent feature reorganization regardless of model capacity or downstream task complexity.

The SSL4EO ViT-S models (\cref{fig:corr_full-ssl4eo_dino,fig:corr_full-ssl4eo_moco}) show a roughly linear decrease in mean feature correlation across scales, but the total reduction is much smaller than for the other models.
This is consistent with their limited slimmability, as there is little redundancy to remove when correlations stay high at small scales.
Two factors may contribute: the smaller overall model size and the smaller embedding dimension.
However, these models cannot be compared directly to the families above, as they are not different scales of the same training method, architecture, and dataset.
The shape is also consistent across all four datasets and both models.

CopernicusFM (\cref{fig:corr_full-copernicusfm}) reaches a minimum in mean correlation around scale 0.6.
This coincides with its slimming profile, where downstream performance on some datasets is slightly higher at similar scales.

Panopticon (\cref{fig:corr_full-panopticon}) shows a smooth, strictly declining curve, which fits its slimming profile.
We would like to note, however, that due to its large patch embedding layer, the curve is not strictly comparable to those of the other models (see Appendix~\ref{sec:slimdims}).

The consistency of these correlation shapes across model sizes and datasets suggests that, within the three model families (DOFA, TerraMind, and Prithvi), each develops a characteristic feature organization that is stable across downstream tasks.
In details, DOFA's multi-modal training coincides with high-to-low correlation transitions, TerraMind-1.0's land cover focus with a smooth, mostly monotonic decrease, and Prithvi-EO-2.0's temporal modeling with a non-monotonic U-shaped pattern.
The remaining models do not allow a family comparison, but they also keep their correlation shape across all four datasets.
\section{Additional Experiments on Explained Variance}\label{sec:add_evr}
\begin{figure*}
    \centering
    \noindent\includegraphics[width=0.98\linewidth]{figures/plots/legends/post_hoc_slimming_existing_fms_legend_only.pdf}\par
    
    % Column headers
    %\hspace{8mm}%
    %\makebox[0.3\textwidth]{\textsf{\textit{\underline{\textbf{0.33}}}}}%
    %\makebox[0.3\textwidth]{\textsf{\textit{\underline{\textbf{0.67}}}}}%
    %\makebox[0.3\textwidth]{\textsf{\textit{\underline{\textbf{1.00}}}}}%

    % Row 1: m-brick-kiln
    \rowgrouplabel[10.5mm]{m-brick-kiln}\hspace{2mm}%
    \begin{subfigure}[t]{0.3\textwidth}
        \centering
        \tlabeledimage
            {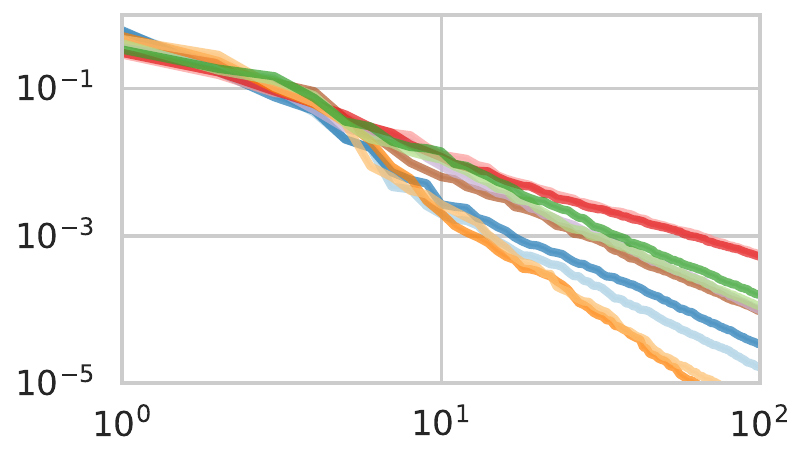}
            {Expl. Var. Ratio}
            {\phantom{Principal Component}}
        \vspace*{-6mm}
        %\caption{}
        %\label{fig:evr_033_appx_brick-kiln}
    \end{subfigure}%
    \begin{subfigure}[t]{0.3\textwidth}
        \centering
        \tlabeledimage
            {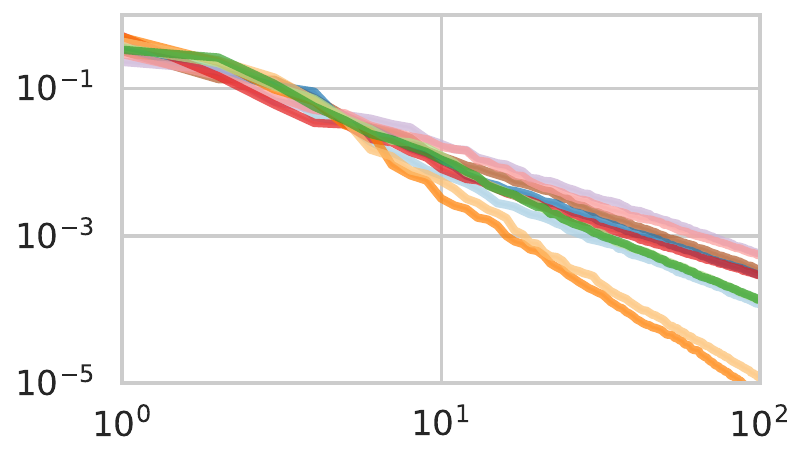}
            {\phantom{Expl. Var. Ratio}}
            {\phantom{Principal Component}}
        \vspace*{-6mm}
        %\caption{}
        %\label{fig:evr_067_appx_brick-kiln}
    \end{subfigure}%
    \begin{subfigure}[t]{0.3\textwidth}
        \centering
        \tlabeledimage
            {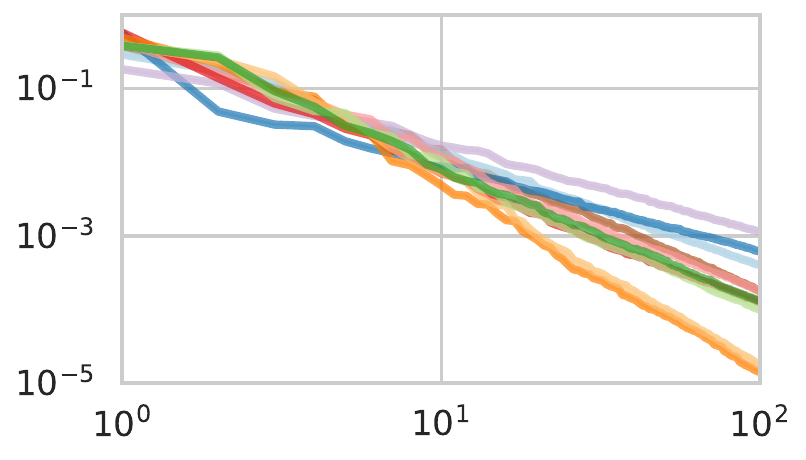}
            {\phantom{Expl. Var. Ratio}}
            {\phantom{Principal Component}}
        \vspace*{-6mm}
        %\caption{}
        %\label{fig:evr_100_appx_brick-kiln}
    \end{subfigure}%

    % Row 2: m-eurosat
    \rowgrouplabel[11mm]{m-eurosat}\hspace{2mm}%
    \begin{subfigure}[t]{0.3\textwidth}
        \centering
        \tlabeledimage
            {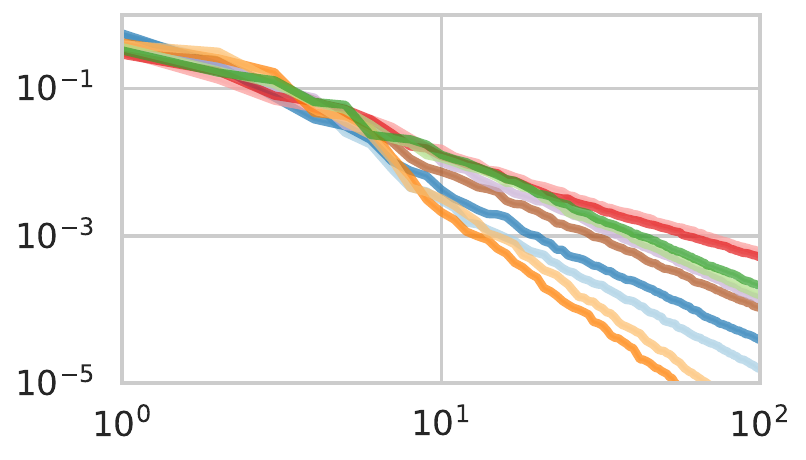}
            {Expl. Var. Ratio}
            {\phantom{Principal Component}}
        \vspace*{-6mm}
        %\caption{}
        %\label{fig:evr_033_appx_eurosat}
    \end{subfigure}%
    \begin{subfigure}[t]{0.3\textwidth}
        \centering
        \tlabeledimage
            {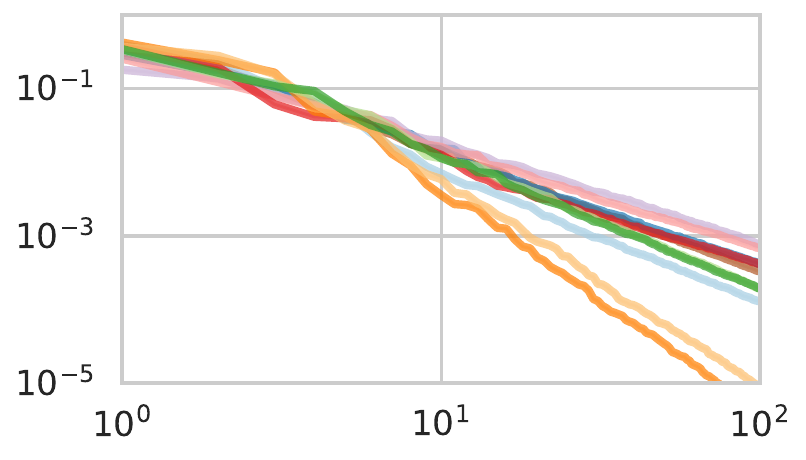}
            {\phantom{Expl. Var. Ratio}}
            {\phantom{Principal Component}}
        \vspace*{-6mm}
        %\caption{}
        %\label{fig:evr_067_appx_eurosat}
    \end{subfigure}%
    \begin{subfigure}[t]{0.3\textwidth}
        \centering
        \tlabeledimage
            {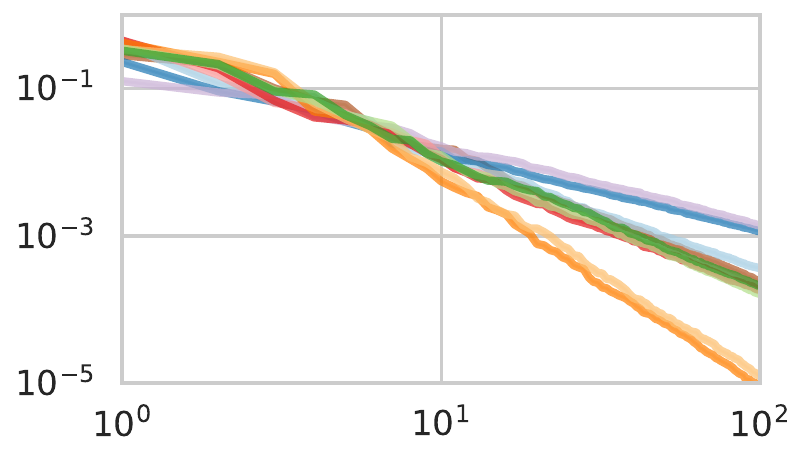}
            {\phantom{Expl. Var. Ratio}}
            {\phantom{Principal Component}}
        \vspace*{-6mm}
        %\caption{}
        %\label{fig:evr_100_appx_eurosat}
    \end{subfigure}%

    % Row 3: m-so2sat
    \rowgrouplabel[12mm]{m-so2sat}\hspace{2mm}%
    \begin{subfigure}[t]{0.3\textwidth}
        \centering
        \tlabeledimage
            {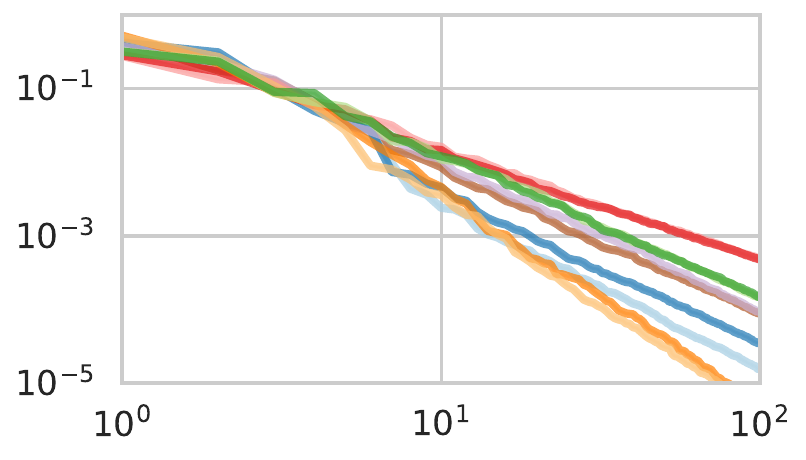}
            {Expl. Var. Ratio}
            {\phantom{Principal Component}}
        \vspace*{-6mm}
        %\caption{}
        %\label{fig:evr_033_appx_so2sat}
    \end{subfigure}%
    \begin{subfigure}[t]{0.3\textwidth}
        \centering
        \tlabeledimage
            {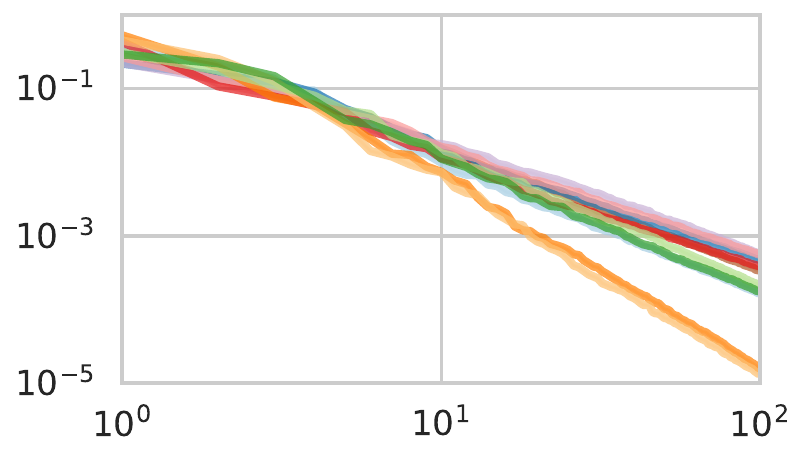}
            {\phantom{Expl. Var. Ratio}}
            {\phantom{Principal Component}}
        \vspace*{-6mm}
        %\caption{}
        %\label{fig:evr_067_appx_so2sat}
    \end{subfigure}%
    \begin{subfigure}[t]{0.3\textwidth}
        \centering
        \tlabeledimage
            {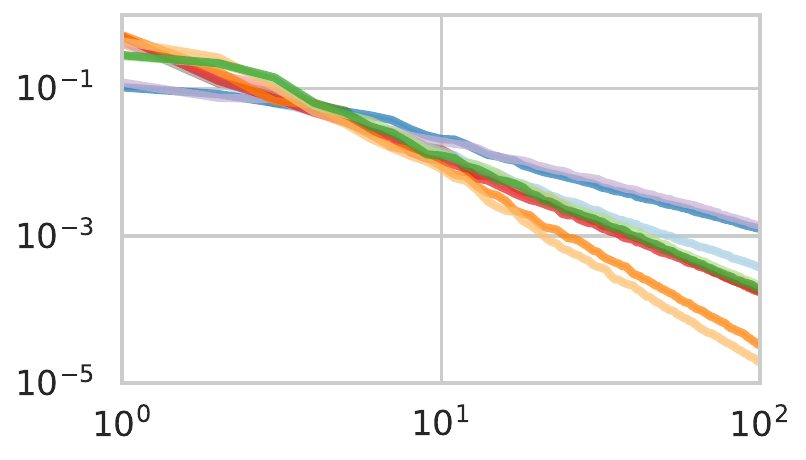}
            {\phantom{Expl. Var. Ratio}}
            {\phantom{Principal Component}}
        \vspace*{-6mm}
        %\caption{}
        %\label{fig:evr_100_appx_so2sat}
    \end{subfigure}%

    % Row 4: m-bigearthnet
    \rowgrouplabel[8mm]{m-bigearthnet}\hspace{2mm}%
    \begin{subfigure}[t]{0.3\textwidth}
        \centering
        \tlabeledimage
            {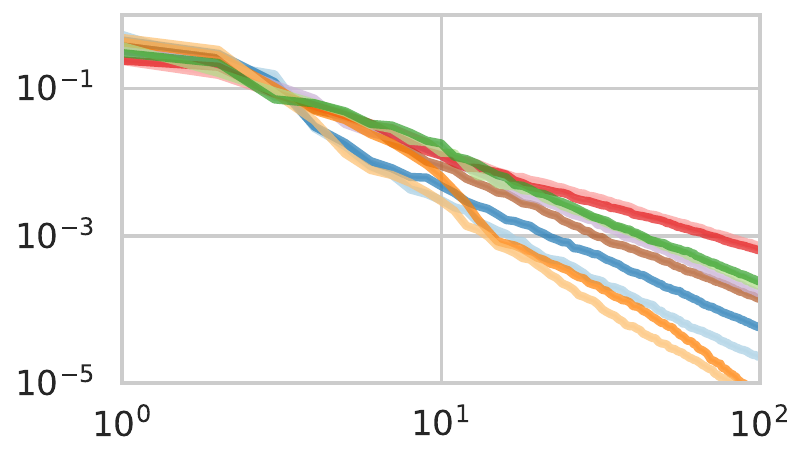}
            {Expl. Var. Ratio}
            {\phantom{Principal Component}}
        \vspace*{-3mm}
        \caption{}
        \label{fig:evr_033_appx_bigearthnet}
    \end{subfigure}%
    \begin{subfigure}[t]{0.3\textwidth}
        \centering
        \tlabeledimage
            {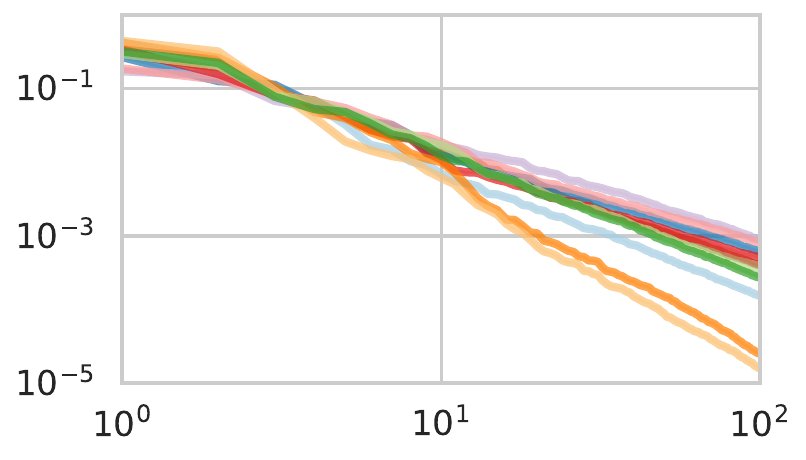}
            {\phantom{Expl. Var. Ratio}}
            {Principal Component}
        \vspace*{-3mm}
        \caption{}
        \label{fig:evr_067_appx_bigearthnet}
    \end{subfigure}%
    \begin{subfigure}[t]{0.3\textwidth}
        \centering
        \tlabeledimage
            {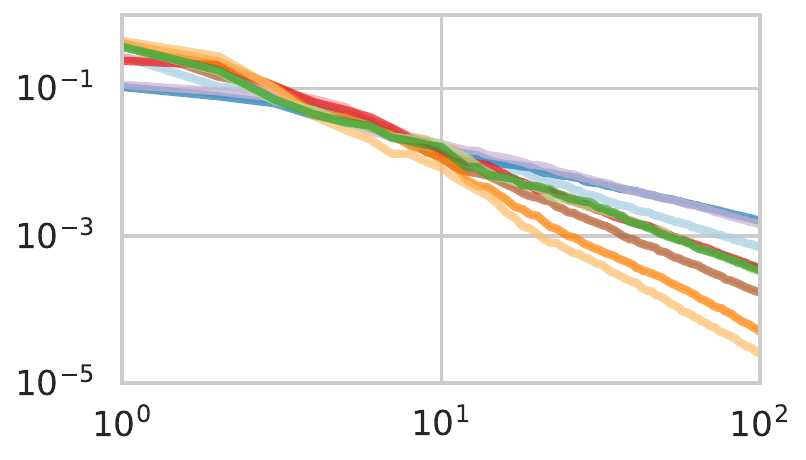}
            {\phantom{Expl. Var. Ratio}}
            {\phantom{Principal Component}}
        \vspace*{-3mm}
        \caption{}
        \label{fig:evr_100_appx_bigearthnet}
    \end{subfigure}%

    \caption{%
        \ac{EVR} with respect to the singular value rank index for ten \ac{RS}
        \acp{FM} on four classification datasets from the geobench benchmark collection at three
        relative compute requirements: 
        a)~at 0.33; 
        b)~at 0.67; and 
        c)~1.0 on (from top to bottom): m-brick-kiln; m-eurosat; m-so2sat; and m-bigearthnet.
    }
    \label{fig:evr_appx}
\end{figure*}
We provide an extended \ac{EVR} analysis across ten \ac{RS} \acp{FM} and four datasets in \cref{fig:evr_appx}.
The extended analysis confirms the log-log slope patterns observed in \cref{fig:loglog_slope_evr} while revealing model family-specific consistency across datasets and scales.
Both Prithvi-EO-2.0 and both TerraMind-1.0 models exhibit remarkable internal consistency: within each model family, both sizes demonstrate nearly identical \ac{EVR} decay patterns across all datasets and scales.
This consistency indicates that these models' variance distribution strategies are influenced by their training objectives.
The Prithvi-EO-2.0 models maintain their characteristic non-monotonic pattern (becoming more isotropic at intermediate scales before reconsolidating at full scale) across all datasets, while the TerraMind-1.0 models preserve stable intermediate anisotropy regardless of whether evaluated on binary, multi-class, or multi-label tasks.
In contrast, the DOFA models exhibit systematic scale-dependent behavior that varies between base and large variants but remains consistent across datasets.
Both DOFA models demonstrate steeper slopes (more concentrated variance) at smaller scales and shallower slopes (more distributed variance) at larger scales across all four datasets, with the base model maintaining systematically steeper slopes than the large variant at equivalent relative compute budgets.
This pattern confirms the observation that DOFA's multi-modal training drives monotonic variance redistribution as capacity increases, with larger models distributing variance more evenly.
The consistency of this pattern across binary (m-brick-kiln), intermediate (m-eurosat), fine-grained (m-so2sat), and multi-label (m-bigearthnet) tasks demonstrates that DOFA's scaling strategy is an architectural property rather than a task-specific adaptation.
The four additional models (SSL4EO ViT-S DINO, SSL4EO ViT-S MoCo, CopernicusFM, and Panopticon) do not belong to a model family, so within-family consistency cannot be assessed. 
They nonetheless show stable \ac{EVR} decay patterns across all four datasets, indicating that their variance organization is largely independent of the downstream task.

These findings reinforce the intuition that training objectives may influence how models organize information across scales, with this organization remaining stable across diverse downstream task requirements.
\section{Failure Modes and Edge Cases}\label{sec:failure}
\paragraph{Class-wise average precision of m-bigearthnet}
\begin{figure}[ht]
    \centering
    \noindent\includegraphics[width=0.98\linewidth]{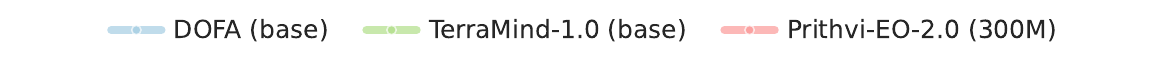}\par\vspace{-2mm}
    \noindent\includegraphics[width=0.98\linewidth]{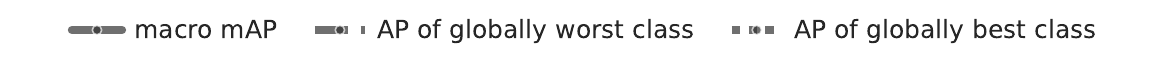}\par\vspace{-2mm}
    \begin{subfigure}[t]{0.3\textwidth}
        \centering
        \tlabeledimage
            {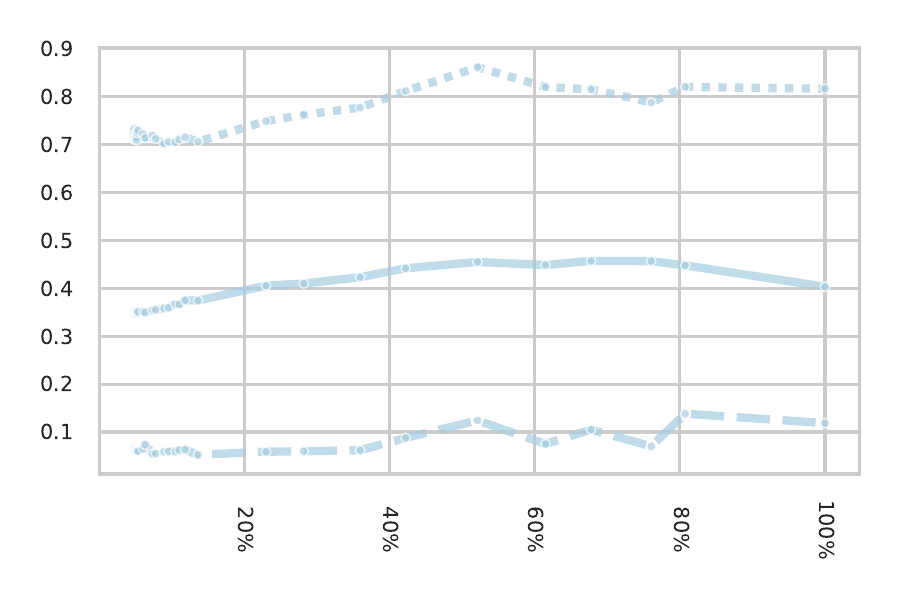}
            {\ac{mAP}}
            {\phantom{Rel. Compute Requirement}}
        \vspace*{-3mm}
        \caption{}
        \label{fig:classwise_dofa}
    \end{subfigure}%
    \begin{subfigure}[t]{0.3\textwidth}
        \centering
        \tlabeledimage
            {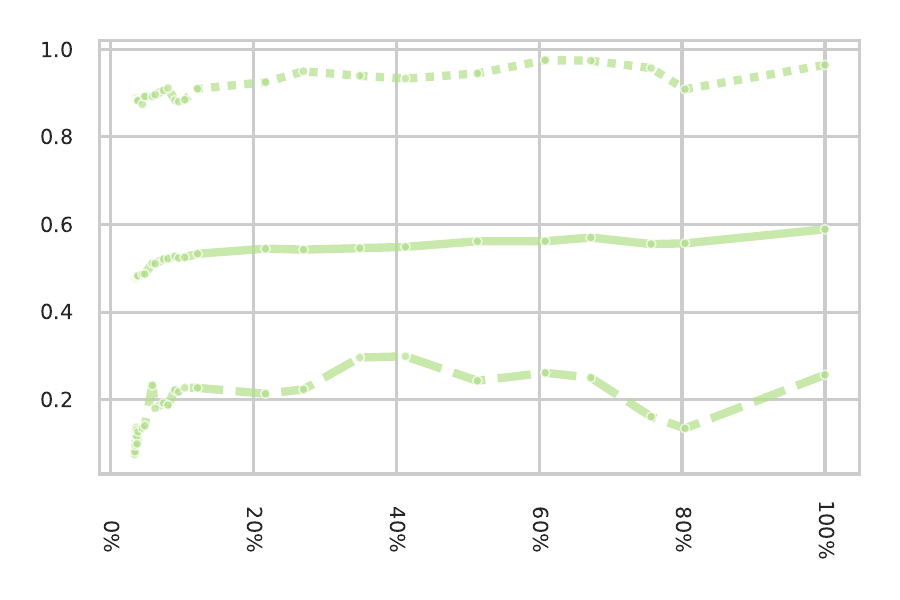}
            {Accuracy}
            {Rel. Compute Requirement}
        \vspace*{-3mm}
        \caption{}
        \label{fig:classwise_terramind}
    \end{subfigure}%
    \begin{subfigure}[t]{0.3\textwidth}
        \centering
        \tlabeledimage
            {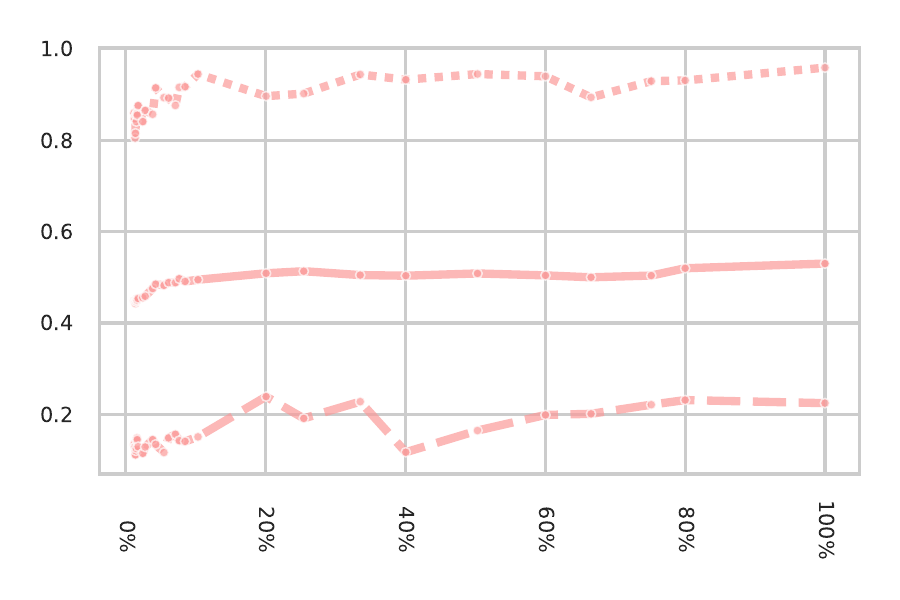}
            {\phantom{\ac{mAP}}}
            {\phantom{Rel. Compute Requirement}}
        \vspace*{-3mm}
        \caption{}
        \label{fig:classwise_prithvi}
    \end{subfigure}%
    \caption{KNN classification performance with respect to relative compute requirements on 
    the test split of m-bigearthnet evaluating the global best, worst, and macro-mean averaged precision in multi-label classification of:
    a)~DOFA (base); 
    b)~TerraMind-1.0 (base); and
    c)~Prithvi-EO-2.0 (300M).}
    \label{fig:classwise}
    \vspace{-3mm}
\end{figure}

To check whether aggregate \ac{mAP} hides uneven degradation, we report the class-wise \ac{AP} of the best and worst class of the full-width model at each scale during \ac{KNN} evaluation in \cref{fig:classwise}. 
The easiest classes keep most of their \ac{AP} at \SI{1}{\percent} compute (\SIrange{89}{92}{\percent} of full-width \ac{AP} across DOFA, Prithvi-EO-2.0, and TerraMind). The hardest classes lose a larger share: Prithvi and TerraMind retain roughly \SIrange{50}{60}{\percent} of their full-width \ac{AP}, while DOFA’s hardest class is already very low (0.11) at full width. 
Slimming therefore degrades classes unevenly, and aggregate metrics partly conceal this, but hard classes lose accuracy gradually rather than collapsing.

\paragraph{Geographical shift in m-so2sat}
\begin{figure}[ht]
    \centering
    \noindent\includegraphics[width=0.98\linewidth]{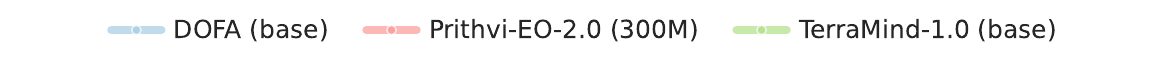}\par\vspace{-2mm}
    \noindent\includegraphics[width=0.98\linewidth]{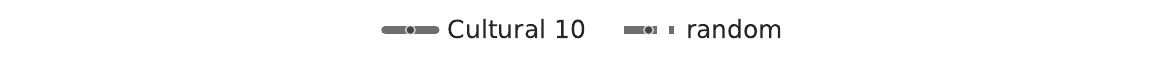}\par\vspace{-2mm}
    \begin{subfigure}[t]{0.3\textwidth}
        \centering
        \tlabeledimage
            {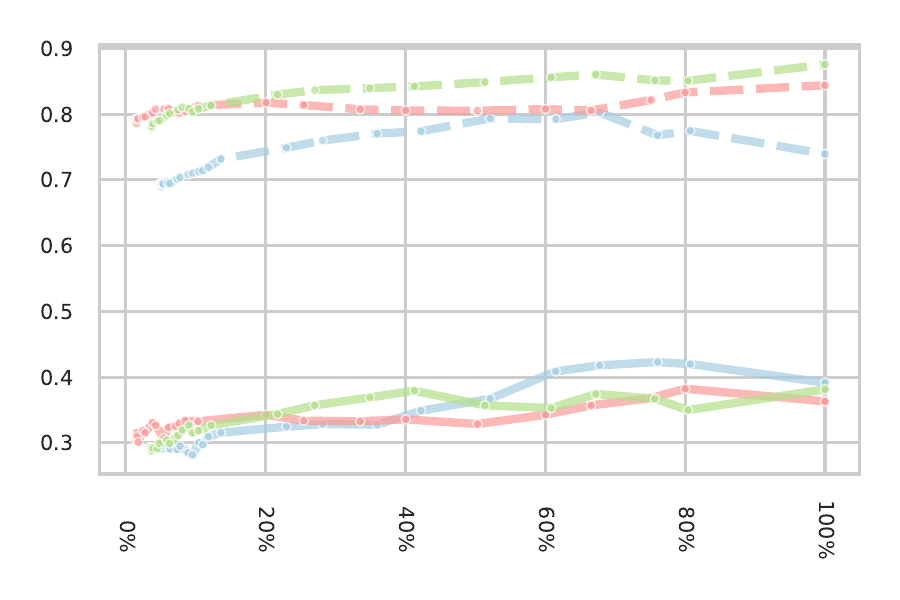}
            {Accuracy}
            {Rel. Compute Requirement}
        \vspace*{-3mm}
        %\caption{}
        \label{fig:geoshift_sub}
    \end{subfigure}%
    \caption{KNN classification performance with respect to relative compute requirements on 
    m-so2sat under random and predefined disjoint (Cultural 10) geographic shifts of:
    DOFA (base); 
    TerraMind-1.0 (base); and
    Prithvi-EO-2.0 (300M).}
    \label{fig:geoshift}
    \vspace{-3mm}
\end{figure}
To quantify the effect of geographic shift, we compare the geographically disjoint Cultural-10 split of the geobench benchmark collection against a random split. 
The random split pools train and test data, reshuffles, resamples to identical sizes, and averages five draws, using the same \ac{KNN} protocol as used in all other experiments. 
As shown in \cref{fig:geoshift}, retention at minimum compute is higher on the random split (about \SI{93}{\percent} for DOFA (base), accuracy drops from \SIrange{74}{69}{\percent}) than on Cultural-10 (about \SI{77}{\percent}, accuracy drops from \SIrange{39}{30}{\percent}), with the same pattern for Prithvi-EO-2.0 (300M) and TerraMind-1.0 (base). 
Geographic shift thus reduces the measured redundancy, but substantial redundancy remains under geographically disjoint evaluation.

\paragraph{Class imbalance}
\begin{figure}[ht]
    \centering
    \noindent\includegraphics[width=0.98\linewidth]{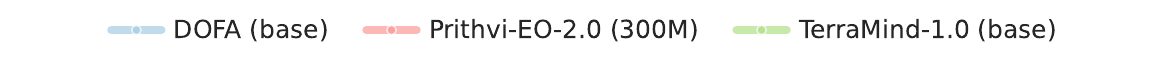}\par\vspace{-2mm}
    \noindent\includegraphics[width=0.98\linewidth]{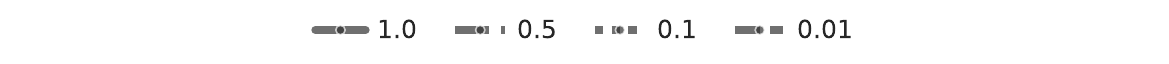}\par\vspace{-2mm}
    \begin{subfigure}[t]{0.3\textwidth}
        \centering
        \tlabeledimage
            {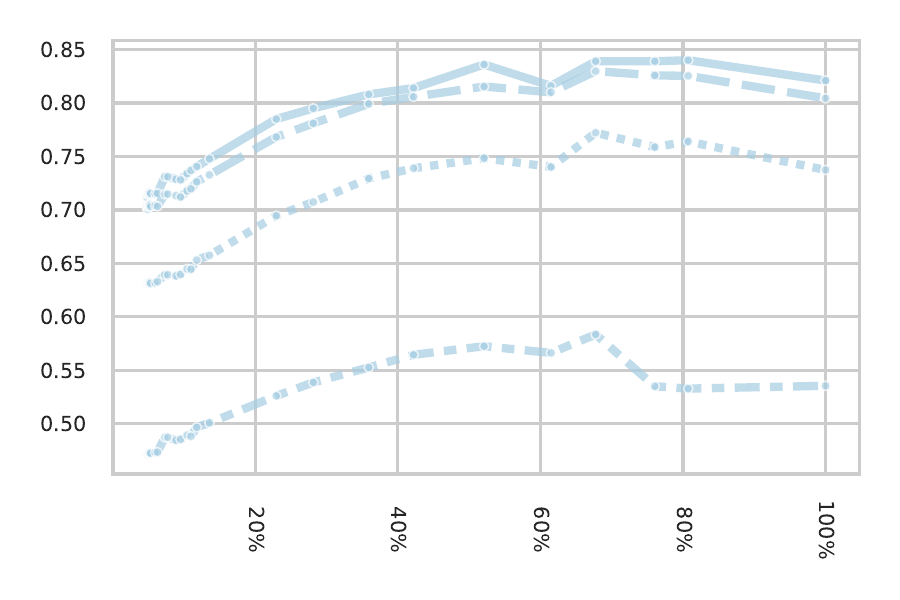}
            {Accuracy}
            {\phantom{Rel. Compute Requirement}}
        \vspace*{-3mm}
        \caption{}
        \label{fig:imbalance_dofa}
    \end{subfigure}%
    \begin{subfigure}[t]{0.3\textwidth}
        \centering
        \tlabeledimage
            {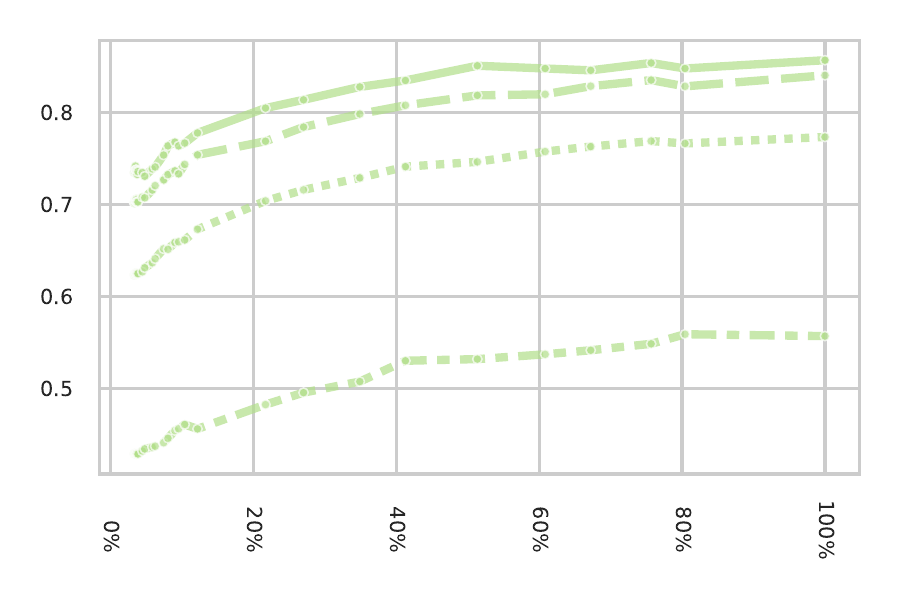}
            {\phantom{Accuracy}}
            {Rel. Compute Requirement}
        \vspace*{-3mm}
        \caption{}
        \label{fig:imbalance_terramind}
    \end{subfigure}%
    \begin{subfigure}[t]{0.3\textwidth}
        \centering
        \tlabeledimage
            {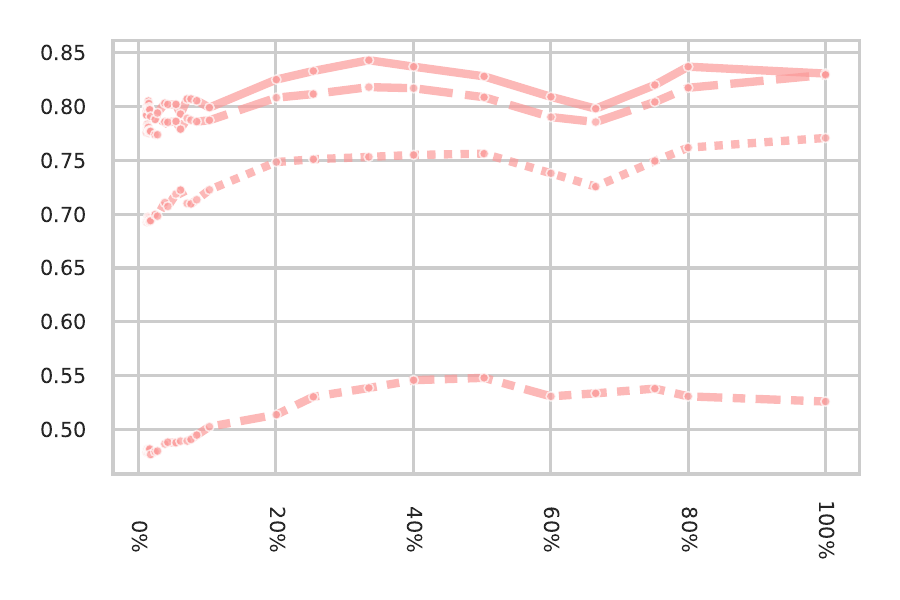}
            {\phantom{Accuracy}}
            {\phantom{Rel. Compute Requirement}}
        \vspace*{-3mm}
        \caption{}
        \label{fig:imbalance_prithvi}
    \end{subfigure}%
    \caption{KNN classification performance with respect to relative compute requirements on 
    the test split of m-eurosat evaluating a synthetic class imbalance in the training set of:
    a)~DOFA (base); 
    b)~TerraMind-1.0 (base); and
    c)~Prithvi-EO-2.0 (300M).}
    \label{fig:imbalance}
    \vspace{-3mm}
\end{figure}

To test whether slimming interacts with class imbalance, we impose an exponential imbalance on the training set of m-eurosat: class c keeps a fraction $r^{c/9}$ of its samples, so class 0 keeps all samples and class 9 keeps the imbalance ratio $r$, with $r$ between 0.01 and 1.0 and the classes in between decaying geometrically. 
The result is averaged over five subsampling seeds.
The result of this subsampling and the following \ac{KNN} evaluations are shown in \cref{fig:imbalance}.
Absolute accuracy drops substantially at every width as imbalance grows. 
The retention curves stay roughly parallel across imbalance ratios for DOFA (base) and Prithvi-EO-2.0 (300M). 
We therefore conclude that slimming does not interact strongly with imbalance for these models. 
TerraMind (base) shows a mild interaction at the most extreme imbalance ratio (0.01), where retention at \SI{1}{\percent} compute falls to about \SI{77}{\percent} from \SI{86}{\percent} when balanced.
\section{Computational Resources}\label{sec:compute}
All \ac{SSL} pretraining experiments were conducted on $4\times$NVIDIA A100 GPU or $4\times$NVIDIA H200 GPU.
Regular pretraining runs required approximately one day; slimmable pretraining runs required approximately four days due to the multi-width training protocol.
Post-hoc slimmability evaluation, feature extraction, and representational analyses (\ac{KNN} classification, \ac{EVR}, feature correlation, effective dimensionality) can be performed on CPU.
All such experiments were run using 16 CPU cores and \SI{128}{\giga\byte} RAM.
Feature extraction across all compute scales can be completed in under 30 minutes per model; GPU acceleration is beneficial for this step and was used where available (NVIDIA H200, A100, and A40 GPUs).
Downstream evaluation runtimes vary by task type.

\ac{KNN} classification and change detection complete within a few hours per model.
Linear probing requires moderately longer due to the additional training loop.
Semantic segmentation is the most compute-intensive downstream task, requiring up to 8 hours per model due to dataset size and the higher complexity of dense prediction relative to \ac{KNN} evaluation.
The reported runtimes reflect the final experimental runs; total compute including preliminary and repeated experiments is not tracked.
\section{Licenses and Assets}\label{sec:licenses}
The pretraining data used for learned slimmability experiments is derived from a subsampled subset of Major-TOM Core-S2L2A \citep{francis2024major}, available at \url{https://huggingface.co/datasets/Major-TOM/Core-S2L2A} under the Creative Commons Attribution-ShareAlike 4.0 International license (CC-BY-SA 4.0).
Downstream evaluation datasets are sourced from the geobench benchmark collection \citep{lacoste2023geobench}, which is released under the Apache 2.0 license (\url{https://github.com/ServiceNow/geo-bench}).
All pretrained \ac{RS} \acp{FM} evaluated in this work are accessed via TerraTorch \citep{gomes2025terratorch}, which is released under the Apache 2.0 license (\url{https://github.com/IBM/terratorch}).

%%%%%%%%%%%%%%%%%%%%%%%%%%%%%%%%%%%%%%%%%%%%%%%%%%%%%%%%%%%%

\makeatletter
\if@preprint
  % Do nothing (or add preprint-specific notes here)
\else
  \newpage
  \section*{NeurIPS Paper Checklist}
\begin{enumerate}

\item {\bf Claims}
    \item[] Question: Do the main claims made in the abstract and introduction accurately reflect the paper's contributions and scope?
    \item[] Answer: \answerYes{} % Replace by \answerYes{}, \answerNo{}, or \answerNA{}.
    \item[] Justification: The main claims made in the abstract and introduction are in line with the contributions and scope of the paper. Additionally, a paragraph on scope limitations is included in the conclusion.

\item {\bf Limitations}
    \item[] Question: Does the paper discuss the limitations of the work performed by the authors?
    \item[] Answer: \answerYes{} % Replace by \answerYes{}, \answerNo{}, or \answerNA{}.
    \item[] Justification: A paragraph on limitations is included in the conclusion section.

\item {\bf Theory assumptions and proofs}
    \item[] Question: For each theoretical result, does the paper provide the full set of assumptions and a complete (and correct) proof?
    \item[] Answer: \answerNA{} % Replace by \answerYes{}, \answerNo{}, or \answerNA{}.
    \item[] Justification: The paper does not include theoretical results. 

    \item {\bf Experimental result reproducibility}
    \item[] Question: Does the paper fully disclose all the information needed to reproduce the main experimental results of the paper to the extent that it affects the main claims and/or conclusions of the paper (regardless of whether the code and data are provided or not)?
    \item[] Answer: \answerYes{} % Replace by \answerYes{}, \answerNo{}, or \answerNA{}.
    \item[] Justification: All relevant information to reproduce the results are provided. Additionally, the code for full reproduction is available at \url{https://github.com/hackelle/slimmable-remote-sensing-foundation-models} under MIT license.

\item {\bf Open access to data and code}
    \item[] Question: Does the paper provide open access to the data and code, with sufficient instructions to faithfully reproduce the main experimental results, as described in supplemental material?
    \item[] Answer: \answerYes{} % Replace by \answerYes{}, \answerNo{}, or \answerNA{}.
    \item[] Justification: The code required to reproduce training and evaluation results is available at \url{https://github.com/hackelle/slimmable-remote-sensing-foundation-models} under MIT license.

\item {\bf Experimental setting/details}
    \item[] Question: Does the paper specify all the training and test details (e.g., data splits, hyperparameters, how they were chosen, type of optimizer) necessary to understand the results?
    \item[] Answer: \answerYes{} % Replace by \answerYes{}, \answerNo{}, or \answerNA{}.
    \item[] Justification: The experimental setup follow standard training splits and parameters, which are detailed in the respective sections.

\item {\bf Experiment statistical significance}
    \item[] Question: Does the paper report error bars suitably and correctly defined or other appropriate information about the statistical significance of the experiments?
    \item[] Answer: \answerYes{} % Replace by \answerYes{}, \answerNo{}, or \answerNA{}.
    \item[] Justification: Error bars show 2-sigma across random seeds and are provided for all linear probe and finetuning classification evaluations; pretraining and segmentation experiments are excluded due to computational cost and \ac{KNN} results are deterministic. Error bars may be smaller than the line width in some figures.

\item {\bf Experiments compute resources}
    \item[] Question: For each experiment, does the paper provide sufficient information on the computer resources (type of compute workers, memory, time of execution) needed to reproduce the experiments?
    \item[] Answer: \answerYes{} % Replace by \answerYes{}, \answerNo{}, or \answerNA{}.
    \item[] Justification: A section detailing the computational resources used is included in the appendix.
    
\item {\bf Code of ethics}
    \item[] Question: Does the research conducted in the paper conform, in every respect, with the NeurIPS Code of Ethics \url{https://neurips.cc/public/EthicsGuidelines}?
    \item[] Answer: \answerYes{} % Replace by \answerYes{}, \answerNo{}, or \answerNA{}.
    \item[] Justification: The research conforms with the NeurIPS Code of Ethics. We note that m-bigearthnet results rely on the geobench benchmark, which uses BigEarthNet v1.0 (43 labels), a superseded dataset version; this is a limitation of the benchmark dependency, not an ethical violation. Broader societal risks associated with AI applied to remote sensing data (e.g., surveillance) are discussed in the paper.

\item {\bf Broader impacts}
    \item[] Question: Does the paper discuss both potential positive societal impacts and negative societal impacts of the work performed?
    \item[] Answer: \answerYes{} % Replace by \answerYes{}, \answerNo{}, or \answerNA{}.
    \item[] Justification: The paper contains a section on practical implications as well as an impact statement.
    
\item {\bf Safeguards}
    \item[] Question: Does the paper describe safeguards that have been put in place for responsible release of data or models that have a high risk for misuse (e.g., pre-trained language models, image generators, or scraped datasets)?
    \item[] Answer: \answerNA{} % Replace by \answerYes{}, \answerNo{}, or \answerNA{}.
    \item[] Justification: There are no models or datasets released in this work.

\item {\bf Licenses for existing assets}
    \item[] Question: Are the creators or original owners of assets (e.g., code, data, models), used in the paper, properly credited and are the license and terms of use explicitly mentioned and properly respected?
    \item[] Answer: \answerYes{} % Replace by \answerYes{}, \answerNo{}, or \answerNA{}.
    \item[] Justification: A section detailing the licenses of all datasets and models used is included in the appendix. All used datasets and models are cited with the appropriate specifications.

\item {\bf New assets}
    \item[] Question: Are new assets introduced in the paper well documented and is the documentation provided alongside the assets?
    \item[] Answer: \answerYes{} % Replace by \answerYes{}, \answerNo{}, or \answerNA{}.
    \item[] Justification: All code required to reproduce training and evaluation results is publicly available; no model weights or datasets will be released.

\item {\bf Crowdsourcing and research with human subjects}
    \item[] Question: For crowdsourcing experiments and research with human subjects, does the paper include the full text of instructions given to participants and screenshots, if applicable, as well as details about compensation (if any)? 
    \item[] Answer: \answerNA{} % Replace by \answerYes{}, \answerNo{}, or \answerNA{}.
    \item[] Justification: The paper does not involve crowdsourcing nor research with human subjects.

\item {\bf Institutional review board (IRB) approvals or equivalent for research with human subjects}
    \item[] Question: Does the paper describe potential risks incurred by study participants, whether such risks were disclosed to the subjects, and whether Institutional Review Board (IRB) approvals (or an equivalent approval/review based on the requirements of your country or institution) were obtained?
    \item[] Answer: \answerNA{} % Replace by \answerYes{}, \answerNo{}, or \answerNA{}.
    \item[] Justification: The paper does not involve crowdsourcing nor research with human subjects.

\item {\bf Declaration of LLM usage}
    \item[] Question: Does the paper describe the usage of LLMs if it is an important, original, or non-standard component of the core methods in this research? Note that if the LLM is used only for writing, editing, or formatting purposes and does \emph{not} impact the core methodology, scientific rigor, or originality of the research, declaration is not required.
    %this research? 
    \item[] Answer: \answerNA{} % Replace by \answerYes{}, \answerNo{}, or \answerNA{}.
    \item[] Justification: The core method development in this research does not involve LLMs as any important, original, or non-standard components

\end{enumerate}
\fi
\makeatother

\end{document}